# The Digital Synaptic Neural Substrate: A New Approach to Computational Creativity

Azlan Iqbal[1], Matej Guid[2], Simon Colton[3],
Jana Krivec[4], Shazril Azman[5] and Boshra Haghighi[6]

ABSTRACT

We introduce a new artificial intelligence (AI) approach or technique termed, the 'Digital Synaptic Neural Substrate' (DSNS). This technique uses selected attributes from objects in various domains (e.g. chess problems, classical music, renowned artworks) and recombines them in such a way as to generate new attributes that can then, in principle, be used to create novel objects of creative value to humans relating to any one of the source domains. This allows some of the burden of creative content generation to be passed from humans to machines. The approach was tested primarily in the domain of chess problem composition. We used the DSNS technique to automatically compose numerous sets of chess problems based on attributes extracted and recombined from chess problems and tournament games by humans, renowned paintings, computer-evolved abstract art, photographs of people, and classical music tracks. The quality of these generated chess problems was then assessed automatically using an existing and experimentally-validated computational chess aesthetics model. They were also assessed by human experts in the domain. The results suggest that attributes collected and recombined from chess and other domains using the DSNS approach can indeed be used to automatically generate chess problems of reasonably high aesthetic quality. In particular, a low quality chess source (i.e. tournament game sequences between weak players) used in combination with actual photographs of people was able to produce three-move chess problems of comparable quality or better to those generated using a high quality chess source (i.e. published compositions by human experts), and more efficiently as well. Why information from a foreign domain can be integrated and functional in this way remains an open question for now. The DSNS approach is, *in principle*, scalable and applicable to any domain in which objects have attributes that can be represented using real numbers.

**Keywords**: artificial intelligence, creativity, brain, process, chess.

## 1 INTRODUCTION

Computational creativity can be classified as a relatively new sub-field of artificial intelligence (AI). In principle, it focuses on the generation of creative objects (e.g. music, visual art) using existing, modified and in some cases even new, typically domain-specific AI approaches or techniques. These objects are usually deemed to be creative from the perspective of humans sufficiently competent in the domain; otherwise known as 'experts' (Johnson, 2012). Such systems allow for some of the burden of creative content generation to be passed from humans to machines. They also have the potential of challenging humans with new ideas not considered, discovered or explored by humans before. The *way* these objects are produced may also be relevant to their creative value in some cases (Colton, 2008), but not necessarily. An object of creative value might be deemed so based on its utility or beauty (or both). Remarkable products of engineering and design should therefore not be excluded by default. Notably, there is also the distinction between a 'P-creative' idea or object and an 'H-creative' one, where the former is considered something new to the *person* who generated it whereas the latter is new *historically* (Ritchie, 2007; Boden, 2009).

There is still something significant in terms of design, at least, about computer systems that are capable of generating P-creative content even though the ultimate goal may be H-creative works. It can be argued that chance or serendipity plays an undeniable role in creativity as well (Pease et al., 2013) and to some extent this can be simulated. In any case, we will not here be drawn into philosophical musings about the 'true' nature of creativity or beauty (Sibley, 1959; Apter, 1977). We will instead abide by the standard and most reasonable conventions of assessment pertaining to creativity in the particular domain of investigation and also rely upon the judgements of human experts in that domain (Didierjean and Gobet, 2008). As with all computational approaches, the method must be described in sufficient detail so as to facilitate reproducibility, but due to the nature of creativity in *humans* – whom typically are not expected to explain themselves in such detail – this removes some of the 'mystery' and may affect the perception of quality with regard to the creative objects produced by the system.

[1] *College of Information Technology, Universiti Tenaga Nasional, Putrajaya Campus, Jalan IKRAM-UNITEN, 43000 Kajang, Selangor, Malaysia; e-mail: azlan@uniten.edu.my*

[2] *Faculty of Computer and Information Science, University of Ljubljana, Slovenia; e-mail: matej.guid@fri.uni-lj.si*

[3] *Department of Computing of Goldsmiths College, University of London; e-mail: s.colton@gold.ac.uk*

[4] *Jozef Stefan Institute, Department of Intelligent Systems, Jamova cesta 39, 1000 Ljubljana, Slovenia; e-mail krivec.jana@gmail.com*

[5] *College of Graduate Studies, Universiti Tenaga Nasional, Putrajaya Campus, Jalan IKRAM-UNITEN, 43000 Kajang, Selangor, Malaysia; e-mail: shazril@uniten.edu.my*

[6] *(Formerly with) College of Graduate Studies, Universiti Tenaga Nasional, Putrajaya Campus, Jalan IKRAM-UNITEN, 43000 Kajang, Selangor, Malaysia e-mail: haghighi.boshra@gmail.com*

Chess problems, the main domain of investigation in this paper, typically requires considerable expertise, experience, time and most importantly, creativity, to produce. For this reason, and given our experience with the game in previous research works, we sought to test the DSNS approach in terms of its ability to facilitate 'creativity' in the process of composing chess problems. The game of chess has always been a favorite domain of investigation for researchers in artificial intelligence, especially, and we found that to be true here as well. The reason is that it provides a controlled and computationally-amenable environment to perform experiments and test hypotheses, as will be evident later in the paper. Success in this domain, given its complexity, therefore suggests scalability of the DSNS approach and at least *potential* in others which are beyond the scope of the present introductory work. Section 2 provides a brief history and overview of some existing work in the field of computational creativity. Section 3 details the DSNS approach and the experimental methodology. In section 4 and its subsections we present and discuss the experimental work. Section 5 consolidates the experimental results in point form with a discussion about the possible limitations of the DSNS and how it may be applied in other domains. Section 6 concludes with directions for further work.

## 2 REVIEW

A good coverage of the concept and theories of creativity across various disciplines (primarily psychology) can be found in (Sternberg and Kaufman, 2010) but these are not particularly relevant here. The *concept* of creativity with regard to AI, however, is pertinent but not particularly new. It was mentioned in the proposal that lead to the famous 1956 Dartmouth Summer School often remembered as the birth of the field (Boden, 2009). Even so, the field was in its infancy back then and challenges specific to creativity as opposed to the 'mere' mechanization of tasks requiring human intelligence were generally unheard of (McCorduck, 2004; Levy, 2006; Ekbia, 2008; Warwick, 2011). Put simply, getting a computer to play *good* even though bland chess was enough of a challenge then than trying to make it play good and *creatively*, e.g. in a way that would surprise even the most stylistic human grandmasters, such as the late Mikhail Tal. Analogously, generating a relatively simple painting was enough of a computational challenge than creating a masterpiece (Cohen, 1999).

With the defeat of then-world chess champion Garry Kasparov to IBM's Deep Blue supercomputer in 1997 (Newborn, 1997) and the general successes of AI in many small but significant ways in other fields (e.g. medical diagnosis, law, stock trading), demand and expectations from machines grew. Even with the resounding success of computer chess today such as computer programs running on smartphones playing at the grandmaster level and computers being indispensable in the training of human grandmasters, some experts still wonder, *where is it going?* (Kasparov, 2014). The many popular science-fiction films such as *Colossus: The Forbin Project* (Sargent, 1970), *Blade Runner* (Scott, 1982), *Terminator 2* (Cameron, 1991), *The Matrix* (Wachowski and Wachowski, 1999) and *A.I.* (Spielberg, 2001) depicting intelligent and lifelike machines may also have had an influence on societal expectations from the field, regardless of their generally dystopian outlook.[7]

Computational creativity therefore began to receive more serious attention from academia at the turn of the second millennium (Colton and Steel, 1999; Buchanan, 2001) perhaps to address some of these expectations and new challenges, in particular those related to the burgeoning Internet (Battelle, 2006). While most applications in this area are generally small and task-specific (like with traditional AI), others are far more ambitious and aim to replicate the workings of the entire human brain. For example, to 'give rise' to creativity and perhaps even what is known as the technological 'singularity' (Holmes, 1996; Seung, 2012; Kurzweil, 2012). This is when AI supposedly will overtake human intelligence and radically alter the world we live in. Even an attempt to mimic the functioning of a cat's brain (Ananthanarayanan et al., 2009) was mired in controversy about its actual performance (Shachtman, 2009) so we can be fairly skeptical. There are also approaches that lie somewhere in between traditional AI and computational creativity such as IBM's *Watson* supercomputer which is apparently capable of extracting information and determining proper context (in English, at least). In short, it can 'read' and answer questions meaningfully (Chandrasekar, 2014). However, it cannot create any novel content of its own, like producing an article such as this.

[7] These productions are becoming more common and are dealing with even more sophisticated issues related to AI such as human-machine social interaction in the film, *Her* (Jonze, 2013; Saunders et al., 2013) and the transcendence of genuine human existence into the digital domain such as in the motion picture, *Transcendence* (Pfister, 2014).

In principle, many of the fundamental approaches or techniques[8] that have been developed in AI such as artificial neural networks, genetic algorithms and evolutionary computation (McCulloch and Pitts, 1943; Box, 1957; Fraser and Burnell, 1970, Back et al., 1997) can and have been applied to varying degrees in computationally creative systems (Abe et al., 2006; Terai and Nakagawa, 2009; Correia et al., 2013; Machado & Amaro, 2013). Such systems may also take the approach of combining information or knowledge from within the same domain using mathematical logic, statistical modeling or some form of machine learning, which is also used in mainstream AI (Cope, 2005; Eigenfeldt and Pasquier, 2013). A tempting idea is for a computationally creative system to also learn from its own 'experience', much like humans are thought to do (Iqbal, 2011; Grace et al., 2013). However, the methods these systems use tend to be domain or task-specific as well. There is nothing inherently wrong with this if the system performs well but it is certainly not any kind of *general* intelligence or 'creativity machine' that the human brain is; some brains more so than others.

In summary, the arsenal of approaches and techniques available in mainstream AI have trickled down to computational creativity and given rise to new types of applications, e.g. that generate creative objects, assess beauty or 'interestingness' (Iqbal et al., 2012; Pérez and Ortiz, 2013). At the same time, computational creativity researchers have found new ways to adapt mathematical logic and existing AI to suit their purposes. However, we could not find any approach documented in the literature that was able to successfully integrate discrete information from two or more unrelated domains so as to produce viable creative output in either one of the original sources. This is different from producing simultaneous outputs in more than one domain based on a particular set of inputs and with the help of an AI engine (Bengi and Ronchi, 2013). The successful integration of discrete information from multiple, unrelated domains for computational creativity purposes may be important if one subscribes to the notion that creativity in a particular domain can be borne out of 'inspiration' obtained from more than one domain. For instance, a musical piece composed that the composer says was inspired by the photo or memory of a beautiful companion. Whether or not such information in the brain is aggregated through systematicity (Phillips, 2014; Clement and Gentner, 1991) is presently unproven and therefore not assumed to be something our approach should be based on.

So, is codification of context-specific, high-level concepts really necessary for this or do low-level 'mindless' mathematical interactions suffice? The process of biological evolution, for instance, is said to be a mindless and semantic-independent one that actually produces minds and even 'free will' (Dennett, 1996, 2012) even though there is much debate regarding the nature of the latter, particularly in association with the notion of 'consciousness' (Chalmers, 1995, 2012; Harris, 2013). This is worth noting because consciousness and free will are often held by many as prerequisites to creativity. The idea is that 'genuine' (and general) creativity cannot be described in terms of a computable (and hence fundamentally mathematical or algorithmic) process. In the specific case of chess problems there has never, to our knowledge, been an approach to automatic composition that relied on a 'creative process' that was independent of human involvement and that could also, in principle, be applied to other areas. Section 2.4 of (Iqbal, 2008) provides a more detailed review of the main approaches in this regard that have been used in the past.

In the following section, we nevertheless present our novel DSNS approach to computational creativity as applied to chess problems. We have intentionally left out the numerous other techniques in AI and computational creativity to compare against to keep the length of this paper manageable and more importantly, because they are fundamentally different to the one we propose even though they may be applicable to the same problem discussed. Ours, however, not only works given the aforementioned problem of quality automatic chess problem composition but is also scalable, in principle, to other domains. We therefore present our research in the practical spirit of fundamental science in that if something can be shown to work – even in a limited domain initially – *why* it works is of secondary importance. In fact, we readily concede there are open questions of this nature which are briefly discussed in section 6.

---

[8] This is what makes them *fundamental*, i.e. they work to a reasonable extent and have had demonstrable applications beyond a single domain or task. Note that being *task*-specific is even more constrained than domain-specific. There is also the 'no free lunch theorem'.

# 3 METHODOLOGY

## 3.1 The DSNS Approach Explained

The Digital Synaptic Neural Substrate (DSNS) approach[9] was intended to provide AI researchers with a generic method to combine information, or rather data, from different domains (e.g. chess, music, paintings) such that they could be integrated in such a way as to lead to the automatic creation of new, original objects from any one of those domains. A sort of catalyst for the 'spark' of creativity, if you will. An analogy may be the creation of a new painting by a human after having been exposed to or 'inspired' by a musical piece or a different painting *and* a musical piece. It would be difficult for the human painter to say for certain which aspects of each lead to which aspects of the new painting (or how) except that these two objects 'come to mind' when asked about the 'inspiration' behind the new painting. It could also be that only one such object is mentioned while the other(s) remain buried deep in memory or the subconscious. We could not find any significant and descriptive neuroscientific or neurological bases for the creation of creative thought so researchers in other areas should be free to posit their own ideas for testing. There is also no 'requirement' for a functionally creative computational process to be grounded in the method employed by the *human* brain, whatever it may be.

Typically, it should be self-evident that the human mind or rather brain must contain bits and pieces of information from various objects we have perceived through our experiences (in the form of generic neuro-chemical substances) and is able to integrate them through volition in poorly-understood ways in order to (sometimes) create new objects of creative value.[10] This fundamental basis of the 'elements of creativity' is therefore not mere conjecture. It is virtually unheard of that a person should be creative in mathematics knowing absolutely nothing about it. Some basic pieces of knowledge about mathematics – along with pieces of all sorts of other information the person has perceived and retained – must first reside in the brain. The precise method through which these pieces of information are intermingled – in particular with regard to producing creative objects or ideas – is far from understood so we freely posit our own ideas to be tested. With this concept of creative information processing in mind, the 'DSNS' terminology and approach was developed.

First, we need two collections, sets or samples of objects from the same or different domains. For example, 100 chess problems and 100 paintings (two domains); or 200 of either divided into two subsamples (one domain). Each object, depending on the domain it is from, is described using a set of attributes. A chess problem, for example, may be described using the attributes: *the number of white pieces on the board*, *the Shannon value of those white pieces* and *the difference in value between the white and black pieces*. These are things that could be perceived or found out about the objects in question. There is theoretically no limit to the number of attributes that can be used to describe an object. However, it should be numerically representable. A painting, on the other hand, may be described using the attributes: *the number of pixels in the image*, *the number of colors used* and *the year it was painted*. It is acknowledged that these attributes may not be related in any meaningful way; this is intentional to reflect the influx of various types of conscious and subconscious information into the brain.

In selecting attributes, one might prefer unique attributes in the domain that do not 'overlap' much. For example, *the number of pixels in the image* should not be included along with say, *the number of 16x16 blocks of pixels in the image* because these are simply variations of the same concept. *The number of white pieces on the board* should not be included along with *the number of white pawns on the board* because the latter is a subset of the former. There are no fixed rules with regard to attribute selection (which is perhaps a strength when it comes to creativity) and we hesitate to state any but a little common sense and good judgement can go a long way. The number of attributes in a domain therefore also becomes more manageable. It is presumed that we have at least some basic knowledge about a domain sufficient for us to identify at least some of the attributes that may partially describe it. So it is fair to say that if we have absolutely no knowledge about a domain, we cannot identify and represent any of its attributes, much less in the form of numbers.

Second, the attribute values for each object should be tabled as a single row along with its object identifier. This row and its values can be identified as a DSNS *string*. An example of three chess problem DSNS strings is provided in Table 1. The Forsyth-Edwards Notation (FEN) is a representation of the starting position of the chess problem. As an identifier or 'key' for the object, the FEN is typically included along with the attributes and their values to make up the DSNS string for a chess problem. So in Table 1, each row, including the FEN, is a separate DSNS string.

---

[9] Iqbal, M. A. M. (2014). A Process of Integrating Information from Different Domains for the Purpose of Generating Novel Creative Objects. Malaysia Patent Application No.: PI 2014703983. Filing Date: 24 December 2014.

[10] At present, this is experimentally somewhat beyond our present efforts with, for example, fMRI and brain hemodynamics to identify regions of the brain related to particular tasks or activities (Ruiz et al., 2014; Strang et al., 2014).

| FEN | White Pieces | Black Pieces | Value of White Pieces | Value of Black Pieces | Value Difference |
|---|---|---|---|---|---|
| 8/1p2BN1K/4Qp2/n1R4p/3k2P1/P5n1/4P3/1r6 w - - 0 1 | 8 | 7 | 23 | 14 | 9 |
| 5rk1/5qpn/8/3N4/3B4/1B6/1KP3R1/8 w - - 0 1 | 6 | 5 | 15 | 18 | 3 |
| 5Q2/b2k1P2/1n1NNn2/1P1p4/6P1/8/8/7K w - - 0 1 | 7 | 5 | 18 | 10 | 8 |

**Table 1**: Example DSNS strings for three chess problems.

An example of DSNS strings for three paintings is shown in Table 2. Here, a unique number serves as the identifier for each painting. The number of attributes for an object, especially between different domains, need not be fixed. In the example case presented, the chess problem sample and the painting sample each contain three individual objects with five attributes each. However, the DSNS approach also allows for an unequal number of attributes between the two domains being used.

| Painting ID | Pixel Count | Colors | Identifiable Objects | Year | Aspect Ratio |
|---|---|---|---|---|---|
| 115 | 398321 | 81935 | 4 | 1626 | 0.729 |
| 16 | 282576 | 68624 | 3 | 1513 | 0.762 |
| 175 | 430407 | 41510 | 2 | 1917 | 0.611 |

**Table 2**: Examples DSNS strings for three paintings.

Third, a random DSNS string is selected from each sample in order to generate a ‘deviation’ value. The idea behind the deviation value is that it can be thought to act as a measure of ‘creative difference’ between two objects. Table 3 shows two DSNS strings (from the same domain, for illustrative purposes) and their attributes. The two columns to the right, i.e. |d| and (∑÷) represent ‘absolute difference’ and ‘summative division’, respectively. The former is simply the absolute difference between the two values of a particular attribute given the two strings whereas the latter is a new concept; it represents the sum of the division of the first string’s attribute value with the second string’s attribute value, and vice-versa; e.g. for attribute 1 in Table 3, it is 6/7 + 7/6. What we call ‘summative division’ was used because we wanted something simple yet functional enough to create sufficient variation in the deviation values that was also not, to our knowledge, previously described in the mathematical literature.

| Attribute | String 1 | String 2 | \|d\| | (∑÷) | Deviation |
|---|---|---|---|---|---|
| 1 | 6 | 7 | 1 | 2.024 | (7+6.275) – 6 = 7.275 |
| 2 | 5 | 7 | 2 | 2.114 | |
| 3 | 13 | 9 | 4 | 2.137 | |
| | | | **7** | **6.275** | |

**Table 3**: Absolute difference, summative division and deviation.

The deviation for the two DSNS strings is therefore = [∑|d| + ∑ (∑÷)] - CID. The CID or ‘creative indifference value’ (‘6’ in this case) is obtained from the sum of the absolute differences and summative divisions (i.e. [∑|d| + ∑ (∑÷)]) for two DSNS strings with *exactly the same attribute values*; a sort of baseline. In theory, since the attribute values are exactly the same, the objects should also be (creatively) the same. Imagine if, in Table 3, for attribute 1, strings 1 and 2 both had a value of 6, the absolute difference would be ‘0’ and the summative division would be (6/6 + 6/6 = 2). This would be true for all the attributes in the strings. The sum of 0+2 = 2; and the number of attributes, *n*, would lead to a CID of 2*n*.

Fourth, this deviation of 7.275 (see Table 3) is then used to generate two *new* DSNS strings by randomly selecting possible attribute values from the original sample (so the values are ‘realistic’) until a pair is found that matches that deviation (i.e. 7.275) using the same process just described above.[11] So suppose the sample has 300 DSNS strings,[12]

[11] Note that the deviation value can be rounded to one or two decimal places to improve performance. The requirement of a pair of objects is fundamental and built into the DSNS approach. A ‘randomization’ approach on the attribute values of just one DSNS string would, in principle, not provide any basis of comparison with another creative object (that would have its own DSNS string), and as is shown in section 4.4, would likely not

there are 300 possible values for each attribute, or somewhat less if the values for each attribute are not unique. A random change of all these attribute values, in new combinations, might lead to two *new* DSNS strings such that the same, precise deviation value results.[13] This can be called 'stage 1' of the search for a perfect match. The idea put forth here is to obtain two new strings which have a similar contrast (i.e. difference) in 'creative value' as the original pair. If after 30 seconds (the amount of time is flexible but should be reasonably long), no exact match can be found using the 'realistic' attribute values in the sample, then the search is 'widened' to include *all* possible values (i.e. to one decimal place or more, depending on the nature of the attribute values and time constraints) between the lowest and highest attribute values in the sample. This is 'stage 2'. So if for a particular attribute, there were only five values in the sample: 3, 7, 9, 10, 12, now *any* value between 3-12 (i.e. including 4, 5, 6, 8 and 11) would be sought after. This goes on for another 30 seconds. If still no match is found, then the final 'stage 3' takes effect where the two new DSNS stings that produced the closest deviation to the one desired are taken. 'Contemplation time' is therefore considered a factor in the creative process.

So in the case of the 7.275 deviation, after 60 seconds, taking into account all the strings generated in stages 1 and 2 (which means they should be stored in an historical array of some sort), if the closest deviation produced was 7.15, then those two DSNS strings that produced it would be retained. These two new strings can then be used to generate one or more new objects based on the same set of attributes (but with potentially different values) used to describe the other objects. The generation tool or technique is a separate issue beyond the DSNS and domain specific; in the case of *this* research, the chess problem generation process is explained in section 4.2 and Appendix A.

In summary, the first two DSNS strings selected represent two unique objects within the domain (e.g. two chess problems). Their 'creative difference' is represented by the deviation value generated. This deviation value is then used to produce two new strings (which themselves would generate the same or similar deviation value) but with different *attribute* values. These new attribute values can ultimately be used to generate new objects of potentially creative value using whatever generation system is available, i.e. by feeding it the attribute values it needs to build objects of a particular type. In very simple terms, it is like having or being able to derive new (realistic) measurements of height, width and length to build a new box. The box-building system is not part of the DSNS approach itself.

## 3.2 Using Different Domains

If objects from two different domains should be integrated using the DSNS approach, the process is the same except that *two* deviations are generated. Each deviation is obtained by choosing two random DSNS strings from the same domain, following the steps described in the previous section. These two deviations are then 'merged'. So if the first domain was chess problems and the second music, and the first deviation obtained was 126.21 whereas the second was 35722.11 (the numbers can be quite different), the two deviations can be merged as shown in Figure 1. If an object from the chess problem domain was desired, then the deviation value would need to relate to that domain, i.e. be the sort of number that a deviation derived entirely from objects in that domain would look like. However, with two deviations and only one of them relating to the chess problem domain, they need to be merged as shown on the left side of Figure 1. This may be seen as a sort of 'lateral thinking' approach to merging the two unrelated deviation values so they become functionally one in the desired or target domain.

**Chess Problem Domain Desired**

| | | | | | | | |
|---|---|---|---|---|---|---|---|
| | | 1 | 2 | 6 | . | 2 | 1 |
| ~~3~~ | ~~5~~ | 7 | 2 | 2 | . | 1 | 1 |
| | | 1 | 2 | 2 | . | 1 | 1 |

**Music Domain Desired**

| | | | | | | | |
|---|---|---|---|---|---|---|---|
| 3 | 5 | 7 | 2 | 2 | . | 1 | 1 |
| | | 1 | 2 | 6 | . | 2 | 1 |
| 3 | 5 | 1 | 2 | 2 | . | 1 | 1 |

**Figure 1**: Merging deviations from different domains.

What happens is the chess problem deviation (i.e. 126.21) is placed in the first row with the music deviation (i.e. 35722.11) directly beneath it, aligning the digits so they match. The '3' and '5' are simply crossed out because there are no digits of that value in the row above. The remaining digits follow this rule: *If [top digit] modulo [bottom digit]*

---

work as well. Indeed, there are likely hundreds of different ways the DSNS approach could be executed differently but we had to decide on the one we expected to work best.

[12] The sample of 300, in this case, is used simply because it was the source of the original two DSNS strings and readily available. A slightly larger or even smaller 'knowledge base' (e.g. 290, 310) to look through can also be used at this point but it would likely not have a significant impact on the result. On the other hand, a much larger knowledge base of say, 600 or even 1,000 objects (even from an altogether different sample than the two strings came from) might yield better results.

[13] The approach we suggest is to randomly select new values for *all* the attributes in each tentatively new string at the same time before testing to see if the new strings produce the desired deviation. Systematically changing the value of just one attribute in either string to see if it brings you closer to the desired deviation is also possible, as are presumably other methods.

*= 0 then result = [bottom digit] else result = [top digit]*. So the result here is 122.11. On the right side of Figure 1, a deviation relating to the music domain is desired so that deviation is put in the first row with the chess problem domain deviation in the second. The '3' and '5' in this case simply 'fall down' and the remaining digits follow the same rule mentioned above. The result in both cases (i.e. the last row) is a *single* deviation value that can be used as described in section 3.1 (the fourth step) to generate two new DSNS strings. In some cases there may be no difference at all between the merged deviation and the original deviation from the same domain. This process of merging may reflect or mimic the imperfect recollection of information by humans which inherently produces variation that can be functional in the creative thought process. In other words, they recall information that is similar and functional in the domain but not necessarily identical to what they have seen before. We consider this to be self-evident as well.

By the same token, *more* than two domains can, in principle, be merged. The chess problem and music domain can be merged to produce a deviation of say, the chess problem type. This deviation can then be merged with a deviation from the painting domain to produce a deviation of the painting type. So data from the chess problem, music and painting domains can be combined to produce content in the painting domain. Theoretically, there is therefore no limit to the number of domains that can be merged in this way.

### 3.3 Using Strings of Different Length

In some cases, a sample of DSNS strings may contain certain objects or 'individuals' that lack a particular attribute value. For instance, the 'year' attribute may not have been available for a particular object such as a painting. In such cases, it is still usable as a DSNS string along with other strings that actually have the year attribute value. The one that does not is simply given the value 'NULL'. During the deviation-generation process (e.g. a string with 5 attributes and one with 4 attributes + 1 NULL attribute happens to be randomly selected) the one with 5 is reduced to 4 attributes by removing the corresponding NULL one. So we are left with an even 4 to 4 DSNS string length in that particular case. This is different from having an attribute value of '0', which is used like any other value. In general, a DSNS string of a different length implies that it either has fewer or more attributes.

This is typically an issue only for strings from the same domain because as explained in section 3.2, using different domains implies generating two deviations (four DSNS strings involved) and only then merging the deviations. It would not be advisable trying to process two DSNS strings from the same domain but with entirely different attribute types. They should match so that the attribute values carry *some* meaning, even if occasionally the pair needs to default to using one attribute fewer due to a NULL entry in either one of them.

### 3.4 Assumptions and Implications

This DSNS approach is assumed to be able to generate attributes that can then be used to generate creative content. All or some of the attributes may be used for this purpose, depending on the domain and external generation system available. Essentially, we randomly select two objects from a domain, 'reduce' them to DSNS strings and then use those strings to generate a deviation value that represents the 'creative difference' between those two objects. From this deviation value, two *new* DSNS strings are generated (with a similar 'creative difference'). Each or even both of these can then be used to build or create one or more new objects that potentially have creative value. How these objects are created is a separate issue and dependent upon the domain in question and the generation or 'building' system available. Figure 2 shows a general diagram of the DSNS approach.

It is therefore implied that raw data 'extracted' from an object – even from different domains – can be integrated such that new objects of creative value from either of the original domains, can be generated. This is comparable to how a human being perceives, say, a painting, which is likely represented in memory as a neuro-chemical substance or reaction in the brain, and then a piece of music, which is also represented in the 'similar format' neuro-chemical substance or reaction in the brain (analogous to the DSNS strings) so that they can be integrated (analogous to the deviation value) to produce either a new painting or new piece of music (analogous to the building specifications supplied by the *new* DSNS strings). It stands to reason that if the deviation value (i.e. the 'creative difference') between two DSNS strings derived from two objects can be matched by two new strings with realistic attribute values other than what was contained in the original DSNS strings, then these two new strings – in particular the attribute values they contain – can be used to generate or create new objects that would resemble the objects from which the original DSNS strings were derived, i.e. in terms of creative value.

The *contrast* between the two DSNS strings which is represented by the deviation value is not necessarily a measure of how weak and strong, creatively, the two objects in question are. It could just as well be a measure of how different two equally strong (or equally weak) objects are in terms of creativity. Chances are, however, one is better or more interesting than the other, however slightly, even. This is also true for the two new strings from which two or more new objects can potentially be created.

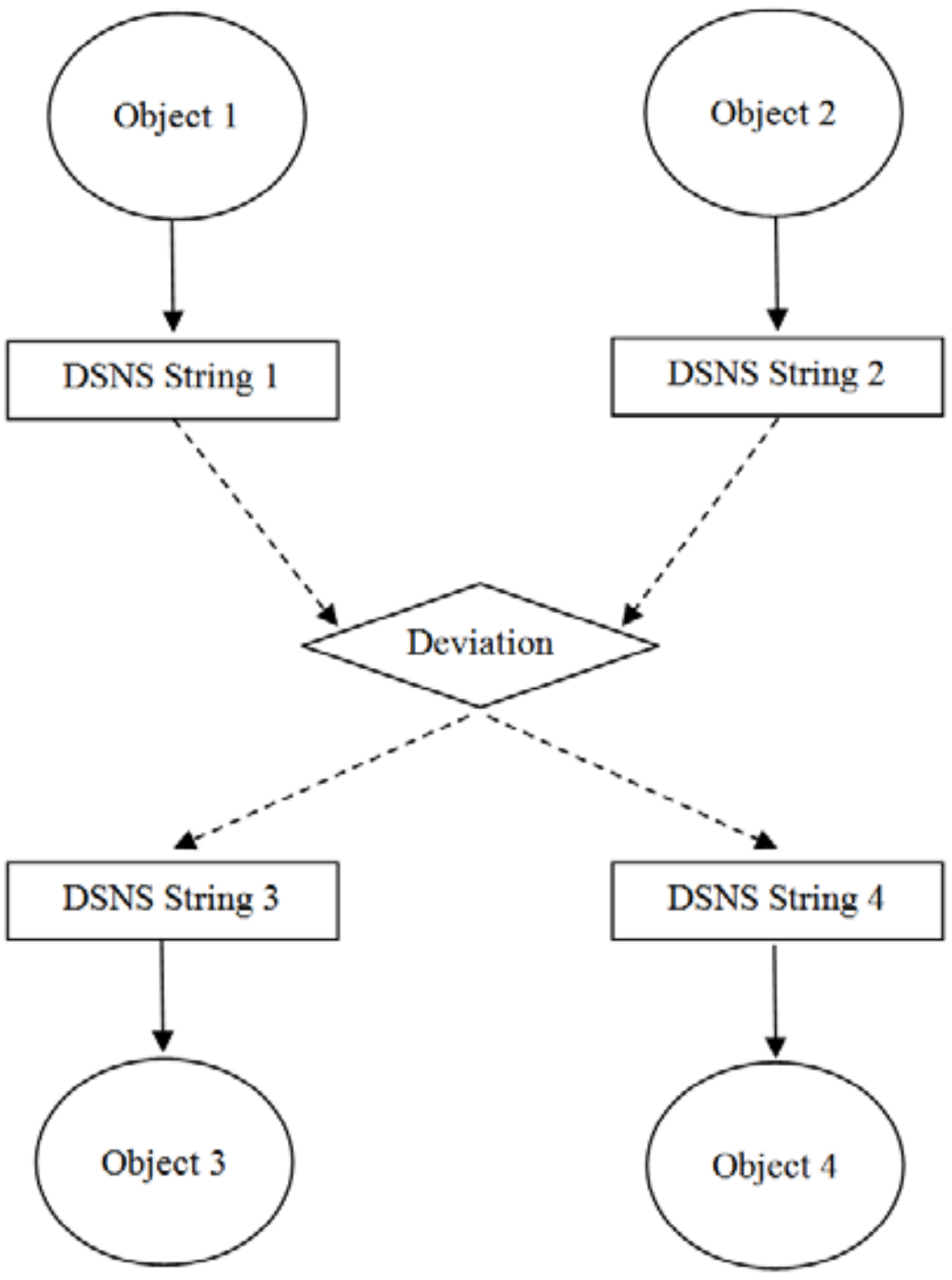

**Figure 2**: General diagram of the DSNS approach.

## 4 EXPERIMENTAL WORK AND DISCUSSION

### 4.1 The Domain of Investigation

In order to test if the DSNS approach actually worked, we had to first select a domain of investigation that required creativity (Hesse, 2011). Given our prior experience and background, the domain of chess problems or compositions was selected. In particular, three-move, forced-mate problems (i.e. against any defense) otherwise known as 'three-movers' (#3, for short) since it is one of the most common and illustrative. Chess problems are basically like puzzles with a stipulation (e.g. *White to play and checkmate in three moves*) and the solver is challenged to discover the winning sequence of moves. In most cases, the solution is somewhat unexpected which is why problems are often considered works of art and beautiful or aesthetic in nature. Not all 'working' or valid problems, however, are equally aesthetic. For instance, the final forced three-move mate sequence taken from a tournament game between two expert players might *resemble* a composed chess problem in terms of the basic requirements but it lacks the forethought and 'design' that a human composer puts into a chess problem intended for publication in a magazine or book.

Figures 3a and 3b show a typical three-mover and three-move checkmate sequence taken from an actual game between two expert players, respectively. The two are 'physically' similar in the sense that we have a starting position (in the case of the real game it is taken just prior to move 37) and the existence of a forced mate-in-three sequence. They differ in the sense that the composition (3a) was designed to be economical and attractive whereas the real game occurred over the board in a tournament. Composition convention dictates that compositions should be designed such that they *could* have occurred in a real game (e.g. you cannot have a black bishop on the **a8** square *and* a black pawn on the **b7** square) and *should* appear realistic (e.g. you should not have 6 queens and 4 rooks of the same color).

Compositions also tend to feature chess themes or motifs such as the *grimshaw* or *plachutta* with varying degrees of sophistication in their solutions, i.e. the main line of play and possible variations (Velimirovic and Valtonen, 2013). These can be planned during the design phase of a chess problem or they could simply emerge in a tournament game winning sequence. A good (as in of publishable quality) composition may therefore take an expert human composer hours or even weeks to compose; perhaps even longer. For these reasons, compositions are considered, on average, to be more attractive or aesthetic than sequences taken from real games (even between experts) where 'design' factors cannot be controlled and the players are furthermore typically under time constraints to win by any legal means possible, rather than to win artfully. These issues are explored in greater detail in (Iqbal, 2008), for the interested reader.

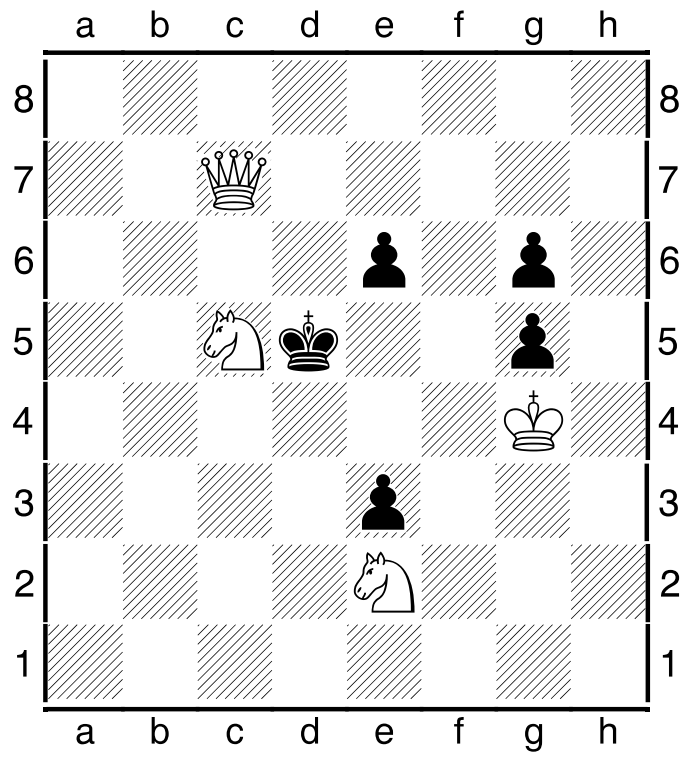

Vincente Maria Norberto Portil
*The Westminster Papers*
1872
1. Ne4 Kxe4 2. Qd8 e5 3. Ng3# 1-0
**(a)**

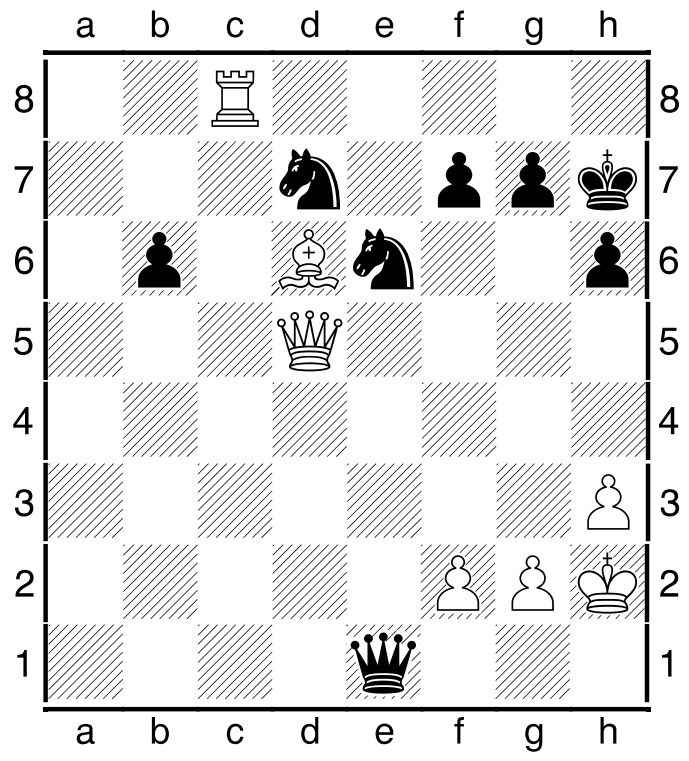

Wells, P. K. vs. Baburin, A.
*BCF-chT 9900 (4NCL)*
1999
37. Qf5+ g6 38. Qxf7+ Ng7 39. Qg8# 1-0
**(b)**

**Figure 3**: A typical #3 chess problem and forced three-move mate sequence from a real game.

4.2 The Experimental Tool

We used an experimentally-validated computational chess aesthetics model to evaluate the chess problems and tournament game sequences mentioned in the previous section.[14] This model was incorporated into a computer program called CHESTHETICA (Iqbal et al., 2012). The present version is 9.22. Generally, the model uses formalizations of a selection of aesthetic principles and themes derived from the chess literature, in conjunction with stochastic technology (i.e. some randomness) to evaluate beauty in #3 sequences. We found it advisable to use the formalization or 'equations' approach to aesthetics (McCarthy, 2005) even though other approaches (e.g. an artificial neural network) may be effective as well; at present, however, these remain untested. Assessments based on the model – typically in the form a single digit number precise to three decimal places – correlate positively and well with domain-competent human aesthetic evaluation (Iqbal et al., 2012).

A notable characteristic of the program is that it may produce a slightly different aesthetics score the second or third time it evaluates the same move sequence. However, this is generally inconsequential so only one cycle of evaluation was used per chess problem, i.e. not an average of 10 or 20 cycles per problem which would have been a waste of time and resources. We have not found a way to apply the model in its entirety as an heuristic which can be used in any sort of known hill-climbing search because the dynamics of the chessboard are not really predictable; which is what makes it such a creative endeavor and full of paradoxes. As an analogy, the ability to assess the beauty of an object (what the model does) does not automatically render that ability capable of producing such objects or improving upon them continuosly in a systematic way. We will not go into further detail about the model here or try to revalidate it in contrast to other generic ideas on aesthetics (Ritchie, 2007). Interested or skeptical readers are invited to review the aforementioned reference for a complete explanation.

A downloadable version of CHESTHETICA ENDGAME (CEG) which supports both #3 and endgame studies is also available (Iqbal, 2012). Endgame studies are typically a longer, more sophisticated type of chess problem (Nunn, 2011). The reason we used this computational aesthetics model to perform the aesthetic evaluations of the three-movers is simply because it would be too much to ask of human experts to assess hundreds of problems and sequences. Doing so would also likely result in less consistent results due to fatigue and human error. CHESTHETICA v9.22 was modified and also used to automatically compose three-movers based on the DSNS approach. It is therefore the 'generation tool' in this particular case that relies on the new DSNS string attribute values, as alluded to in section 3.1. The newer composing module is independent of the program's aesthetics evaluation components mentioned earlier. The program's earlier composing module included a

[14] Good references with regard to the field of computational aesthetics can be found in (Iqbal, 2015) and (Galanter, 2012).

‘random’ and ‘experience table’ approach not particularly relevant to this research (Iqbal, 2011). Details about how the existing composing algorithm was modified to accommodate the DSNS approach is provided in Appendix A.

The program is capable of imposing a ‘filter’ of up to 5 common composition conventions, namely, *no cooked problems*, *no duals*, *no checks in the key (i.e. first) move*, *no captures in the key move* and *no key move that restricts the enemy king’s movement*. Only the third one was applied for all experiments to keep the composing rate reasonably high and consistent for experimental purposes. The composing rate of CHESTHETICA drops exponentially the more filters are applied. This means it would take very much longer to get the same number of compositions. In general, a composition adhering to only one convention is considered of lower quality than one adhering to say, five. However, this is not necessarily true in every case and there is some evidence to suggest that composition conventions may not be as important to non-esoteric aesthetics as composers tend to think they are (Iqbal, 2014a).

### 4.3 The First Experiment: Same Domain

The first experiment was intended to test the hypothesis that the DSNS approach – using one domain as a source – can be used to generate objects of creative value in the same domain, i.e. forced three-movers (compositions and from tournament games). The term ‘creative value’ here implies *aesthetic* value. Aesthetics was chosen as a measure of creativity because it stands to reason that beautiful man-made objects are typically the product of creative minds. We also intended to use the experimental tool, CHESTHETICA (see section 4.2) which is the only known experimentally-validated chess aesthetics evaluation software in the world. The Meson Chess Problem Database[15] was used as a source of compositions by (largely) experienced human composers. From a total of 29,453 three-movers, 300 assessed to have an aesthetics score above 3.5 (by CHESTHETICA) were used.[16] Another 300 assessed to have a score below 1.25 were also used. The two sets represent both high and low quality samples in terms of creative content. Each of these sets were randomly divided into two subsets of 150 compositions each, for a total of four sets or subsamples.

Forced three-move checkmate sequences taken from tournament games were also used. These were taken from the ChessBase Big Database 2011. From a total of 5,063,778 games, 300 games between players with an Elo rating above 2,500 were used. Another 300 between players with an Elo rating below 1,500 were used. These two sets also represent high and low creative content, respectively. Each of these sets were further divided into two subsets of 150 compositions each, for a total of four sets or subsamples. It stands to reason that with higher quality source materials, the DSNS approach should produce chess compositions of higher quality, on average, than with lower quality materials. For the compositions, the two subsets of 150 compositions were used to generate the deviation values (see section 3.1) and produce new pairs of DSNS strings for use by CHESTHETICA in composing new chess problems (see Appendix A). The same was done for the tournament game subsets. For brevity, these sets will be referred to as Comp3.5, Comp1.25, TG2500 and TG1500.

The ten attributes used in the DSNS process for this domain represent some feature of the chess problem or tournament game sequence that a human observer might notice or be able to find out. They are, in principle, arbitrary, but should be obtainable by some means and describable using real numbers. These include:

1. The number of white pieces in the initial position
2. The number of black pieces in the initial position
3. The Shannon value of those white pieces
4. The Shannon value of those black pieces
5. The difference between the two Shannon values (i.e. material difference)
6. The number of moves in the sequence (fixed at 3)
7. The year the chess composition was composed/tournament game was played
8. The first piece to move in the sequence (P = 1, N = 2, B = 3, R = 4, Q = 5, K = 6)
9. The last piece to move in the sequence (P = 1, N = 2, B = 3, R = 4, Q = 5, K = 6)
10. The sparsity value of the initial position

[15] http://www.bstephen.me.uk

[16] Aesthetics scores by the program typically range between 0 and 5 with hardly any at either extreme (none known to date, in fact). It is also difficult to determine what the highest possible aesthetics score for a three-move sequence could be given the dynamics and virtually infinite possibilities on the chessboard.

The idea here is that these 'scraps' of information are similar to how the brain stores pieces of information from objects we observe. It is clear that few, if any, of us can remember objects in precise detail. More likely is that we remember particular attributes of an object that appeal or have some significance to us. Taken collectively, these features represent that object in our brains. The first two attributes are clear. The third is the total value of the chess pieces as described by Claude Shannon in his seminal paper on computer chess (Shannon, 1950). He attributed relative values to the chess pieces as follows: queen = 9, rook = 5, bishop/knight = 3 and pawn = 1.[17] The king is of infinite value because losing it means losing the game. With approximate values such as this, computers are able to gain a good idea of the material imbalances on the board and play a decent game of chess. Modern chess programs may use slightly different weights and hundreds of other game-playing heuristics but these values still serve well as a rough guide with regard to which army is doing better; sufficient for the purposes of our experiments. The fifth attribute is clear.

The sixth attribute is the same for all the chess problems, i.e. '3'. In future implementations of the DSNS where chess problems of different lengths may be used, this attribute will vary, of course. The fact that this one attribute has the same value for all compositions and sequences actually makes the DSNS experiment more realistic because it is unlikely that objects from any domain would have attributes that differed in *every* regard. Paintings may have the same frame size or weight whereas musical tracks may all have the same frequency. All the attributes values taken collectively, however (as is the case in the DSNS approach) still provide for sufficient variation in the new DSNS strings to be generated. This is also not equivalent to 'artifically' leaving the sixth attribute out and using only nine. The seventh attribute is more subtle and not obvious from the moves itself. The sequence could have been composed or played yesterday or 500 years ago. This piece of information is nevertheless something that a user may care to find out by checking the game details available in most chess databases or by doing an online search.

The eighth and ninth attributes are integer codes that represent the first and last pieces to move in the sequence. So if a bishop was the first piece to move in the sequence, the value for attribute 8 would be '3' and if a knight was played in the last move of the sequence, the value of attribute 9 would be '2'. The tenth and final attribute used refers to an approximation of how spread out the pieces are on the board in the initial position. It is described using the following equation. *s()* denotes the number of pieces in the field of a particular piece, $p_n$, i.e. in the squares immediately around it. A detailed explanation of the sparsity concept is available in (Iqbal and Yaacob, 2008).

$$sparsity = \left[\left(n^{-1}.\sum_{1}^{n} s\left(p_n\right)\right)+1\right]^{-1}$$

A screen capture showing an excerpt of the spreadsheet with DSNS strings on the left and right (150 on each side) and the deviation value in the middle is available in Appendix B. A 'random' composing approach was also used as a control. Essentially, this approach does not use any 'technology' in composing chess problems and places the pieces on the board purely at random. Details are available in (Iqbal, 2011). CHESTHETICA was allowed to automatically compose using this DSNS approach (and the random one) for a total of 24 hours, i.e. 12 hours on one machine and another 12 hours on another. The first computer (PC 1) was a notebook: Intel(R) Core(TM) i7-3820QM CPU @ 2.70 GHz with 16 GB of RAM running Microsoft Windows 7 Pro SP1 64-bit. The second computer (PC 2) was a desktop: Intel(R) Core(TM) Duo CPU E8400 @ 3.00 GHz with 4 GB of RAM running Microsoft Windows 7 Pro SP1 32-bit. Table 4 shows the 'real world' performance of CHESTHETICA in generating compositions using the DSNS approach.

| | **Sources for DSNS Approach** | | | | **Random Approach** |
|---|---|---|---|---|---|
| | **Comp3.5** | **Comp1.25** | **TG2500** | **TG1500** | |
| **PC 1** | 30 | 106 | 39 | 78 | 110 |
| **PC 2** | 41 | 108 | 17 | 37 | 106 |
| **Total** | 71 | 214 | 56 | 115 | 226 |

**Table 4**: Compositions generated by CHESTHETICA using the DSNS approach (one domain) in 24 hours total.

The average composing rate (using the DSNS approach) for PC 1 was 5.27 compositions per hour (*cph*) whereas for PC 2 it was 4.23 *cph*. The sample sizes are too small, at this point, for statistical analysis but it is fair to assume that processing power and memory do not make very much of a difference in this case. Table 5 shows the mean aesthetics scores and variations for the (total) generated compositions derived from each set. The higher the number of variations, the 'richer' or 'more complex' the composition is typically regarded to be, according to convention. Despite the precision of these aesthetics scores they are typically used for *ranking* purposes. One should therefore not make claims that an aesthetics score of 2.0 implies that a composition is *twice* as beautiful as one with an aesthetics score of 1.0. At the same time, a small yet statistically significant difference in mean aesthetics (between groups) is not necessarily meaningless. A single factor analysis of variance (ANOVA) test was performed across all the five sets comparing the aesthetics means and the differences were found to be statistically significant: F (4, 667) = 4.33, p = 0.0018. Even excluding the 'random approach' control, the differences in means were still statistically significant: F (3, 452) = 3.14, p = 0.0251.

[17] Alan Turing proposed slightly different values, i.e. P=1, N = 3, B = 3.5, R = 5, Q = 10 (Turing, 1953).

| | **Sources for DSNS Approach** | | | | **Random Approach** |
|---|---|---|---|---|---|
| | **Comp3.5** | **Comp1.25** | **TG2500** | **TG1500** | |
| **Aesthetics Score** | 2.438 | 2.278 | 2.308 | 2.278 | 2.208 |
| **Variations** | 49.9 | 31.9 | 30.0 | 22.0 | 35.5 |

**Table 5**: Mean aesthetics scores and variations for the compositions generated (one domain).

So we see a clear distinction in mean aesthetic value between compositions generated from higher quality sources (Comp3.5, TG2500) and lower quality sources (Comp1.25, TG1500) using the DSNS approach. Interestingly, compositions generated using higher quality tournament games fared slightly better, aesthetically, than low quality compositions. The random approach, expectedly, fared the poorest aesthetically. There was no statistically significant difference in the mean variations across all the five sets. This supports previous work that suggested variations are relevant but only to a point (Iqbal et al., 2012). In other words, simply having *more* variations does not necessarily imply higher aesthetic quality.

### 4.4 The Second Experiment: Different Domains

The first experiment (see section 4.3) demonstrated that the DSNS approach can indeed produce compositions of higher creative value (than a purely random approach) based on the sources used for the DSNS strings. In the second experiment, we tested to see if materials sourced from *different* domains could also be used successfully. The four domains included: renowned human artworks (i.e. paintings), computer-generated abstract art pieces (from the Elvira system), photographs (with people in them, but not selfies) and renowned classical music pieces. The paintings, photographs and music were identified and selected (300 objects each) by a female research assistant[18] who is also a co-author. The abstract art pieces (1,000 in total) were supplied by another two co-authors and from these, the first 300 based on the file names were used. These pieces were created using the Elvira system (Colton et al., 2011). The selection process of these objects were considered sufficiently random for our purposes.

The paintings and classical music pieces were considered high quality sources in contrast to the abstract art pieces that were computer-evolved and considered of comparatively lower quality. The photographs were considered to be of 'moderate' quality; perhaps somewhere between the computer-evolved abstract art pieces and the renowned paintings. Each of these four sets of 300 objects were divided into subsets of 150 as in the first experiment. The same DSNS approach was then used; refer to section 3.2 for information about merging deviations from different domains. The attributes used for the paintings, photographs and abstract art pieces are as follows. As with the chess domain they represent unique features that a human observer might notice or be able to find out. Similarly they are arbitrary but should be obtainable by some means and describable using real numbers.

1. Resolution (i.e. the total number of pixels)
2. Number of colors used in the image
3. Number of distinct objects in the image
4. The year it was created
5. Aspect ratio (i.e. width divided by height)
6. Brightness
7. Contrast
8. Noisiness
9. Lightness
10. File size (in bits)

[18] We specify the gender of the selector here in case it is ever found to have influenced the results.

The third attribute was determined through ‘manual’ and direct observation of each image by our female research assistant. These included, for example, people, tables and vases; so two people, one table and three vases would equal six distinct objects. ‘Brightness’ is a reference to how much all the pixels go from black (dark) to white (bright) (Bezraydin et al., 2007). The ‘contrast’ value was calculated as explained in (Koren, 2006). ‘Noisiness’ relates to how random or unrelated, pixels are with other pixels surrounding them (Kerr, 2008). ‘Lightness’ as in HSL (or Hue, Saturation, Lightness) was calculated as explained in (Yu et al., 2006). Much of this attribute information was obtained through programmatic means by our other (male) research assistant who is also a co-author. As with the previous experiment, attributes that can be used to describe the object are arbitrary as long as they can be represented using real numbers. Figures 4, 5 and 6 show examples of the renowned paintings, computer-evolved abstract art pieces and photographs of people used, respectively.

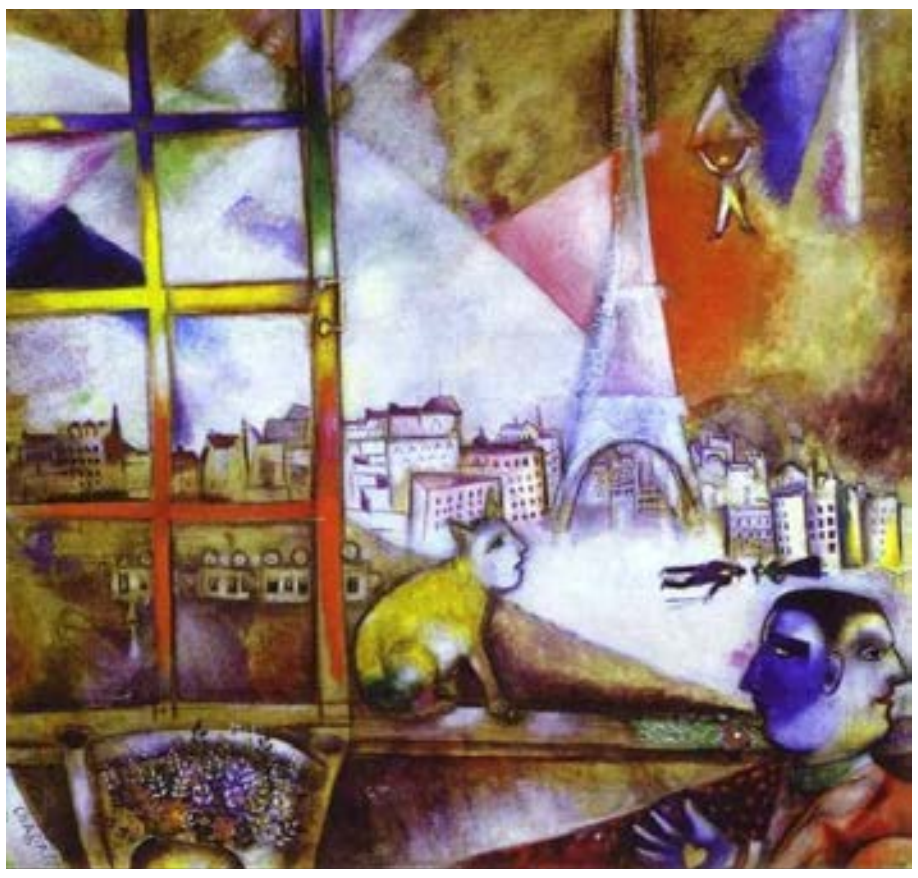

*Paris Through the Window* by Marc Chagall
1913
**(a)**

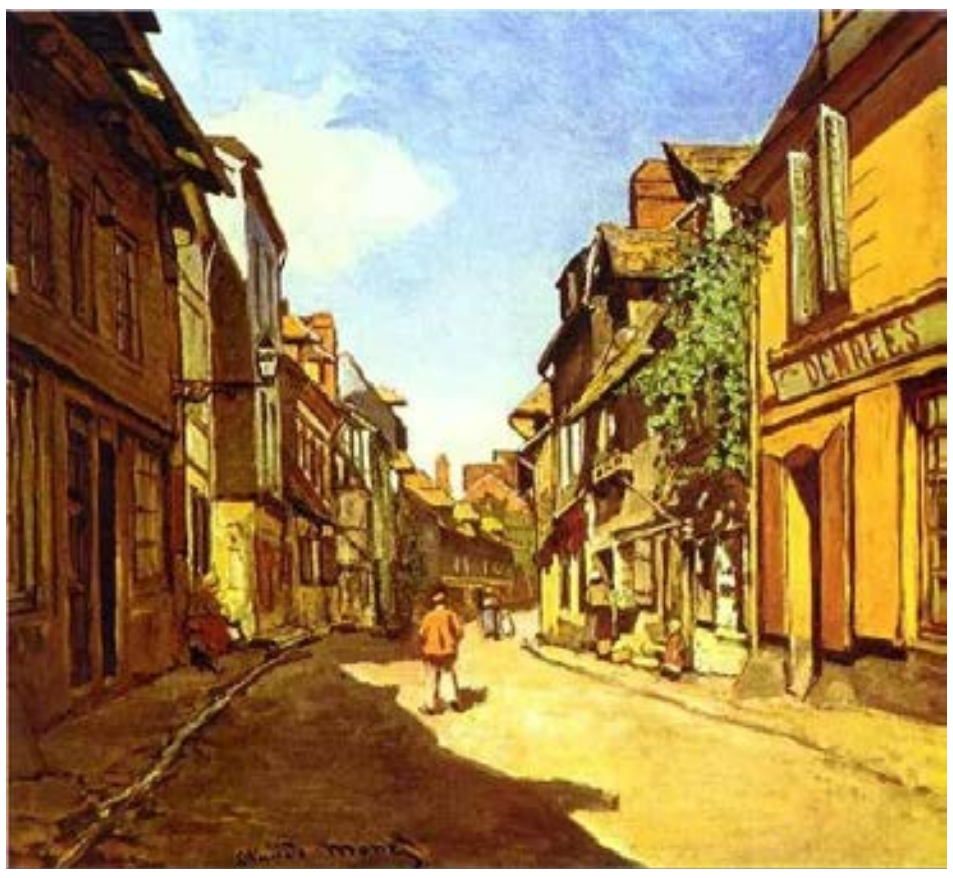

*Rue de la Bavolle, Honfleur* by Claude Monet
1864
**(b)**

**Figure 4**: Examples of the renowned paintings used.

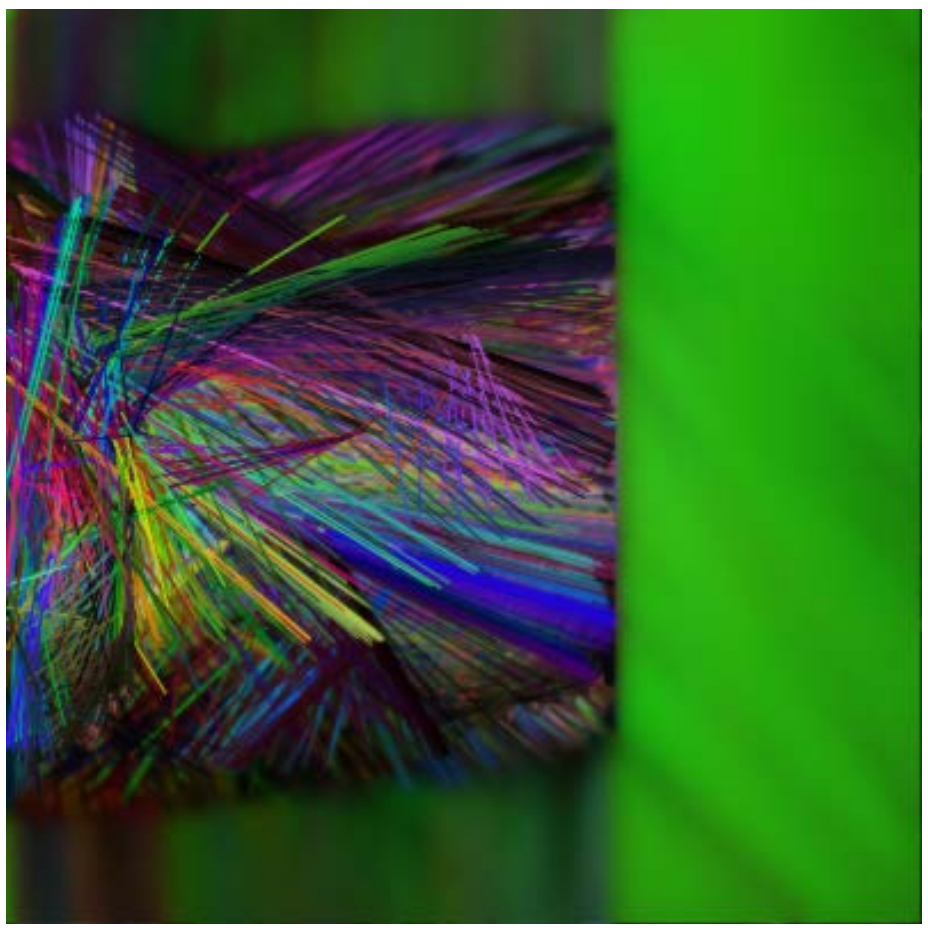

*Byline* by the Elvira system, 2010
**(a)**

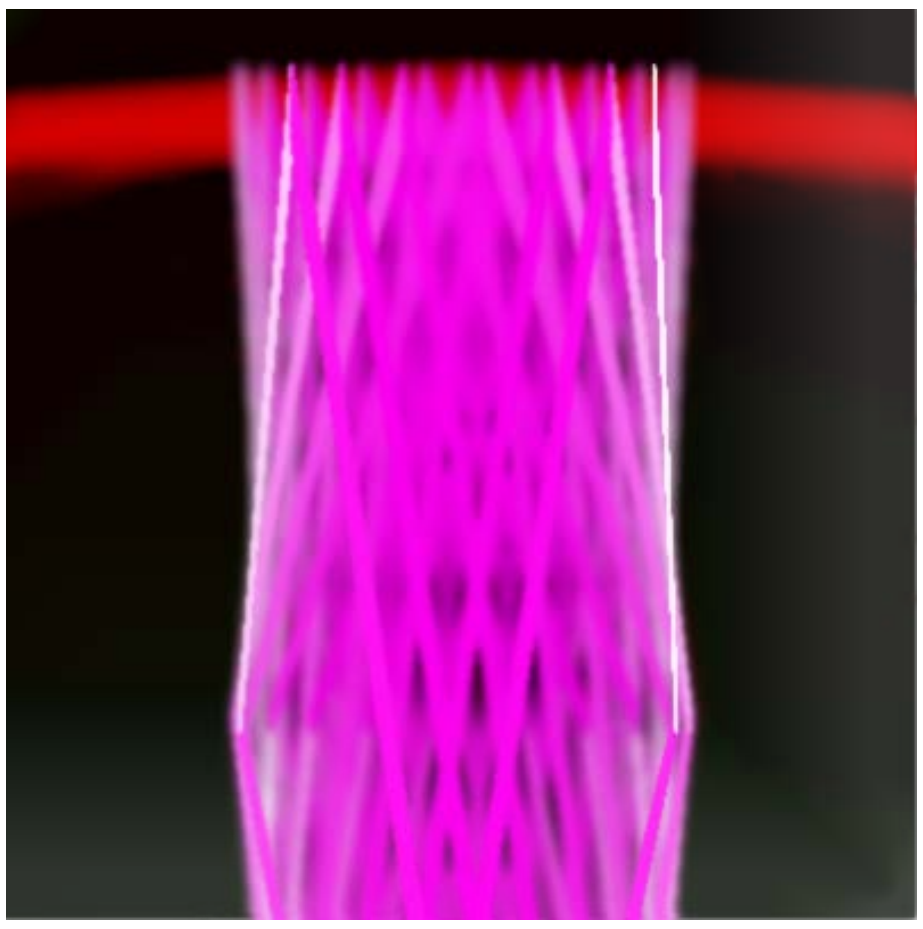

*Crown* by the Elvira system, 2010
**(b)**

**Figure 5**: Examples of the computer-evolved abstract art pieces used.

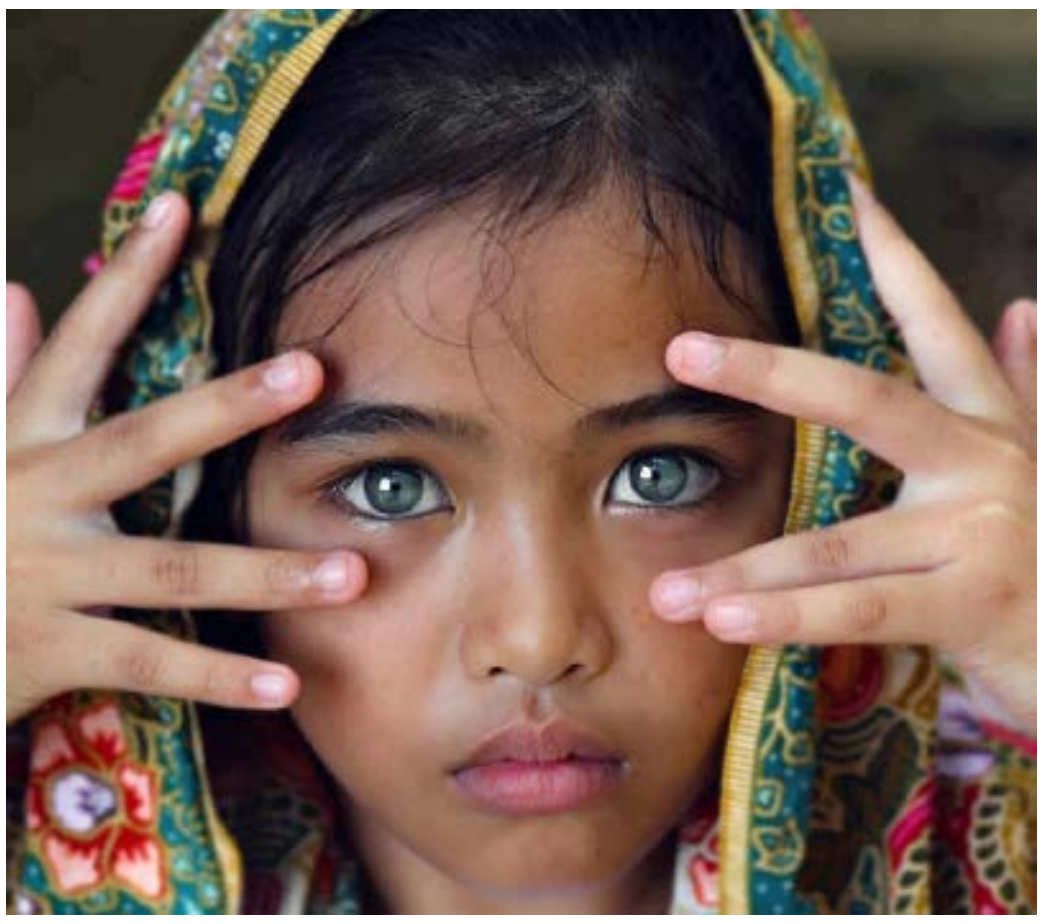

*Cute, beautiful and lovely kids* on Pinterest
by Patricia St Louis
http://www.pinterest.com/pin/177188566561408314
2013
**(a)**

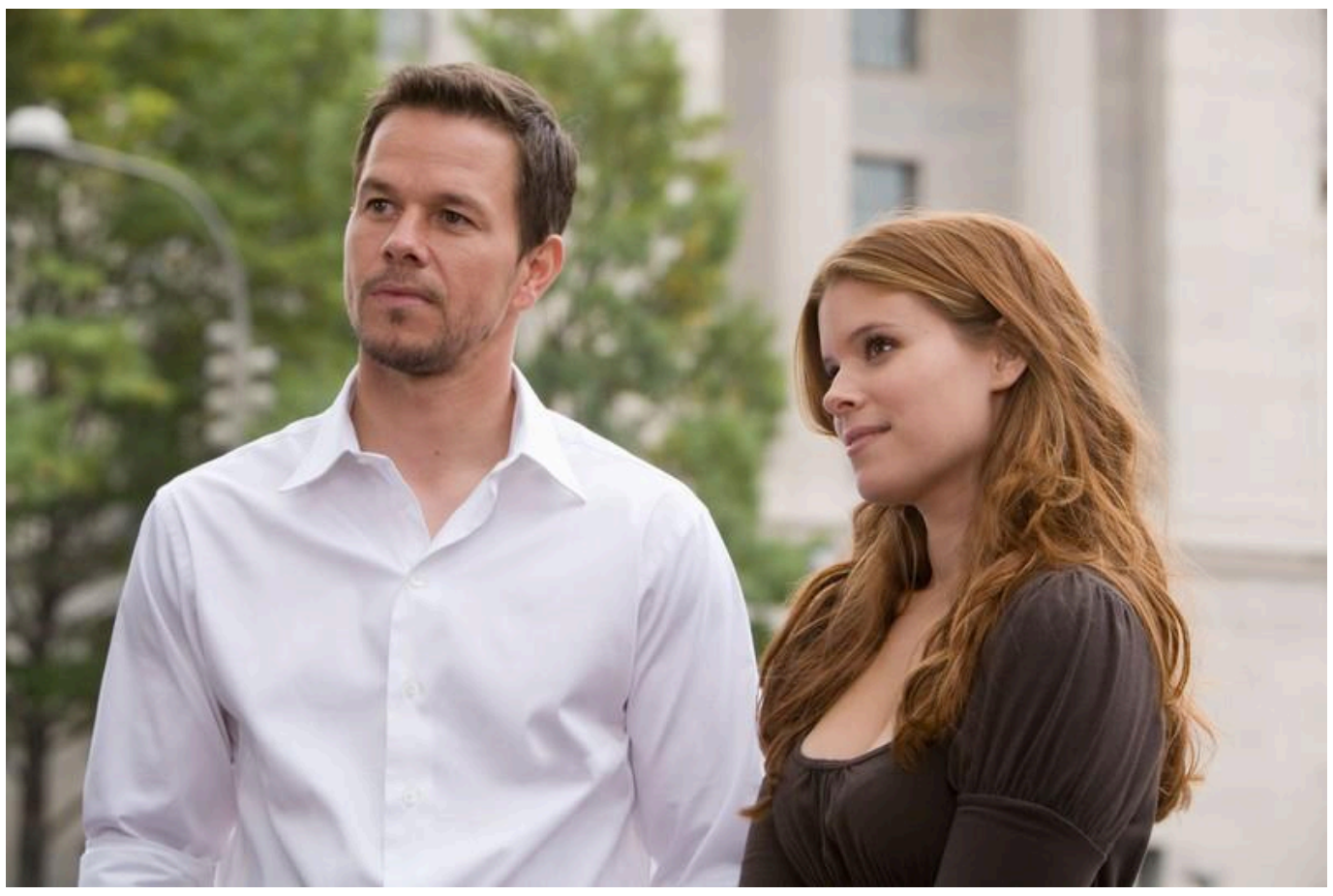

*Mark Wahlberg and Kate Mara*
Shooter (motion picture), Paramount Pictures
http://www.fdb.cz/film/odstrelovac-shooter/fotogalerie/43638
2007
**(b)**

**Figure 6**: Examples of the photographs of people used.

The attributes used for the classical music pieces comprised of the following.

1. Number of channels
2. Sample rate / frequency
3. File size (in bytes)
4. The year it was composed
5. Average loudness
6. Sound energy
7. Dynamic range
8. Bitrate
9. Sample size / quantization
10. Sound efficiency

'Average loudness' refers to the average peak to peak amplitude as calculated using the Python programming language *Average (audio file)* function. 'Sound energy' refers to the zero-crossing rate that the sound goes through, i.e. the amplitude going to zero. The greater the zero-crossing rate, the more energy it produces (Srinivasan et al., 1999). The 'dynamic range' is basically the noise floor subtracted from the peak signal. The 'bitrate' is the number of bits of data used per second (e.g. 128 kbps) in the audio file. The 'sample size' refers to the quantization; typically 8 or 16 bits; 24 bits is also possible but less common. The 'sound efficiency' refers to the effectiveness of the sound signal measured by its crest factor, i.e. a measure of the impulsiveness of a noise or vibration signal (Norton and Karczub, 2003). Examples of the classical music pieces used include: *Handel – Sarabande*, *Shostakovich – Romance* and *Mozart – Lacrimosa*.

Once again, for brevity, the chess source sets will be referred to as Comp3.5, Comp1.25, TG2500 and TG1500. The other domain sets with which the chess sets are merged with will be referred to as *art* (renowned paintings), *Elvira* (evolved abstract art pieces), *photo* (photographs of people) and *music* (renowned classical music pieces). In Table 6, for each pairing, the topmost row represents the number of compositions generated by CHESTHETICA for PC1, the middle row PC2 and the bottom row, in bold, is the total. So, for example, the Comp3.5 domain merged with the Elvira domain yielded 84 generated compositions in total.

In principle, PC 1 is more powerful than PC 2 in terms of the processing power and available memory. In the previous experiment, there were too few samples to present any statistical analysis of the results. However, combining the information related to the DSNS approach in Table 4 with the information in Table 6, we have a sample size of 20 each for PC 1 and PC 2, which is not unreasonable for statistical purposes. A two-sample F-test for variances was first applied to the samples to determine whether a two-sample, two-tailed, T-test assuming equal or unequal variances (i.e. TTEV or TTUV) should be used to compare the means (at the 5% significance level), i.e. the average number of compositions generated in 12 hours on a particular machine. This approach will henceforth be used where the means of two samples need to be compared.

| | Comp3.5 | Comp1.25 | TG2500 | TG1500 |
|---|---|---|---|---|
| | 37 | 75 | 75 | 65 |
| **+Art** | 38 | 94 | 94 | 50 |
| | **75** | **169** | **169** | **115** |
| | 46 | 128 | 128 | 51 |
| **+Elvira** | 38 | 90 | 90 | 44 |
| | **84** | **218** | **218** | **95** |
| | 38 | 94 | 94 | 36 |
| **+Photo** | 32 | 88 | 88 | 33 |
| | **70** | **182** | **182** | **69** |
| | 35 | 81 | 81 | 58 |
| **+Music** | 42 | 98 | 98 | 57 |
| | **77** | **179** | **179** | **115** |

**Table 6**: Compositions generated by CHESTHETICA using the DSNS approach (two domains) in 24 hours total.

It turns out that, based on a TTEV test, there was no statistically significant difference between the average number of compositions composed on PC 1 (worked out to a rate of 5.729 *cph*) and PC 2 (a rate of 5.321 *cph*). This finding is important because some readers may assume all of this is simply brute force calculation related directly to how fast a computer can process data and how much memory it has to work with. No doubt, these *are* factors but it does not necessarily indicate the output potential of the DSNS composing approach. A useful analogy may be assuming that simply turning up the heat in a room will dry wet clothes proportionately faster. Table 7 shows, for each pairing, the mean aesthetics scores (top row) and mean variations (bottom row).

| | Comp3.5 | Comp1.25 | TG2500 | TG1500 |
|---|---|---|---|---|
| **+Art** | 2.243 | 2.289 | 2.250 | 2.363 |
| | 63 | 33.6 | 28 | 19.8 |
| **+Elvira** | 2.277 | 2.328 | 2.352 | 2.330 |
| | 41.2 | 43.2 | 58.3 | 24.6 |
| **+Photo** | 2.323 | 2.304 | 2.137 | 2.449 |
| | 35.9 | 33.6 | 21.5 | 44.2 |
| **+Music** | 2.270 | 2.302 | 2.364 | 2.400 |
| | 63 | 30.2 | 60.8 | 31.2 |

**Table 7**: Mean aesthetics scores and variations for the compositions generated (two domains).

For Comp3.5, an ANOVA test across 5 sets, i.e. Comp3.5 on its own (2.438 mean aesthetics score, see Table 5) and Comp3.5 merged with the four other domains (see Table 7), showed a statistically significant difference between the means for their aesthetics scores: F (4, 372) = 2.69, p = 0.0031. This suggests that, given a high quality source (Comp3.5), using 'information' from other domains to compose chess problems does not result in higher quality compositions since the aesthetics scores using two domains were all lower. There was no difference of statistical significance for the average number of variations between the sets, which is the same outcome as the previous experiment.

For Comp1.25, a similar ANOVA test showed no statistically significant difference between the means for their aesthetics scores or the average number of variations between the sets. This suggests that given a *lower* quality chess source (2.278 mean aesthetics score for Comp1.25 alone, see Table 5), using information from other domains neither negatively nor positively affects the quality of the generated compositions. For TG2500, however, the same ANOVA test showed a statistically significant difference between the means for their aesthetics scores: F (4, 327) = 3.48, p = 0.0084, but not for the variations. This suggests that using a high quality source of *tournament game sequences* (2.308 mean aesthetics score for TG2500 alone, see Table 5), the quality of the generated compositions can actually be improved slightly if used in combination with the Elvira or classical music sets. Yet the opposite is true if combined with the art and photo domains in this case.

Finally, for TG1500, the ANOVA test showed a statistically significant difference between the means for their aesthetics scores: F (4, 504) = 2.39, p = 0.0497; and *also* for the variations: F (4, 504) = 2.51, p = 0.0412. This suggests that using a *low* quality set of tournament game sequences (2.278 mean aesthetics score for TG1500 alone, see Table 5), the quality of the generated compositions can actually be improved if used in combination with *any* of the four other domains. In contrast with the result obtained for Comp1.25 above, the result also suggests that there is something 'intrinsically' different about compositions and tournament game sequences given that, in this case, their mean aesthetics scores using a single domain source were exactly the same (i.e. 2.278). If, in principle, we can accept that compositions, even low quality ones, are generally better than low quality tournament game sequences, then the results so far would suggest that the *lowest* quality source domain (i.e. TG1500) used in combination with other domains actually produces among the best results.

Even though the difference in the average number of variations in the case of TG1500 was also statistically significant, it is difficult to prove that having more or fewer variations (and what the thresholds really are) improves the quality of a composition. We can only say that the DSNS approach, in combining information from other domains with low quality tournament game sequences, is able to influence the average number of variations for the compositions generated both positively and negatively. Returning to the mean aesthetics scores, it is not clear why, exactly, the lowest quality chess source used in combination with *all* the other completely unrelated domains tested, through the DSNS approach, should produce compositions of higher quality than using the low quality chess source alone. It suggests that 'unrelated' information from different domains can indeed be aggregated meaningfully as is likely occurring in some form in the human brain.

To test this further, we performed another comparison of three sets involving the TG1500 and photo domains. This is because the 'TG1500 + photo' combination (TG1500p for brevity) yielded the highest, statistically significant mean aesthetics score (2.449). We introduced a new source material set, i.e. 'photorandom'. This is basically the same set of 300 photographs of people except that the values for each attribute were randomly-generated within the range of highest and lowest for that attribute. So, for instance, if the 'number of colors' attribute for all 300 photos ranged from 927 to 238,034, then a 'random' photo would have a new value for that attribute based on a random number generated within that range. This was done for all 10 attributes. The result was a collection of 'garbage' photos where the attributes of each made no sense and probably did not exist in reality.

It stands to reason that, if information is indeed being 'successfully' integrated or aggregated between unrelated domain types using the DSNS approach, *actual* photos (with *real* attribute values) should do better than those with randomly-generated attribute values (garbage photos) and also better than *not* using photos at all. So we compared the TG1500 (alone with no photo integration), the TG1500p *and* the TG1500 + photorandom (TG1500pr for brevity) sets together. The results are shown in Table 8.

| | **TG1500** | **TG1500p** | **TG1500pr** |
|---|---|---|---|
| **Aesthetics Score** | 2.278 | 2.449 | 2.325 |
| **Variations** | 22.0 | 44.2 | 19.6 |

**Table 8**: Mean aesthetics scores and variations for the compositions generated (one and two domains).

An ANOVA test showed a statistically significant difference between the means for their aesthetics scores: $F(2, 286) = 3.62$, $p = 0.0281$; and also for the variations: $F(2, 286) = 3.60$, $p = 0.0286$. So it would seem that using actual photos (with real attributes) performs better than garbage photo data and also better than not using photos at all. Using actual photos also yields the highest number of variations, on average, of the three sets but that alone is not necessarily an improvement in terms of quality because we can see that the non-DSNS random approach in Table 5 also had a relatively high number of variations (35.5) but the lowest mean aesthetics score (2.208) compared to the other single domain chess sources. In summary, the DSNS approach, at least in the case of low quality tournament game sequences integrating information from real photographs, is able to produce compositions of higher aesthetic quality than otherwise. It remains an open question for now why this should happen at all, even though the human brain likely successfully mingles information from different domains in unusual and poorly-understood ways as well.

## 4.5 The Third Experiment: Variations in the Number of Objects and Attributes

In the third experiment, we wanted to test if the number of *objects* used in the DSNS approach and the number of *attributes* influenced the quality of the compositions generated. For this purpose, we used the Comp3.5 and TG1500 samples contrasted against variations of each where only 150 objects were used (instead of 300) and also against the same sample size but where only 5 attributes were used, i.e. the first 5 attributes (see section 4.3). The automatic composing process (see Appendix A) would simply skip the constraints that depended upon the missing attributes. Comp3.5 and TG1500 were used because these had, respectively, the highest and lowest aesthetics scores (see Table 5). The results are shown in Tables 9 and 10.

| | **Comp3.5 300 objects 10 attributes** | **Comp3.5 150 objects 10 attributes** | **Comp3.5 300 objects 5 attributes** |
|---|---|---|---|
| **Aesthetics Score** | 2.438 | 2.263 | 2.272 |
| **Variations** | 49.9 | 50.6 | 26.3 |

**Table 9**: Mean aesthetics scores and variations for the compositions generated (one high quality source, variable objects and attributes).

| | **TG1500 300 objects 10 attributes** | **TG1500 150 objects 10 attributes** | **TG1500 300 objects 5 attributes** |
|---|---|---|---|
| **Aesthetics Score** | 2.278 | 2.257 | 2.335 |
| **Variations** | 22.0 | 22.1 | 29.3 |

**Table 10**: Mean aesthetics scores and variations for the compositions generated (one low quality source, variable objects and attributes).

For Comp3.5 (300 objects, 10 attributes) against Comp3.5 (150 objects, 10 attributes), the difference between the mean aesthetics scores based on a TTEV test was statistically significant: t(148) = 2.439, P = 0.0159. For Comp3.5 (300 objects, 10 attributes) against Comp3.5 (300 objects, 5 attributes), the difference between the mean aesthetics scores based on the same test was statistically significant as well: t(169) = 2.749, P = 0.0033. This suggests that, given a high quality chess source, using more objects or using more attributes improves the quality of the generated compositions, which implies scalability of the DSNS approach. There were no differences of statistical significance (TTUV) between the average number of variations comparing the same groups.

For TG1500 (300 objects, 10 attributes) against TG1500 (150 objects, 10 attributes), the difference between the mean aesthetics scores based on a TTEV test was not statistically significant. For TG1500 (300 objects, 10 attributes) against TG1500 (300 objects, 5 attributes), the difference between the mean aesthetics scores based on a TTUV test was not statistically significant either. This suggests that using more objects or more attributes makes no difference given a low quality chess source. There were no differences of statistical significance (TTUV) between the average number of variations comparing the same groups. We also tested a merged set, i.e. TG1500p (see previous section) using just 150 objects and 5 attributes as follows; for the photos, we used the first 5 attributes (see section 4.3).

| | **TG1500p 300 games 300 photos 10 attributes** | **TG1500p 150 games 150 photos 10 attributes** | **TG1500p 150 games 300 photos 10 attributes** | **TG1500p 300 games 150 photos 10 attributes** |
|---|---|---|---|---|
| **Aesthetics Score** | 2.449 | 2.171 | 2.363 | 2.306 |
| **Variations** | 44.2 | 24.5 | 39.9 | 35.1 |

**Table 11**: Mean aesthetics scores and variations for the compositions generated (two domains, variable objects).

| | **TG1500p 300 games 300 photos 10 attributes** | **TG1500p 300 games 300 photos 5 attributes** |
|---|---|---|
| **Aesthetics Score** | 2.449 | 2.355 |
| **Variations** | 44.2 | 25 |

**Table 12**: Mean aesthetics scores and variations for the compositions generated (two domains, variable attributes).

For TG1500p (300 games, 300 photos, 10 attributes) against TG1500p (150 games, 150 photos, 10 attributes), the difference between the mean aesthetics scores based on a TTEV test was statistically significant: t(153) = 3.779, P = 0.0001. For TG1500p (300 games, 300 photos, 10 attributes) against TG1500p (150 games, 300 photos, 10 attributes), the difference between the mean aesthetics scores based on the same test was not statistically significant. For TG1500p (300 games, 300 photos, 10 attributes) against TG1500p (300 games, 150 photos, 10 attributes), the difference between the mean aesthetics scores based on the same test was statistically significant: t(204) = 2.219, P = 0.0276. So using more *objects* for both sources given these two source domains improves the quality of the compositions generated. This is consistent with the previous finding when using only a high quality chess source (Comp3.5, Table 9). However, for TG1500p sources, reducing just the number of games does not make a difference; but reducing the number of *photos*, interestingly, does. There were no differences of statistical significance across these four sets between the average number of variations based on an ANOVA test.

For TG1500p (300 games, 300 photos, 10 attributes) against TG1500p (300 games, 300 photos, 5 attributes), the difference between the mean aesthetics scores based on a TTUV test was not statistically significant. This suggests that, given these two source domains, using more *attributes* makes no difference. This is somewhat consistent with the previous finding when using only a low quality chess source (TG1500, Table 10), in that an increase in the number of *attributes* used does not make a difference, regardless of the inclusion of photos as a source. There were no differences of statistical significance between the average number of variations based on a TTUV test.

## 4.6 The Fourth Experiment: Human Expert Assessment

As explained in section 4.2, enlisting the help of human experts to evaluate aesthetics in this research would have its share of problems. Fatigue, imprecision and inconsistency are hallmarks of the human condition especially when it comes to typically subjective issues such as aesthetics. Nonetheless, if the amount of work required can be kept manageable and interesting for the experts, it is still useful to see what their evaluations might be like. So in the fourth experiment, we used the compositions generated using three approaches, i.e. TG1500p, TG2500 + photo (TG2500p for brevity) and Comp3.5 (see Tables 4 and 6). The following positions (see Table 13) were excluded from the compositions generated in all the sets because they are too simplistic and common to retain the attention of the human experts. K = King, Q = Queen, R = Rook, P = Pawn. A digit before the alphabet indicates the piece count.

| | | |
|---|---|---|
| K, Q vs K<br>K, Q vs K, P<br>K, Q, P vs K<br>K, Q, P vs K, P<br>K, Q, R vs K<br>K, Q, R vs K, P | K, R vs K<br>K, R vs K, P<br>K, R, P vs K<br>K, R, P vs K, P | K, 2R vs K<br>K, 2R vs K, P<br>K, 2R, P vs K<br>K, 2R, P vs K, P |

**Table 13**: Composition types that were excluded for the benefit of the human experts.

From the remainder, the top 30 from each set based on their aesthetics scores were chosen, and these randomly mixed together into a PGN (Portable Game Notation) file or database of 90 compositions. These DSNS-generated compositions and their solutions are provided in Appendix C. All identifying information was also removed from each composition in the file so the human experts would not know which set they came from, or even if they were composed by a human or computer. For a balanced view of aesthetics in the game, the human experts chosen were co-authors Matej Guid (FIDE[19] Master), Jana Krivec (Woman Grandmaster) and Vlaicu Crisan (International Master of Chess Solving and FIDE Master of Composition). Together they represent more or less the spectrum of expertise pertaining to the game of chess and where aesthetics in the game has been recognized, i.e. over the board and also in the domain of composing chess problems. They were considered sufficiently ‘domain competent’ with regard to the game. All the experts were completely unaware of the details of this fourth experiment until it was completed.

The human experts were asked to rate the 90 compositions on a scale of 0.0 to 5.0 (one precision point) “*based on their individual expertise and perception of beauty/aesthetics in the game*”. This would also mean that some compositions would have to have the same rating based on human perception, which is acceptable. A larger scale of say, 0 to 10 was not used because this has already been attempted in previous research work related to the aesthetics model (Iqbal, 2008; Iqbal et al., 2012) and we wanted to try something different. Comments for each composition were optional. After rating the 90 compositions, the experts were asked to then choose 15 from the 90 that were “*most likely to have been composed by a human being*”. The relevance of this second part will become evident later. The PGN file and Microsoft Excel evaluation sheet were e-mailed to the experts independently and they were asked to respond within two weeks. One expert took slightly longer due to other commitments. This is understandable given the complexity of having to (comparatively) rate 90 compositions on a discrete scale based on the subjective aspect of beauty. Their ratings are shown in Appendix D. Two of the human experts provided optional commentary about the merits and issues pertaining to each composition and this is provided in Appendix E.

[19] Fédération Internationale des Échecs (World Chess Federation).

In general, the experts were of the opinion that the compositions they were given were of low quality or too easy. This is to be expected given the filter of just one composition convention (see section 4.2) and possibly because the automatic composer 'reduces' each problem to be as economical as possible (see Appendix A). Imposing all five composition conventions using CHESTHETICA would have taken far too long for our experimental purposes. One expert – the master solver and composer – suspected all the compositions were generated by computer but then went on to choose 16 instead of 15 that he thought were composed by a human. The female grandmaster selected 27 instead of just 15 whereas the FIDE master kept to the instructed 15. This alone suggests that the compositions generated, by expert opinion, were more human-like than to be expected; especially given that they were indeed *all* generated by computer using the DSNS approach. In fact, the experts did tend to have nicer things to say about the ones they thought were composed by a human (see Appendix E).

The average expert score for the 90 compositions they evaluated were assessed based on the original three source sets (which the experts had no idea about), i.e. Comp3.5, TG1500p and TG2500p. An ANOVA test showed no statistically significant difference between the means (i.e. 0.847, 0.722 and 0.776, respectively). This was on a scale of 0.0 to 5.0. The result could have been due to the small sample sizes and the overall difficulty of the task in discerning between many compositions that may appear quite similar in quality, if not form as well. There was also no statistically significant (Pearson) correlation between any two of the human expert evaluations. This was not entirely unexpected as humans tend to factor in personal tastes and biases often without realizing it (Iqbal, 2014a).

The same sets were tested using CHESTHETICA and an ANOVA test showed a statistically significant difference between the means (i.e. 2.678, 2.538 and 2.343, respectively): $F(2, 87) = 6.154$, $p = 0.0032$. So the human experts were unable to discern between the groups based on their average aesthetics ratings but the computer program seemed to think that Comp3.5, given this subset of 30 compositions, produced the highest quality chess problems compared to TG1500p and TG2500p, with the former doing better than the latter. This contradicts the means for these groups (i.e. 2.438, 2.449 and 2.137, respectively) shown in Tables 5 and 7 which had larger sample sizes because an ANOVA test showed the differences in means here to be significant as well: $F(2, 214) = 12.358$, $p = 8.3E\text{-}06$. Here, TG1500p is slightly better than Comp3.5, and TG2500p is the worst of the lot.

The deciding factor, we thought, must reside in subtle form somewhere in the 'undecided' human expert aesthetic evaluations. Looking back at them, the suggestive answer was actually in their selections of the compositions *thought to have been composed by a human* (which also tended to receive the most favorable comments). Considering only the compositions where *two* or more experts agreed – classified in terms of the sources used to produce them – the experts' selections tended to agree with CHESTHETICA's assessment just mentioned. See Table 14.

| # | Experts Agreed | Source |
|---|---|---|
| 24 | 2/3 | Comp3.5 |
| 44 | 2/3 | TG1500 + photo |
| 55 | 2/3 | TG2500 + photo |
| 62 | 2/3 | TG1500 + photo |
| 63 | 2/3 | TG1500 + photo |
| 66 | 2/3 | Comp3.5 |
| 81 | 3/3 | Comp3.5 |
| 88 | 2/3 | TG1500 + photo |

**Table 14**: Expert agreement on the computer compositions thought to be human compositions.

The final standings of the compositions thought to have been composed by a human when tallied in terms of the number of experts (in cases where the majority agreed) stood at: TG1500p (8), Comp3.5 (7), TG2500p (2). This correlates perfectly with the order of CHESTHETICA's assessment of the mean aesthetics scores earlier using all available compositions for these sets, i.e. 2.449, 2.438 and 2.137, respectively. There was absolutely no way the experts could have known the compositions they were selecting favorably had come from these particular source sets. The rough probability of any two experts agreeing with each other that any particular composition was composed by a human = $[[(15+16+27)/3]/90]^2 \times 100 = 4.61\%$. The rough probability of all *three* experts agreeing with each other on a particular composition is 0.99%. Figure 7 shows two of the DSNS-generated compositions the majority of experts agreed upon were composed by a human being. The solutions are in Appendix C.

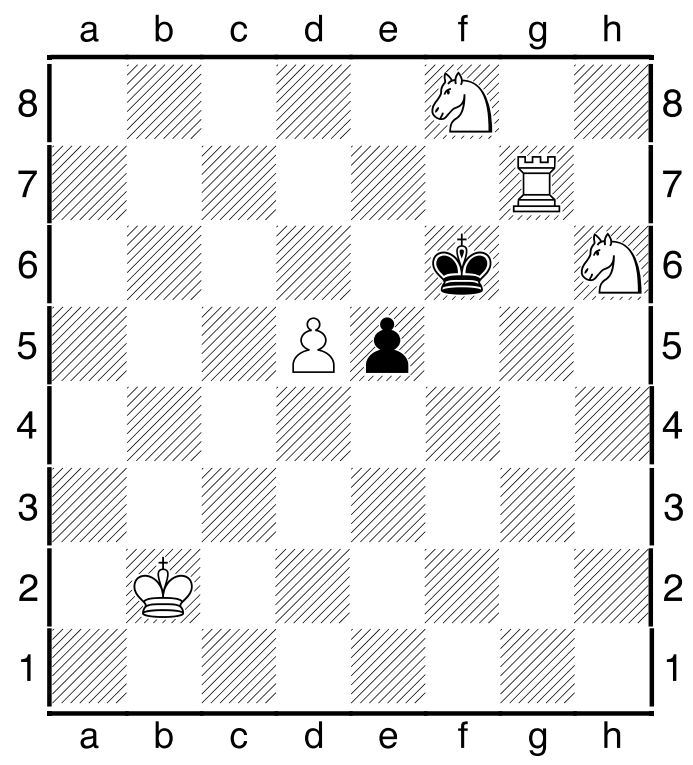

#63
**(a)**

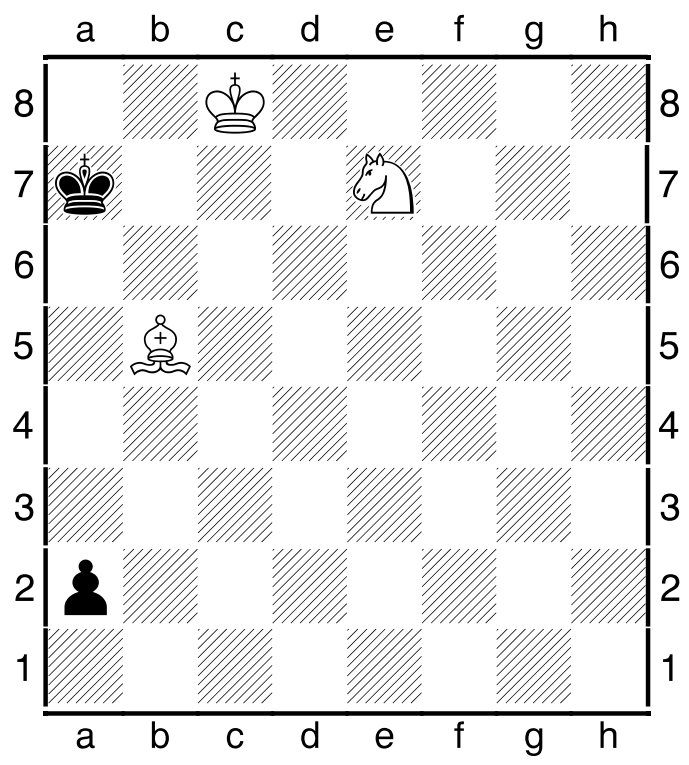

#81
**(b)**

**Figure 7**: Examples of the DSNS-generated #3 compositions.

So it would seem that, based on the available evidence, TG1500p does indeed, on average, produce compositions that are *slightly* better in quality than Comp3.5 alone; and that TG2500p is indeed the worst of the three in terms of average composition quality. This suggests that real (i.e. actual) photographic information, used in conjunction with a low quality chess source (TG1500) via the DSNS approach is able to produce creative objects in the domain of chess of comparable quality or even slightly better than using a high quality chess source alone (Comp3.5). Again, the reason for this remains an open question for now.

### 4.7 The Fifth Experiment: Comparing TG1500p and Comp3.5 Overall Performance

To be sure, an experiment was performed to gauge the overall performance differences between the TG1500p and Comp3.5 approaches. This time, three different computers were used, i.e. PC1 and PC2 (see section 4.3) and PC3, a notebook: Intel(R) Core(TM) i3 CPU M 370 @ 2.40 GHz with 4 GB of RAM running Microsoft Windows 7 Home Premium 64-bit. Each computer had different processing power and capabilities; in general, in reducing amount from PC1 to PC3. CHESTHETICA was set to run for 24 cycles of automatic composition on each computer, comparing these two approaches. Each cycle was set for 6 hours and the composing efficiency in compositions per hour (*cph*) was measured per cycle. The composing process was this time tested using two convention filters (i.e. *no checks in key move* and *no captures in key move*) and also using no convention filters; the former was expected to produce compositions of higher quality but at reduced efficiency.

Compiling the composing efficiency readings from all 72 cycles, we found that the average *cph* using two convention filters based on the TG1500p approach (i.e. 3.32 *cph*) was indeed higher than the Comp3.5 approach (i.e. 2.34 *cph*) to a statistically significant degree based on a TTUV test : $t(108) = -4.406$, $P < 0.0001$. However, using no convention filters the TG1500p approach (i.e. 8.97 *cph*) was indeed lower than the Comp3.5 approach (i.e. 10.58 *cph*) based on a TTEV test: $t(122) = 2.924$, $P = 0.004$. The consolidated aesthetic scores for all of the compositions produced during those 72 cycles showed that there was no statistically significant difference between TG1500p (i.e. 2.360) and Comp3.5 (2.343) using two convention filters but there was when using none, i.e. 2.3 and 2.271, respectively, based on a TTUV test: $t(8415) = -3.262$, $P = 0.0011$. We also confirmed, as predicted, that using two convention filters, the aesthetics scores do, on average, improve compared to using no filters. This was true for both Comp3.5 (two filters: 2.343, none: 2.271) and TG1500p (two filters: 2.36, none: 2.3) based on a TTUV and TTEV test, respectively: $t(1666) = 5.322$, $P < 0.0001$ and $t(5393) = 4.828$, $P < 0.0001$.

So in summary, for more beautiful compositions (i.e. using two convention filters), the TG1500p approach is more efficient than the Comp3.5 approach. It produces more compositions in the same amount of composing time. Comp3.5 produces comparable compositions (in terms of aesthetic quality) but fewer. Using no convention filters, Comp3.5 produces more compositions than TG1500p but of lower aesthetic quality. So the TG1500p is clearly the better of the two approaches. Somehow, combining information from weak games and photographs of people (using the DSNS approach) outperforms using information from high quality chess compositions alone. Interestingly, based on ANOVA tests, there was also no statistically significant difference in average composing efficiency between the three computers used in all cases but one (i.e. TG1500p, two convention filters), where the weakest computer (PC3) actually performed the best: $F(2, 69) = 3.1475$, $p < 0.05$. Perhaps an indicator of a machine employing 'genuine' creativity is a demonstration that raw processing power does not necessarily improve performance; and that perhaps even a relatively low power machine can outperform higher power ones. In principle, creativity largely happens when it happens and can only be improved using a better approach or process rather than a faster machine.

## 4.8 The Sixth Experiment: Comparisons Against the State-of-the-Art

The state-of-the-art with regard to a computational creativity approach and chess problem composition, to the best of our knowledge, is the 'experience table' described in (Iqbal, 2011). Essentially, it uses domain knowledge extracted from a database of 29,453 mostly published chess compositions by human composers, many of whom are of the expert and master level. The experience table is a table of probabilities that a piece of a particular type should exist on a particular square on the board. This information is used in the automatic composing process and was shown to outperform a random-placement approach (similar to the one used in section 4.3) and also experience tables derived from other 'weaker' sources. Full details are in the aforementioned reference.

So in order to compare the DSNS approach with the experience table one, we used three pairings of chess three-movers and photographs for the DSNS, and the original 29,453 source database for the 'experience table' approach. We used three pairings of the DSNS because there are many different ways even the best pairing can be represented, i.e. it is not limited to a particular set or size of chess sequences or photographs. The three pairings therefore provide a more balanced view of the DSNS. The experience table, on the other hand, was shown to perform best using the aforementioned dabatase so we used only that. The three DSNS pairings included the original TG1500p (as explained in section 4.4), and two others, i.e. 4,698 games between weak players with an Elo rating below 1600 in conjunction with 1,000 photographs of people (TG1600p1K for brevity), and TG1500 (see section 4.3) in conjunction with the same 1,000 photographs of people (TG1500p1K for brevity).

We tested using no chess composition conventions, using two conventions (*no checks in key move* and *no captures in key move*) and using three conventions (the previous two and *no cooked problems*). In general, the more conventions applied, the higher the quality of the chess problems (Iqbal, 2014a). So for the tests we used one on the low end (i.e. no conventions) and two tests on the higher end. Table 15 shows the mean aesthetics scores of the compositions generated and the composing efficiency in compositions per hour (cph) based on 24 cycles (as explained in section 4.7). For three conventions, we used 36 cycles because it takes longer to compose a reasonable amount of problems for statistical analysis.

| **Conventions** | **None** | | **Two** | | **Three** | |
|---|---|---|---|---|---|---|
| **Composing Approach** | **DSNS: TG1500p** | **Experience Table** | **DSNS: TG1600p1K** | **Experience Table** | **DSNS: TG1500p1K** | **Experience Table** |
| **Mean Aesthetics Score** | 2.194 | 2.218 | 2.285 | 2.257 | 2.316 | 2.240 |
| **Mean Efficiency (cph)** | 4.709 | 4.119 | 0.799 | 0.993 | 0.555 | 0.667 |

**Table 15**: Comparison of the DSNS against the state-of-the-art 'experience table' approach.

For no conventions, we used the TTEV test (see section 4.4) because the sample sizes were too large but for the rest we used the Mann-Whitney U test (two-tailed, 5% significance level) to compare the means, which is suitable if the sample sizes are smaller, i.e. typically below 200. There was no statistically significant difference between any of the three paired sets (in terms of both aesthetics and efficiency) except in the case of the mean aesthetics scores when applying *three* conventions (U-value = 7382.5, Z-score = -2.1385, $n_1 = 121$, $n_2 = 144$, $P = 0.03236$). This suggests that, under stricter composing conditions, the DSNS approach outperforms the experience table approach by producing compositions that, on average, rank higher aesthetically. Furthermore, the composing rate is still statistically equivalent. Under less strict composing conditions, there is no difference between the two approaches. Even then, it still leaves something to be said about the DSNS considering that the experience table approach uses tens of thousands of published compositions by human composers as a source whereas the DSNS uses sequences from games between weak players and the entirely unrelated domain of photographs of people.

## 4.9 Human vs. Computer Composition

For the interested reader, in terms of mean aesthetics (as assesed by the aesthetics model and perceived by the majority of domain-competent human players), there was no statistically significant difference between the 69 compositions created using the TG1500p DSNS approach (2.449) and 69 randomly-selected compositions by human experts from the The Meson Chess Problem Database (2.300) explained in section 4.3; TTEV test. There was, however, a statistically significant difference between the average number of variations where TG1500p had 44.2 variations on average and the Meson sample 143.1 on average; TTUV test: $t(93) = 2.937$, $P = 0.00418$ . Again, it is important to note that chess problems have various types (e.g. three-movers, endgame studies, constructs) and different requirements which go beyond the common ground of aesthetics assessed by the aesthetics model.

For instance, compositions by humans tend to feature more variations by design and there is insufficient evidence that simply having more variations leads to improved aesthetics. Likewise, a complex yet award-winning problem may be difficult for the *majority* of chess players to perceive as beautiful whereas an unexpected yet direct tactical checkmate (e.g. from a famous tournament game or construct) might be easier to fathom and more appealing to them. So the comparison between the TG1500p and Meson Database samples suggest that, on average, *aesthetically*, at least, the computer-generated compositions are comparable to (but not better than) those created by human experts. A growing collection of computer-generated chess problems composed using the DSNS approach is available online[20] for interested readers. In addition to being appreciated aesthetically, they can also be used by players of all levels as puzzles to train and improve their game. Additionally, between 5 January and 19 February 2015, for instance, 1,263 three-movers were composed in total using anywhere between one and nine different instances of CHESTHETICA and no repetitions were detected.

# 5 CONSOLIDATION OF RESULTS

Bringing all of the experimental data of this research together, we present the following findings regarding the DSNS approach, including some extended findings.

<u>DSNS</u>

1. Computer processing power and memory do not have a significant effect on the composing efficiency.
2. Given the use of just a single domain, the DSNS approach produces higher quality output compared to a completely random approach regardless of the quality of that DSNS source.
3. Given the use of just a single domain, a higher quality source results in higher quality output.
4. Given the use of just a single domain of high quality, combining with another unrelated domain has a negative effect on the quality of the output.
5. Given the use of just a single domain of ‘moderate’ quality (e.g. TG2500) – considering all chess samples used – combining with another unrelated domain improves the quality of the output in some cases but has a negative effect in other cases.
6. Given the use of just a single domain of low quality (preferably *lowest* quality, e.g. TG1500), combining with another unrelated domain improves the quality of the output in *all* cases.
7. Given the use of just a single domain of low quality (i.e. TG1500), combining with *actual* photographs of people produces higher quality output than garbage photo data and not using photos at all.
8. Given the use of just a single domain of high quality, using more objects or more attributes has a positive effect on the quality of the output.
9. Given the use of just a single domain of low quality, using more objects or more attributes has no effect on the quality of the output.
10. For TG1500p (low quality chess domain used in combination with photo domain), increasing the number of objects for both improves the quality of the output. Reducing the number of photos alone has a negative effect. Reducing the number of attributes of both has no effect.
11. Overall, taking into account composing efficiency and quality using different machines, TG1500p outperforms Comp3.5.
12. The DSNS also outperforms the present state-of-the-art composing approach (i.e. the ‘experience table’) in terms of average aesthetic quality of compositions generated under stricter composing conditions such as using three convention filters. Under lesser conditions, it is equal.

[20] Search YouTube for channel “Azlan Iqbal” or go to https://www.youtube.com/c/AzlanIqbal

13. The computer-generated compositions using the TG1500p approach are of comparable aesthetic quality to compositions created by human experts, even though award-winning compositions are also judged by other factors not specifically accounted for by the aesthetics model used.

Extended Findings

1. The number of variations in a composition likely has little to no effect on its quality beyond a limited point.
2. Human expert aesthetic assessment of computer-generated compositions (of not particularly high quality) tends to be inconclusive using a discrete scale. Other methods such as them selecting those which are thought to have been composed by a human tend to reveal more.

These findings can be explored in more detail by the reader by referring back to section 4. Even so, there appears to be a pattern with regard to the DSNS approach, at least when aimed at creating objects of creative value within the domain of chess. It would seem that if the intended domain (i.e. the one in which objects of creative value are to be created) has a high quality source upon which to draw upon (e.g. Comp3.5), then using objects from unrelated domains may be unnecessary. However, a larger number of such objects and a larger number of attributes that describe each of those objects would likely improve the results.

On the other hand, where a high quality source upon which to draw upon is not available – or only a low quality source is available (e.g. TG1500) – then using objects from other domains (e.g. actual photos of people) somehow improves results. Perhaps even to the point that is comparable to using a high quality source alone or even slightly better. The reason for this is not clear. The only explanation that appears to make some sense at this point is that the DSNS approach must be working in a way that is somewhat analogous to the neuro-chemical substances and reactions that represent objects and experiences of all kinds in our brains. When these representations of objects combine and interact in poorly-understood ways, we sometimes are able to produce new objects that are deemed, 'creative'.

## 5.1 Limitations of the DSNS and Its Possible Application in Other Domains

While we have shown experimentally that the DSNS not only works but also outperforms other comparable approaches in automatic chess problem composition (of the three-mover variety, at least), there are certain limitations to consider. For instance, the 'process' of attribute or feature selection in a given domain is generally arbitrary, though they must be numerically representable. This usually goes against our scientific inclinations to be very precise and well-defined in our methods. However, we must realize that there is no known 'formula' for creativity and that different brains tend to perceive even the same things somewhat differently. This may be why some people are more creative than others. So the lack of a well-intended yet constraining process of attribute selection as may be found in other areas of artificial intelligence creates some confusion about 'what works' and what does not.

There is also a question of how many objects in a particular domain should be used. The permutations can be endless and the only way to be sure, at least to a certain extent, is to run experiments as we have done in sections 4.3, 4.4 and 4.5. We now know, for instance, that games between weak players and photographs of people can be used to generate relatively good three-move chess problems. However, the experiments do not explain *why* this is so. Is there a reason? Perhaps. Does it matter to scientists and engineers? Probably not because they can still study and build DSNS systems that work and create both artful and useful things (like chess problems). If we found out *why*, could we improve or optimize the DSNS process? Possibly. So the lack of knowing why can be seen as an additional limitation but perhaps just a temporary one. A lot more work by a lot more people in many different areas (e.g. neuroscience, AI, psychology, philosophy) is clearly required to make significant headway in answering the 'why' question.

Can the DSNS be applied in other areas? In principle, the approach itself does not constrain it to chess. If we have or can imagine an existing system that is able to construct say, paintings, given a set of inputs like the number of colors, the size of the canvas, the frequency of a particular kind of stroke and the intensity of certain colors, the DSNS can indeed supply various combinations of these numbers that might result in interesting paintings. There is no way to be sure unless detailed and rigorous experiments, like the ones in this paper, are done. Perhaps drawings by young children in combination with classical music would produce the best results. So any domain that is able to construct creative objects given a set of inputs, whatever and however many there are (that can be represented using numbers) can indeed be tested for compatibility with the DSNS approach; but this is clearly beyond the scope and budget of the present work.

## 6 CONCLUSIONS

Human creativity has proven to be quite an elusive concept to formalize; more so than 'mere' intelligence. Our creativity is not only a source of enjoyment in terms of being able to create beautiful objects but it is also the source of all our technological advancements such as useful ideas and designs that have materialized, be it vaccines to prevent disease or computerized vehicles to take us into space. Essentially, everything around us of value is the product of human creativity. So the incentive to 'mechanize' this process - perhaps in a way that could far exceed our own abilities – is clear. To date there have been many attempts by computer scientists and artificial intelligence researchers to develop software that can 'come up' with creative products such as music, poetry and paintings. Some more successful than others; yet improvements over time cannot be denied.

However, to our knowledge, there is no single underlying process pertaining to computational creativity that is domain-independent. In this article, we have proposed just such a thing: the digital synaptic neural substrate (DSNS), named for its analogous nature to the generic, neuro-chemical substances that likely represent objects and our experiences in our brains. That generic substance that mingles and interacts in poorly-understood ways so as to sometimes, in some people, create that 'Eureka!' moment. The DSNS has been shown in this research to be effective in the process of composing chess problems of reasonably high quality[21] – though not quite yet on the level of the *best* human composers – using information obtained from the same domain and *also* using information taken from the same domain *and* a completely unrelated domain. The reason for this is still an open question. A significant advantage of the DSNS approach is the lack of the need to program domain-specific, rule-based heuristics for creating creative content. It can be seen as an underlying process or backbone approach that can be 'plugged in' to any creative content generation system or tool by supplying the necessary attributes that system needs to do its work. The DSNS applies the necessary 'constraints' to the system to hone its output.

There are virtually an infinite number of different ways or permutations to further test the DSNS approach. Variations might include in terms of the domain type, domain subtypes, desired output type, length of random search for a matching deviation value, and even combining more than two domains. Not to mention variations in the number of attributes and objects used. Which combinations work best and why? So the DSNS approach appears to be quite scalable and open to many further inquiries for interested researchers. However, these tests are beyond the scope of this already lengthy paper which aims primarily to introduce the DSNS approach and concept for use in the field of computational creativity and perhaps even other fields.

From an artificial intelligence standpoint, one might imagine the DSNS approach implemented in an advanced robot capable of fully exploring its environment, using a variety of senses; perhaps even more and more sensitive than our own. A robot that is able to automatically extract attributes or features from the objects it perceives and 'pool' them together in the form of DSNS strings in its computerized brain. These strings will then be free to mingle until something of value – in any domain the robot is familiar with – can likely be produced. The robot would have a built-in system of trial and error that learns automatically which combinations of attributes from which domains produce the best results and would create a cycle of consistently improving output. This process would be, in principle, inexhaustible because the same creative objects produced can then be fed back into the robot's perception for further processing, assuming an ever-changing environment was not enough. A reality in the not-so-distant future, perhaps. In any case, at the very least, the DSNS approach has been shown to be quite good at composing valid chess problems of the three-move variety (that requires creativity). This is what we set out to do and have hopefully demonstrated in this article.

## ACKNOWLEDGEMENTS

This research was supported by the Ministry of Science, Technology and Innovation (MOSTI) in Malaysia under their *eScienceFund* research grant (01-02-03-SF0240). We would also like to sincerely thank Vlaicu Crisan who provided detailed feedback pertaining to the DSNS-generated compositions, and Cameron Browne for his various contributions to the project.

---

[21] This is with reference to 'traditional' chess problems that must abide by various composition conventions; not all of which have aesthetic merit. However, there is a new, comparable class of chess problems known as 'chess constructs' which have aesthetic value yet need not abide by such conventions (Iqbal, 2014b, 2014c). These are therefore less esoteric and more accessible to a wider audience of players and composers. By this standard, the compositions generated are likely of even higher quality.

## REFERENCES

1. Abe, K., Sakamoto, K. and Nakagawa, M. (2006). A Computational Model of Metaphor Generation Process. In Proceedings of the 28th Annual Meeting of the Cognitive Science Society, pp. 937-942.

2. Ananthanarayanan, R., Esser, S. K., Simon, H. D., & Modha, D. S. (2009). The Cat Is Out of the Bag: Cortical Simulations with 109 neurons, 1013 Synapses. In High Performance Computing Networking, Storage and Analysis, Proceedings of the Conference On. IEEE, pp. 1-12.

3. Didierjean, A. and Gobet, F. (2008). Sherlock Holmes – An Expert's View of Expertise. British Journal of Psychology, 99(1), 109-125.

4. Apter, M. J. (1977). Can Computers Be Programmed to Appreciate Art? Leonardo, pp. 17-21.

5. Back, T., Hammel, U. and Schwefel, H. P. (1997). Evolutionary Computation: Comments on the History and Current State. Evolutionary Computation, IEEE Transactions on, 1(1), pp. 3-17.

6. Battelle, J. (2006). The Search: How Google and Its Rivals Rewrote the Rules of Business and Transformed Our Culture, Portfolio, Penguin Group (USA) Inc.

7. Benghi, C. and Ronchie, G. (2013). An Artificial Intelligence System to Mediate the Creation of Sound and Light Environments. In Proceedings of the Fourth International Conference on Computational Creativity, p. 220.

8. Bezryadin, S., Bourov, P. and Ilinih, D. (2007). Brightness Calculation in Digital Image Processing. In International Symposium on Technologies for Digital Photo Fulfillment, Vol. 2007, No. 1, pp. 10-15. Society for Imaging Science and Technology.

9. Boden, M. A. (2009). Computer Models of Creativity. AI Magazine, 30(3), 23.

10. Box, G. E. (1957). Evolutionary Operation: A Method for Increasing Industrial Productivity. Applied Statistics, pp. 81-101.

11. Buchanan, B. G. (2001). Creativity at the Metalevel: AAAI-2000 Presidential Address. AI Magazine, 22(3), 13.

12. Cameron, J. (Director). (1991). Terminator 2: Judgment Day [Motion Picture]. United States: Carolco Pictures.

13. Chalmers, D. J. (1995). Facing Up to the Problem of Consciousness. Journal of Consciousness Studies, 2(3), pp. 200-219.

14. Chalmers, D. J. (2012). Personal Communication (with main author). 17 November.

15. Chandrasekar, R. (2014). Elementary? Question Answering, IBM's Watson, and the Jeopardy! Challenge. Resonance, 19(3), 222-241.

16. Clement, C. A. and Gentner, D. (1991). Systematicity as a Selection Constraint in Analogical Mapping. Cognitive Science, 15(1), 89-132.

17. Cohen, H. (1999). Colouring Without Seeing: A Problem in Machine Creativity. AISB Quarterly, 102, pp. 26-35.

18. Colton, S. and Steel, G. (1999). Artificial Intelligence and Scientific Creativity. Artificial Intelligence and the Study of Behaviour Quarterly, Vol. 102.

19. Colton, S. (2008). Creativity Versus the Perception of Creativity in Computational Systems. In AAAI Spring Symposium: Creative Intelligent Systems, pp. 14-20.

20. Colton, S., Cook, M. and Raad, A. (2011). Ludic Considerations of Tablet-based Evo-art. In Applications of Evolutionary Computation, pp. 223-233. Springer Berlin Heidelberg.

21. Cope, D. (2005). Computer Models of Musical Creativity. Cambridge: MIT Press.

22. Correia, J., Machado, P., Romero, J. and Carballal, A. (2013). Evolving Figurative Images Using Expression-Based Evolutionary Art. In Proceedings of the Fourth International Conference on Computational Creativity, p. 24.

23. Dennett, D. C. (1996). Darwin's Dangerous Idea: Evolution and the Meanings of Life. Simon and Schuster.

24. Dennett, D. C. (2012). Personal Communication (with main author). 27 August.

25. Eigenfeldt, A. and Pasquier, P. (2013). Considering Vertical and Horizontal Context in Corpus-based Generative Electronic Dance Music. In Proceedings of the Fourth International Conference on Computational Creativity, p. 72.

26. Ekbia, H. R. (2008). Artificial Dreams: The Quest for Non-biological Intelligence. Cambridge University Press.

27. Fraser, A. and Burnell, D. (1970). Computer Models in Genetics. McGraw-Hill.

28. Galanter, P. (2012). Computational Aesthetic Evaluation: Past and future. In Computers and Creativity, pp. 255-293. Springer Berlin Heidelberg.

29. Grace, K., Gero, J. and Saunders, R. (2013). Learning How to Reinterpret Creative Problems. In Proceedings of the Fourth International Conference on Computational Creativity, p. 113.

30. Harris, S. (2012). Free Will. Simon and Schuster.

31. Hesse, H. (2011). The Joys of Chess: Heroes, Battles and Brilliancies. New in Chess, 1st Edition.

32. Holmes, B. (1996). The Creativity Machine. New Scientist, pp. 22-26.

33. Iqbal, A. and Yaacob, M. (2008). Computational Assessment of Sparsity in Board Games. In Proceedings of the 12th International Conference on Computer Games: AI, Animation, Mobile, Educational and Serious Games (CGames 2008), Louisville, Kentucky, USA, Vol. 30, pp. 29-33.

34. Iqbal, M. A. M. (2008). A Discrete Computational Aesthetics Model for a Zero-Sum Perfect Information Game, Ph.D. Thesis, University of Malaya, Kuala Lumpur, Malaysia. http://metalab.uniten.edu.my/~azlan/Research/pdfs/phd_thesis_azlan.pdf

35. Iqbal, A. (2011). Increasing Efficiency and Quality in the Automatic Composition of Three-Move Mate Problems, in Entertainment Computing - ICEC 2011, Lecture Notes in Computer Science, Vol. 6972, pp. 186-197. Anacleto, J.; Fels, S.; Graham, N.; Kapralos, B.; Saif El-Nasr, M.; Stanley, K. (Eds.). 1st Edition., 2011, XVI. Springer. ISBN 978-3-642-24499-5

36. Iqbal, A. (2012). A Computer Program to Identify Beauty in Problems and Studies, ChessBase News, Hamburg, Germany. 15 December. http://en.chessbase.com/post/a-computer-program-to-identify-beauty-in-problems-and-studies

37. Iqbal, A. van der Heijden, H., Guid, M. and Makhmali, A. (2012). Evaluating the Aesthetics of Endgame Studies: A Computational Model of Human Aesthetic Perception, IEEE Transactions on Computational Intelligence and AI in Games: Special Issue on Computational Aesthetics in Games, Vol. 4, No. 3, pp. 178-191. ISSN 1943-068X. e-ISSN 1943-0698

38. Iqbal, A. (2014a). How Relevant Are Chess Composition Conventions?, in Communications in Computer and Information Science, Vol. 408, pp. 122-131. Simone Diniz Junqueira Barbosa, Phoebe Chen, Alfredo Cuzzocrea, Xiaoyong Du, Joaquim Filipe, Orhun Kara, Igor Kotenko, Krishna M. Sivalingam, Dominik Ślęzak, Takashi Washio, Xiaokang Yang (Eds.). Springer. Online ISSN 1865-0937, Print ISSN 1865-0929

39. Iqbal, A. (2014b). Introducing 'Chess Constructs'. ChessBase News, Hamburg, Germany, 29 June http://en.chessbase.com/post/azlan-iqbal-introducing-chess-constructs

40. Iqbal, A. (2014c). Best 'Chess Constructs' by ChessBase Readers. ChessBase News, Hamburg, Germany, 26 July http://en.chessbase.com/post/best-chess-constructs-by-chessbase-readers

41. Iqbal, A. (2015). Computer Science and Artificial Intelligence: "Computational Aesthetics", in Encyclopedia Britannica, 11 February. Encyclopedia Britannica, Inc., Chicago, USA. http://global.britannica.com/EBchecked/topic/2011991/computational-aesthetics

42. Johnson, C. G. (2012). The Creative Computer as Romantic Hero? Computational Creativity Systems and Creative Personae; In Proceedings of the Third International Conference on Computational Creativity, May, pp. 57-61.

43. Jonze, S. (Director). (2001). Her [Motion Picture]. United States: Annapurna Pictures.

44. Kasparov, G. (2014). Personal Communication (with main author). 25 April.

45. Kerr, D. A. (2008). The ISO Definition of the Dynamic Range of a Digital Still Camera. http://dougkerr.net/Pumpkin/articles/ISO_Dynamic_range.pdf

46. Koren, N. (2006). The Imatest Program: Comparing Cameras with Different Amounts of Sharpening. In Electronic Imaging 2006, p. 60690L. International Society for Optics and Photonics.

47. Kurzweil, R. (2012). How to Create a Mind: The Secret of Human Thought Revealed. Penguin.

48. Levy, D. N. (2006). Robots Unlimited: Life in a Virtual Age. Wellesley: AK Peters.

49. Machado, P. and Amaro, H. (2013). Fitness Functions for Ant Colony Paintings. In Proceedings of the Fourth International Conference on Computational Creativity, p. 90.

50. McCarthy, J. (2005). Personal Communication (with main author). 17 September.

51. McCorduck, P. (2004). Machines Who Think: A Personal Inquiry into the History and Prospects of Artificial Intelligence. AK Peters Ltd.

52. McCulloch, W. S. and Pitts, W. (1943). A Logical Calculus of the Ideas Immanent in Nervous Activity. The Bulletin of Mathematical Biophysics, 5(4), 115-133.

53. Newborn, M. (1997). Kasparov vs. Deep Blue: Computer Chess Comes of Age. Springer-Verlag New York, Inc..

54. Norton, M. P. and Karczub, D. G. (2003). Fundamentals of Noise and Vibration Analysis for Engineers. Cambridge University Press.

55. Nunn, J. (2011). Personal Communication (with main author). 22 February.

56. Pease, A., Colton, S., Ramezani, R., Charnley, J., & Reed, K. (2013). A Discussion on Serendipity in Creative Systems. In Proceedings of the Fourth International Conference on Computational Creativity, p. 64.

57. Pérez, R. P. Y. and Ortiz, O. (2013). A Model for Evaluating Interestingness in a Computer-Generated plot. In Proceedings of the Fourth International Conference on Computational Creativity, p. 131.

58. Phillips, S. (2014). Analogy, Cognitive Architecture and Universal Construction: A Tale of Two Systematicities. PloS one, 9(2), e89152.

59. Pfister, W. (Director). (2014). Transcendence [Motion Picture]. United States: Alcon Entertainment.

60. Ritchie, G. (2007). Some Empirical Criteria For Attributing Creativity to a Computer Program. Minds and Machines, 17(1), pp. 67-99.

61. Ruiz, S., Buyukturkoglu, K., Rana, M., Birbaumer, N. and Sitaram, R. (2014). Real-time fMRI Brain Computer Interfaces: Self-regulation of Single Brain Regions to Networks. Biological Psychology, 95, pp. 4-20.

62. Sargent, J. (Director). (1970). Colossus: The Forbin Project [Motion Picture]. United States: Universal Pictures.

63. Saunders, R., Chee, E. and Gemeinboeck, P. (2013). Evaluating Human-Robot Interaction with Embodied Creative Systems. In Proceedings of the Fourth International Conference on Computational Creativity, pp. 205-209.

64. Scott, R. (Director). (1982). Blade Runner [Motion Picture]. United States: Warner Bros.

65. Seung, S. (2012). Connectome: How the Brain's Wiring Makes Us Who We Are. Houghton Mifflin Harcourt.

66. Shachtman, N. (2009). Darpa’s Simulated Cat Brain Project a ‘Scam’: Top Scientist. Wired Magazine. http://www.wired.com/2009/11/darpas-simulated-cat-brain-project-a-scam-top-neuroscientist/

67. Shannon, C. E. (1950). Programming a Computer for Playing Chess. Philosophical Magazine, 41(314), pp. 256-275.

68. Sibley, F. (1959). Aesthetic Concepts. The Philosophical Review, pp. 421-450.

69. Spielberg, S. (Director). (2001). A.I. [Motion Picture]. United States: Warner Bros.

70. Srinivasan, S., Petkovic, D. and Ponceleon, D. (1999). Towards Robust Features for Classifying Audio in the CueVideo System. In Proceedings of the Seventh ACM international Conference on Multimedia (Part 1), pp. 393-400. ACM.

71. Strang, S., Utikal, V., Fischbacher, U., Weber, B. and Falk, A. (2014). Neural Correlates of Receiving an Apology and Active Forgiveness: An fMRI Study. PloS one, 9(2), e87654.

72. Sternberg, R. J., & Kaufman, J. C. (Eds.). (2010). The Cambridge Handbook of Creativity. Cambridge University Press.

73. Terai, A. and Nakagawa, M. (2009). A Neural Network Model of Metaphor Generation with Dynamic Interaction. In Alippi, C., Polycarpou, M., Panayiotou, C., Ellinas, G., eds., ICANN 2009, Part I. LNCS, 5768, Springer, Heidelberg, pp. 779–788.

74. Turing, A. (1953). Digital Computers Applied to Games, in Faster than Thought, Chapter 25 (subsection), B. V. Bowden (ed.), Pitman, London.

75. Velimirovic, M. and Valtonen, K. (2013). The Definitive Book: Encyclopedia of Chess Problems - Themes and Terms, Reprinted Edition

76. Wachowski, A. and Wachowski, L. (Directors). (1999). The Matrix [Motion Picture]. United States: Warner Bros.

77. Warwick, K. (2011). Artificial Intelligence: The Basics. Routledge.

78. Yu, Y., Xu, D., Chen, C., Yu, Y. and Zhao, L. (2006). A Surface Errors Locator System for Ancient Culture Preservation. In Digital Libraries: Achievements, Challenges and Opportunities, pp. 360-369. Springer Berlin Heidelberg.

**APPENDIX A: Chess Problem Composing Steps**

Chess problems for this research were composed automatically using CHESTHETICA (see Figure 9) following the essential steps below. It is a modification of the approach used in (Iqbal, 2011). The problems composed are limited to orthodox mate-in-3 problems in standard international chess.

1. Obtain the two *new* DSNS strings produced from the DSNS process (see section 3.1).
2. Set the total number of white pieces and black pieces that can be used in the composition. There are two attribute values pertaining to these features in each DSNS string.
    a. For example, if in string 1 the white piece count is 5 and in string 2 the white piece count is 10, the range possible for this computer composition is between 5 and 10. The same for the black pieces.
3. Calculate the total Shannon value of the white pieces and then the black pieces in both strings and get the average of each. Use these average values to determine the number of piece permutations (i.e. combinations of different pieces) that satisfy them.
    a. For example, an average value of 10 for white could mean having a bishop, two knights and a pawn whereas an average value of 9 for black could mean having just a queen. The total number of piece permutations possible for both the white and black pieces here is totaled.
    b. This total will equal the number of times the same pair of DSNS strings is used in attempting to generate a composition. So, in principle, every legal piece combination can be tested.
4. Determine which permutations of pieces for both white and black satisfy (2) and are 'reasonable', i.e. in total, there are no more than 8 pieces – other than pawns – on the board.
    a. For example, if the range of pieces for white that can be used in the composition is between 4 and 6, then permutations with only 3 pieces are excluded.
    b. 'Reasonable' means that the position should be realistic. Typically, only one or two of the pieces on the board for a particular color would have a pawn promoted to it. So, if the upper limit of the total number of white pieces allowed (as per (2)) is 12 and you have a possible permutation with one queen, four rooks, four bishops and a knight, this will be excluded because even though the total number of pieces is 10 (below 12), it is more than 8 pieces. On the other hand, one queen, three rooks, two bishops and two knights would be acceptable and more realistic.
5. If no permutations for both white and black can be found that satisfy the requirements in (4), return to (1), otherwise use a random, valid one for each.
6. Place the two kings on random squares on the board. Accept them so far as the resulting position is legal; otherwise, keep repeating the process.
7. Choose at random one of six possibilities (i.e. the five remaining piece types and a 'blank square') based on equal probability (i.e. 1 in 6 or 16.67%). Alternate between white (first) and then black,.
    a. If a 'blank' is chosen (which could be for either white or black) return to (7) and choose for the opponent's color instead. So a blank means one color misses its 'turn' and could therefore have fewer pieces on the board in the end.
8. Choose a random square until one that is unoccupied is found.
    a. This is where, if a piece was chosen in (7), it will be placed.
9. Determine, based on the two DSNS strings, which are the first and last pieces to have moved.
10. If the piece chosen in (7) is not the same as any of the piece types determined in (9) and none of the latter are already on the board, set a 50% chance that the former will have to be chosen again.
11. Place the chosen piece on the square determined in (8).
12. Check the legality of the position in terms of chess rules, taking into account the constraints mentioned earlier.
    a. For example, having a pawn occupying the eighth rank is illegal.
    b. The possibility of castling was given a 'neutral' 50% random probability of being legal, assuming a king and one of its rooks happen to be on the right squares. Determination of legality based on retrograde analysis was considered too complicated and unnecessary for the purposes of this research. 'Officially', in compositions, castling in the key move is legal unless it can be proven otherwise.
    c. En passant captures, if plausible, default to illegal. En passant is considered legal only if it can be proven the last move by the opponent permitted it.
13. If the position is illegal, remove the chosen piece from the board and return to (7).
14. Determine if the material difference between white and black for the position at present exceeds the higher of the two Shannon material differences in the two DSNS strings.

15. If (14) is true then clear the board and start composing a new problem; return to (3).
    a. The same DSNS strings are used but with a refreshed piece permutation array.
16. Sum the sparsity values from both DSNS strings.
17. If the total from (16) ≥ 1 (i.e. leaning toward a sparser position) then determine if the sparsity value of the present position is less than 0.25 (i.e. leaning toward density). If so, clear the board and start composing a new problem; return to (3).
18. If the total from (16) < 1 (i.e. leaning toward a denser or crowded position) then determine if the sparsity value of the present position is more than 0.75 (i.e. leaning toward a sparser position). If so, clear the board and start composing a new problem; return to (3).
19. Keep the piece chosen in (7) on the board and return to (7) to choose a new piece for the opponent's army until all the constraints above have been satisfied.
20. Test with a mate-solver engine to determine if the tentatively acceptable position generated has a forced mate-in-3 solution to it. If not, remove the last chosen piece from the board; return to (7).
    a. CHESTHETICA communicates with ChestUCI v5.2 for this purpose (5 second search limit).
21. If there is such a solution, the position is 'optimized' as shown in the code below. This makes the composition more economical in form.
    a. FOR every square

```
FOR every square
    IF not occupied by a king and not empty THEN
        Remove piece
        IF forced mate-in-3 can still be found THEN
            Proceed
        ELSE
            Return piece to its original location
        END IF
    END IF
NEXT
```

    b. To be thorough, optimization is performed three times, starting from the upper left to the lower right of the board; white pieces first, then black, and then white again. Fewer passes proved to be insufficient in certain positions. Optimization generally increases the aesthetic quality of a composition by removing unnecessary or passive pieces, but not always.
    c. Sometimes, optimization may not be possible.
22. Test for conformity to composition conventions specified, if any. If there are any conventions specified not satisfied, pieces are added by returning to (7).
    a. In the case of the convention '*no restricting enemy king movement in key move*', the piece chosen earlier is actually removed before returning to (7) to help avoid this problem the next time around.
    b. The constraints mentioned earlier on need not be re-applied after the optimization and conformity to convention processes.
23. Accept the composition as valid, optimized and in conformity with specified conventions.
    a. The composition is stored in a PGN file.
24. Clear the board and return to (5) if the total number of permutations in 3(b) is not yet reached, otherwise, return to (1).
    a. So a particular pair of DSNS strings can possibly be used to generate more than one composition (or perhaps none).

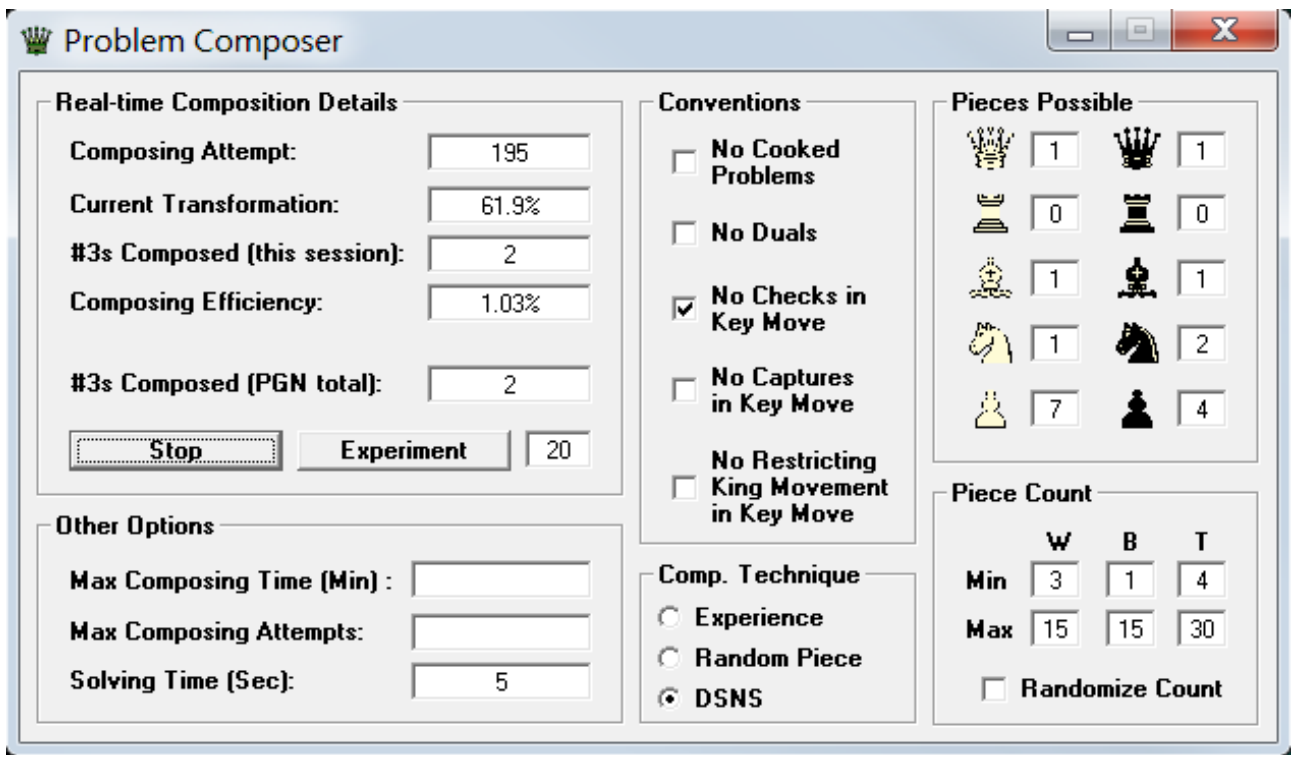

**Figure 9**: CHESTHETICA v9.22 composition interface window.

# APPENDIX B: Two Columns of DSNS Strings in a Spreadsheet

| 2 | COMP #3 >3.5 - Set A.pgn | white pieces | black pieces | shannon white pieces | shannon black pieces | shannon difference | moves | year | first piece to move | last piece to move | sparsity | \|d\| | DEVIATION | Σ÷ | COMP #3 >3.5 - Set B.pgn | white pieces | black pieces | shannon white pieces | shannon black pieces | shannon difference | moves | year | first piece to move | last piece to move | sparsity |
|---|---|---|---|---|---|---|---|---|---|---|---|---|---|---|---|---|---|---|---|---|---|---|---|---|---|
| 3 | | | | | | | | | | | | | | | | | | | | | | | | | |
| 4 | 8/1p2BN1K/4Qp2/n1R4p/3k2P1/P5n1/4P3/1r6 w - - 0 1 | 8 | 7 | 23 | 14 | 9 | 3 | 1908 | 2 | 2 | 0.45454545 | 123.015 | 124.801 | 21.786 | 8/8/4p3/2N1K3/RP2P3/2k5/p1P1Q3/b7 w - - 0 1 | 7 | 4 | 20 | 5 | 15 | 3 | 2009 | 2 | 2 | 0.44 |
| 5 | 5rk1/5qpn/8/3N4/3B4/1B6/1KP3R1/8 w - - 0 1 | 6 | 5 | 15 | 18 | 3 | 3 | 1851 | 4 | 2 | 0.31428571 | 179.017 | 183.768 | 24.751 | 1n1n3q/R1prBN2/1Nk1p3/Qp5r/2p2P1P/2P5/3P2R1/5K1 | 12 | 10 | 35 | 29 | 6 | 3 | 1982 | 1 | 2 | 0.2972973 |
| 6 | 5Q2/b2k1P2/1n1NNn2/1P1p4/6P1/8/8/7K w - - 0 1 | 7 | 5 | 18 | 10 | 8 | 3 | 1883 | 2 | 1 | 0.35294118 | 63.314 | 66.947 | 23.633 | 1k6/1P6/P7/2P1P3/8/Q7/8/K4b2 w - - 0 1 | 6 | 2 | 13 | 3 | 10 | 3 | 1927 | 1 | 1 | 0.66666667 |
| 7 | r4Qb1/p1p3Pk/P6p/7P/8/6K1/5R2/8 w - - 0 1 | 6 | 6 | 17 | 11 | 6 | 3 | 1930 | 4 | 2 | 0.35294118 | 31.004 | 32.305 | 21.301 | 1K4N1/5np1/4N2b/3ppB1k/8/5pP1/Q3p2P/8 w - - 0 1 | 7 | 8 | 20 | 11 | 9 | 3 | 1949 | 2 | 1 | 0.34883721 |
| 8 | b7/3NQ3/3N4/1n1k2n1/8/p2K4/8/8 w - - 0 1 | 4 | 5 | 15 | 10 | 5 | 3 | 1888 | 2 | 2 | 0.52941176 | 60.129 | 62.435 | 22.306 | 8/1K6/8/kp6/1n6/1PB1p3/8/1N6 w - - 0 1 | 4 | 4 | 7 | 5 | 2 | 3 | 1930 | 3 | 2 | 0.4 |
| 9 | r1q1r3/8/k1p1p3/1pK5/7R/1P6/8/4QB2 w - - 0 1 | 5 | 7 | 18 | 22 | 4 | 3 | 1849 | 5 | 1 | 0.54545455 | 60.165 | 61.925 | 21.760 | 8/7r/1p1pN1Q1/np2k3/3R3P/n4p2/KPP5/3b4 w - - 0 1 | 7 | 9 | 20 | 18 | 2 | 3 | 1894 | 2 | 1 | 0.38095238 |
| 10 | 5bb1/5p1n/3p1pkP/3K1NpN/6P1/8/8/8 w - - 0 1 | 5 | 8 | 8 | 13 | 5 | 3 | 1908 | 2 | 1 | 0.22033898 | 95.258 | 103.269 | 28.011 | 8/1P3P1R/1P1k1P2/2p5/2p1K2P/2P5/8/8 w - - 0 1 | 8 | 3 | 11 | 2 | 9 | 3 | 1975 | 1 | 2 | 0.47826087 |
| 11 | 8/1K3Q2/2P5/4k1Pn/1P6/2PBbPpn/8/3b2N1 w - - 0 1 | 9 | 6 | 20 | 13 | 7 | 3 | 1885 | 1 | 1 | 0.48387097 | 115.183 | 118.963 | 23.780 | 3N2R1/r1pN1B2/4pp1r/1P1kqb1R/1Q6/3pnpBp/1b2PP2/ | 11 | 14 | 34 | 37 | 3 | 3 | 1947 | 2 | 1 | 0.30120482 |
| 12 | 3b4/2p5/6p1/6k1/8/3NN3/1Q6/5K2 w - - 0 1 | 4 | 4 | 15 | 5 | 10 | 3 | 1993 | 2 | 2 | 0.57142857 | 179.279 | 193.410 | 34.131 | 1B6/1n2K3/1rprNN2/4k3/1p1pbpP1/R7/2nP1P2/2R5 w - | 9 | 10 | 22 | 23 | 1 | 3 | 1862 | 4 | 1 | 0.29230769 |
| 13 | 8/4p3/7B/4p3/4k1P1/8/5K2/3R4 w - - 0 1 | 4 | 3 | 9 | 2 | 7 | 3 | 1845 | 3 | 4 | 0.77777778 | 230.522 | 252.478 | 41.956 | K7/4QP1B/2pp1rR1/1PPk1r1p/P2p1P1q/BP2bbP1/1N6/4 | 14 | 10 | 35 | 29 | 6 | 3 | 2000 | 2 | 1 | 0.25531915 |
| 14 | 1Q1b4/r1n1ppB1/r7/R1Nb2k1/4B1p1/6P1/5nK1/7R w - - | 8 | 10 | 29 | 25 | 4 | 3 | null | 5 | 4 | 0.36 | 48.107 | 54.732 | 24.625 | k2r4/1b6/N7/1PK1Q3/8/8/8/8 w - - 0 1 | 4 | 3 | 13 | 8 | 5 | 3 | 1814 | 5 | 1 | 0.46666667 |
| 15 | 8/8/1Kp1N3/8/2k1r3/1R2p3/B7/2bR4 w - - 0 1 | 5 | 5 | 16 | 10 | 6 | 3 | 1989 | 2 | 2 | 0.5 | 60.100 | 62.705 | 20.605 | 8/8/B7/1Np4Q/4k3/6K1/7N/1n2n3 w - - 0 1 | 5 | 4 | 18 | 7 | 11 | 3 | 1940 | 2 | 2 | 0.6 |
| 16 | K7/b3p2k/2p1B3/4QPPp/1p6/4p3/1p6/5n2 w - - 0 1 | 5 | 9 | 14 | 12 | 2 | 3 | 1997 | 1 | 1 | 0.5 | 100.152 | 100.768 | 20.616 | 8/3p3Q/2rp1q1B/R1nN2N1/3k4/1P3p1r/K3P3/8 w - - 0 1 | 8 | 8 | 25 | 25 | 0 | 3 | 1929 | 3 | 1 | 0.34782609 |
| 17 | 8/1k2N3/RB5K/1Bp1N3/8/3p4/2P4p/1r5r w - - 0 1 | 7 | 6 | 18 | 13 | 5 | 3 | 1924 | 4 | 1 | 0.39393939 | 77.606 | 87.730 | 30.124 | 4n3/P7/4K1R1/8/8/8/2R5/5k2 w - - 0 1 | 4 | 2 | 11 | 3 | 8 | 3 | 1968 | 1 | 4 | 1 |
| 18 | 5n1k/8/8/5Q2/5q1B/8/8/6KR w - - 0 1 | 4 | 3 | 17 | 12 | 5 | 3 | 1935 | 5 | 5 | 0.63636364 | 30.136 | 34.832 | 24.696 | 4n3/1p2Kp2/8/4k2p/2p1PR2/p7/5P1p/bQ6 w - - 0 1 | 5 | 9 | 16 | 12 | 4 | 3 | 1952 | 5 | 1 | 0.5 |
| 19 | 8/1R6/8/1Q1n4/8/b4K1p/8/6k1 w - - 0 1 | 3 | 4 | 14 | 7 | 7 | 3 | 1996 | 5 | 4 | 1 | 48.625 | 54.372 | 23.747 | B3b2K/1n2Rp2/2Pk4/5Pp1/1P3N2/3pp1B1/1R5b/Q7 w - | 10 | 8 | 31 | 13 | 18 | 3 | null | 4 | 2 | 0.375 |
| 20 | B3R2n/4Pr2/1KNN4/1ppk1Ppb/3p1RP1/P1P5/4P3/Qq2b3 | 13 | 10 | 34 | 27 | 7 | 3 | 1973 | 2 | 3 | 0.29113924 | 112.288 | 119.379 | 27.091 | 3N1B2/8/qP2NkBK/8/3P1P2/7P/8/8 w - - 0 1 | 9 | 2 | 16 | 9 | 7 | 3 | 1911 | 2 | 1 | 0.57894737 |
| 21 | 1b1q4/1N1N4/2B5/4p3/1Pk5/7p/1K6/3RB3 w - - 0 1 | 7 | 5 | 18 | 14 | 4 | 3 | 1971 | 6 | 2 | 0.5 | 106.194 | 108.306 | 22.112 | 8/8/4Rp1r/1Q1p2Bp/nK1kp3/1p3r2/3nP3/8 w - - 0 1 | 5 | 10 | 18 | 21 | 3 | 3 | 1884 | 3 | 1 | 0.30612245 |
| 22 | Kb1Q2b1/n1p5/1R1pr3/1B1k1P2/R1NPN2P/2B2r2/3n1P2/ | 12 | 10 | 35 | 33 | 2 | 3 | 1972 | 5 | 3 | 0.2972973 | 21.026 | 22.298 | 21.273 | rQ4q1/2R5/3N1pRp/r2p3n/1NpP1kPB/5P2/5PK1/2n5 w - | 11 | 10 | 32 | 29 | 3 | 3 | 1964 | 2 | 2 | 0.32307692 |
| 23 | 8/8/6B1/6p1/8/6K1/4k3/2R2N2 w - - 0 1 | 4 | 2 | 11 | 1 | 10 | 3 | null | 4 | 3 | 0.6 | 39.284 | 61.253 | 39.969 | 3n4/K1k1p2n/1Nb2B1r/1P1QP3/8/8/8/8 w - - 0 1 | 6 | 6 | 17 | 15 | 2 | 3 | 1872 | 1 | 1 | 0.31578947 |
| 24 | 8/3PnknP/5N2/6K1/8/8/8/8 w - - 0 1 | 4 | 3 | 5 | 6 | 1 | 3 | 1957 | 1 | 2 | 0.30434783 | 51.096 | 52.746 | 21.650 | 3nK3/8/1b4B1/1p2k3/2pp1R2/3p1p2/7B/8 w - - 0 1 | 4 | 8 | 11 | 11 | 0 | 3 | 1926 | 3 | 3 | 0.4 |
| 25 | 1R6/2PR4/8/1N6/8/6Q1/8/kn3K2 w - - 0 1 | 6 | 2 | 23 | 3 | 20 | 3 | 1952 | 4 | 2 | 0.57142857 | 63.000 | 64.111 | 21.111 | 8/1N4Q1/n7/1N1k4/4b3/K7/8/3B4 w - - 0 1 | 5 | 3 | 18 | 6 | 12 | 3 | 1908 | 3 | 2 | 0.57142857 |
| 26 | 3b3r/1bn1P3/1r5Q/pPk1p3/K2N1P2/1B1P4/4P2p/6B1 w - | 10 | 9 | 23 | 22 | 1 | 3 | 1875 | 3 | 1 | 0.33333333 | 121.095 | 122.949 | 21.854 | b1QN1b2/5p1r/3p1B2/1p1k1KNp/2p5/n1p2P2/2pp4/r5R | 7 | 14 | 24 | 27 | 3 | 3 | 1979 | 4 | 1 | 0.42857143 |
| 27 | 2n5/6K1/4p3/6np/8/8/5N1k/4BQ2 w - - 0 1 | 4 | 5 | 15 | 8 | 7 | 3 | 1914 | 6 | 2 | 0.52941176 | 106.257 | 110.677 | 24.420 | 1rk1bN2/Pp1Rp1Pp/1B1pPp1b/8/KP6/8/R1B4r/1Q6 w - | 11 | 10 | 32 | 21 | 11 | 3 | 1972 | 5 | 1 | 0.27272727 |
| 28 | 1b2K3/rn1P1p2/pP1k1PpP/2Rbp1NQ/3Pp3/7P/2NR1q2/3r | 12 | 12 | 31 | 33 | 2 | 3 | 1968 | 2 | 1 | 0.24489796 | 149.103 | 151.354 | 22.251 | 8/1b5r/1pNp2p1/4pp2/1P2k1BQ/2K1N3/2n3P1/8 w - - 0 | 7 | 9 | 20 | 16 | 4 | 3 | 1858 | 2 | 2 | 0.34782609 |
| 29 | 1Q1b4/3Pr3/3R1B1n/1N2RpP1/K4kp1/4pP2/1Pp1Pp1P/n1 | 12 | 12 | 31 | 33 | 2 | 3 | 1999 | 2 | 4 | 0.25 | 25.053 | 23.885 | 18.832 | N7/KPB5/2P1r3/2Q2p2/1RNPk3/1p4P1/bb2P1p1/3r2q1 v | 11 | 9 | 28 | 28 | 0 | 3 | 2007 | 3 | 2 | 0.3030303 |
| 30 | 1N6/3P1Pb1/r2k2P1/q3R3/ppR1P3/5p1K/Q7/Bn6 w - - 0 | 10 | 8 | 29 | 23 | 6 | 3 | 1981 | 5 | 1 | 0.375 | 56.067 | 70.108 | 34.041 | 8/8/8/8/5NQ1/1Np5/1pR2K2/1k6 w - - 0 1 | 5 | 3 | 20 | 2 | 18 | 3 | 1981 | 2 | 2 | 0.30769231 |
| 31 | 2B5/8/1N1P4/4kpP1/8/B4K2/4P3/8 w - - 0 1 | 7 | 2 | 12 | 1 | 11 | 3 | 1904 | 1 | 2 | 0.52941176 | 148.115 | 182.313 | 54.199 | 1b6/5N2/q2p3p/7p/1nprBk1K/r2p4/3P1N2/6R1 w - - 0 1 | 6 | 11 | 15 | 30 | 15 | 3 | 2003 | 4 | 2 | 0.41463415 |
| 32 | N7/2Q3Nb/2p5/n2p2p1/3B4/2Pk4/8/b4K2 w - - 0 1 | 6 | 7 | 19 | 12 | 7 | 3 | 1910 | 5 | 2 | 0.48148148 | 8.166 | 8.761 | 18.596 | 4r2B/2N5/p7/7r/3p4/2k3N1/1R6/K4B2 w - - 0 1 | 6 | 5 | 17 | 12 | 5 | 3 | null | 3 | 2 | 0.64705882 |
| 33 | 8/8/5Nb1/8/1Q1N1k2/8/7K/8 w - - 0 1 | 4 | 2 | 15 | 3 | 12 | 3 | null | 6 | 2 | 0.75 | 30.446 | 37.231 | 24.786 | 8/4Q3/8/8/1p2pNPb/1Np1Rb2/K1n5/3k1B2 w - - 0 1 | 7 | 7 | 24 | 12 | 12 | 3 | 1958 | 2 | 2 | 0.30434783 |
| 34 | 3n4/6p1/3p4/4k3/1pp4Q/4N3/6K1/4R3 w - - 0 1 | 4 | 6 | 17 | 7 | 10 | 3 | 1941 | 5 | 2 | 0.71428571 | 80.078 | 85.006 | 24.928 | 7r/1k2PP2/3Q4/1N6/8/8/6K1/8 w - - 0 1 | 5 | 2 | 14 | 5 | 9 | 3 | 2005 | 1 | 3 | 0.63636364 |
| 35 | 6k1/5pP1/5K1P/8/B7/5N2/8/8 w - - 0 1 | 5 | 2 | 8 | 1 | 7 | 3 | 1976 | 3 | 1 | 0.36842105 | 90.053 | 114.418 | 44.365 | n7/RPNkp2b/7b/r1p2p2/1p6/3pRK2/1Q6/8 w - - 0 1 | 6 | 10 | 23 | 19 | 4 | 3 | 1936 | 5 | 4 | 0.42105263 |
| 36 | 8/1pR5/4p3/4k1K1/8/2Np2p1/3P2B1/8 w - - 0 1 | 5 | 5 | 12 | 4 | 8 | 3 | null | 2 | 1 | 0.45454545 | 29.066 | 35.001 | 23.935 | 1b1nN2B/7Q/3p4/6k1/6p1/2P2prp/6b1/3K4 w - - 0 1 | 5 | 9 | 16 | 18 | 2 | 3 | 1961 | 2 | 2 | 0.38888889 |
| 37 | 6bK/3B3p/1b1p4/2pkr1NR/R3p3/1P2p3/1nP1P2p/4Q2r w | 9 | 12 | 28 | 25 | 3 | 3 | 1996 | 2 | 1 | 0.33333333 | 210.667 | 268.720 | 78.053 | 8/7P/8/8/8/5K1p/8/7k w - - 0 1 | 2 | 2 | 1 | 1 | 0 | 3 | 1861 | 1 | 4 | 1 |
| 38 | 8/8/8/5N1k/8/5K1N/8/3R4 w - - 0 1 | 4 | 1 | 11 | 0 | 11 | 3 | 1983 | 2 | 4 | 1 | 100.632 | 104.838 | 24.206 | 5Nkb/PPpp2n1/3P4/3P3B/2P2K2/8/6R1/8 w - - 0 1 | 9 | 5 | 16 | 8 | 8 | 3 | 1912 | 5 | 3 | 0.36842105 |
| 39 | n7/r1RP1K2/3p4/B1pk1pPQ/N2bpP2/n7/P1NrB3/3R2q1 w | 12 | 11 | 35 | 32 | 3 | 3 | 1957 | 5 | 1 | 0.31506849 | 141.152 | 143.831 | 22.680 | 8/2b4r/K1p1pN2/4k1pp/8/2PR4/B4Q2/5n2 w - - 0 1 | 6 | 8 | 21 | 15 | 6 | 3 | 1861 | 4 | 2 | 0.46666667 |
| 40 | 4RQ2/1rp5/PbP1NRKp/3k1p2/Pp3pP1/1n3b2/2nBPq2/2rN | 13 | 13 | 36 | 36 | 0 | 3 | 1955 | 5 | 2 | 0.25 | 89.329 | 91.504 | 22.175 | qb5K/2n1R2P/7k/2n5/6Q1/8/8/5b1r w - - 0 1 | 4 | 7 | 15 | 26 | 11 | 3 | 1925 | 4 | 1 | 0.57894737 |
| 41 | 6br/3pP3/pK3Q1p/3k4/8/p7/1pPP1P2/qrn5 w - - 0 1 | 6 | 11 | 13 | 30 | 17 | 3 | 1918 | 1 | 1 | 0.36170213 | 61.044 | 69.378 | 28.334 | 7Q/2q5/n7/4Nn2/r2R4/1Rp1kB2/p1b5/K3B3 w - - 0 1 | 7 | 8 | 28 | 25 | 3 | 3 | 1936 | 5 | 2 | 0.40540541 |
| 42 | 8/1B6/2NB1P2/1p1k3P/2b1R2K/8/3P4/3Q4 w - - 0 1 | 9 | 3 | 26 | 4 | 22 | 3 | 1887 | 3 | 2 | 0.375 | 62.096 | 71.681 | 29.585 | 5r2/4n1p1/1q1RNb2/1p2k3/1Q3NK1/4r1pn/p4p2/8 w - - | 5 | 12 | 20 | 33 | 13 | 3 | 1883 | 2 | 2 | 0.27868852 |
| 43 | b7/4pP2/5pRp/R2q1k1B/4np1K/3P4/1Q3p2/8 w - - 0 1 | 7 | 9 | 24 | 20 | 4 | 3 | 1992 | 5 | 4 | 0.33333333 | 31.086 | 33.843 | 20.757 | 8/8/Kb1B2B1/5R2/2Ppk1P1/2p5/4pp2/2R5 w - - 0 1 | 7 | 6 | 18 | 7 | 11 | 3 | null | 4 | 3 | 0.41935484 |
| 44 | 5q1n/R4P1k/4P1Nb/8/3P4/8/B1Q5/K6R w - - 0 1 | 9 | 4 | 28 | 15 | 13 | 3 | 1935 | 1 | 2 | 0.44827586 | 47.178 | 55.287 | 28.108 | 8/6p1/6Q1/1NnNpp2/1pB1k3/8/p1b1PbK1/5rn1 w - - 0 1 | 6 | 11 | 19 | 22 | 3 | 3 | 1941 | 5 | 1 | 0.26984127 |
| 45 | K6R/1PP1k1n1/6pP/6PP/1rN2N2/Q1p4b/8/1r6 w - - 0 1 | 10 | 7 | 25 | 18 | 7 | 3 | 1924 | 5 | 1 | 0.34693878 | 83.107 | 85.546 | 22.439 | 3r4/2p1pPK1/1qPpkp2/Rp1N1p2/4Np1P/5P2/3RB1n1/Br | 12 | 12 | 35 | 29 | 6 | 3 | 1974 | 3 | 3 | 0.24 |
| 46 | 6Qb/1p1B4/1N3B2/2k2p2/1N3p1q/4p2p/PPb5/2Kn4 w - - | 8 | 10 | 23 | 23 | 0 | 3 | 1966 | 2 | 1 | 0.39130435 | 151.109 | 162.077 | 30.969 | B7/3K4/8/3Q1p2/8/1pk5/1N6/B7 w - - 0 1 | 5 | 3 | 18 | 2 | 16 | 3 | 1871 | 5 | 2 | 0.5 |
| 47 | 5k1K/3p4/3N4/4P1N1/1B6/8/8/8 w - - 0 1 | 5 | 2 | 10 | 1 | 9 | 3 | 1904 | 3 | 2 | 0.63636364 | 52.136 | 60.184 | 28.048 | 5B2/4N1K1/2B5/1Pp1kb2/8/b1P1p3/8/3Q4 w - - 0 1 | 7 | 5 | 20 | 8 | 12 | 3 | 1879 | 5 | 2 | 0.5 |
| 48 | 5B2/5P1P/4P1k1/8/6K1/8/8/8 w - - 0 1 | 5 | 1 | 6 | 0 | 6 | 3 | 1994 | 1 | 1 | 0.42857143 | 110.117 | 123.907 | 33.790 | 6N1/4p1RP/4kp2/KN1r2rp/2Q2B2/p1p1PPn1/q7/8 w - - | 9 | 10 | 26 | 27 | 1 | 3 | 1950 | 1 | 2 | 0.31147541 |
| 49 | Q3B3/6N1/2NpRp1n/2Pk2n1/3P2P1/1Kp4p/1b6/7b w - - ( | 9 | 9 | 26 | 16 | 10 | 3 | null | 5 | 2 | 0.3 | 40.025 | 49.615 | 27.590 | 1n4B1/2p2q1r/p1NPrp1p/K1B5/Q2Nk1Pp/1RP3P1/p3PP2 | 13 | 12 | 32 | 31 | 1 | 3 | 1994 | 3 | 1 | 0.32467532 |
| 50 | 2n5/2K2p2/p7/p1pk1n2/2R4Q/P2b1p1N/1B3N2/8 w - - 0 | 7 | 9 | 24 | 14 | 10 | 3 | 1881 | 2 | 1 | 0.47058824 | 139.183 | 140.642 | 21.459 | 6n1/1n2q2p/2QR1p2/r1PPp3/1p1Nk3/5R2/P4P1P/B4K1E | 12 | 9 | 33 | 24 | 9 | 3 | 1994 | 1 | 1 | 0.28767123 |
| 51 | 8/6K1/pN1Q4/1k2p3/3bN1R1/8/nPP5/7B w - - 0 1 | 8 | 5 | 25 | 8 | 17 | 3 | 1912 | 2 | 2 | 0.41935484 | 61.008 | 70.005 | 28.998 | 5b2/1N6/rpp3K1/5Q2/k1B1n3/p7/N1Pn4/8 w - - 0 1 | 6 | 8 | 19 | 17 | 2 | 3 | 1934 | 5 | 1 | 0.41176471 |
| 52 | k7/1r6/1P4b1/8/8/1R4K1/8/7B w - - 0 1 | 4 | 3 | 9 | 8 | 1 | 3 | 1965 | 4 | 4 | 0.63636364 | 67.265 | 72.229 | 24.964 | 8/8/6Qb/4k2r/8/nKN1p2P/2P1N1Pr/8 w - - 0 1 | 7 | 6 | 18 | 17 | 1 | 3 | 1927 | 2 | 1 | 0.37142857 |
| 53 | q2Q4/2p5/1bk1n3/4PB2/N1n2PK1/3p4/2N5/8 w - - 0 1 | 7 | 7 | 20 | 20 | 0 | 3 | 1880 | 5 | 2 | 0.35 | 171.087 | 170.593 | 19.506 | 2nr4/q1R5/1PN1N2Q/PBPk1P2/1p1r3B/5PP1/b1p1Pn2/b | 15 | 10 | 38 | 33 | 5 | 3 | 2004 | 5 | 2 | 0.26315789 |
| 54 | 8/3K1PBk/8/7P/8/4N3/8/8 w - - 0 1 | 5 | 1 | 8 | 0 | 8 | 3 | 1992 | 1 | 2 | 0.6 | 105.322 | 114.792 | 29.470 | 3Q1br1/1p1rPP1b/3P2p1/N1p3P1/3kpK2/2Np4/1P1n4/8 | 9 | 11 | 20 | 24 | 4 | 3 | 1941 | 1 | 2 | 0.27777778 |
| 55 | q4B2/4P1n1/4k3/1b2NrP1/1Q3P2/4K3/8/8 w - - 0 1 | 7 | 5 | 18 | 20 | 2 | 3 | 1855 | 5 | 2 | 0.33333333 | 105.039 | 110.070 | 25.031 | K1N2R2/6p1/2B1kb1p/1p6/p4P2/1p1pP2P/5Pp1/6Qb w | 9 | 10 | 24 | 13 | 11 | 3 | 1930 | 5 | 1 | 0.37254902 |
| 56 | 5r2/p2R1P1k/5p1p/8/q7/1p6/PP4QP/K7 w - - 0 1 | 7 | 7 | 18 | 18 | 0 | 3 | null | 5 | 4 | 0.41176471 | 19.062 | 18.217 | 17.155 | 6k1/1p1K2n1/8/6N1/rb3QN1/8/8/8 w - - 0 1 | 4 | 5 | 15 | 12 | 3 | 3 | 1889 | 5 | 2 | 0.47368421 |
| 57 | 1K3RBb/n3P2p/1Np1PPp1/1pP1k3/4p2P/2PpB1N1/b4r2/n | 13 | 12 | 32 | 23 | 9 | 3 | 1984 | 1 | 2 | 0.28735632 | 63.141 | 69.504 | 26.362 | QK5k/6Rp/3nb3/5p2/1p3p2/p7/8/BqR1r3 w - - 0 1 | 5 | 10 | 22 | 25 | 3 | 3 | 1955 | 5 | 4 | 0.42857143 |
| 58 | 3k3n/p1R5/8/1B2B3/4K3/8/8/8 w - - 0 1 | 4 | 3 | 11 | 4 | 7 | 3 | 1911 | 4 | 4 | 0.63636364 | 47.011 | 49.615 | 22.603 | 1R6/3P4/8/2Pk1PNB/8/B1K5/8/5N2 w - - 0 1 | 9 | 1 | 20 | 0 | 20 | 3 | 1901 | 2 | 2 | 0.625 |
| 59 | 8/5N2/2p5/8/3RPk2/NB5P/2P2K2/8 w - - 0 1 | 8 | 2 | 17 | 1 | 16 | 3 | 1954 | 4 | 2 | 0.55555556 | 34.056 | 49.667 | 33.612 | 1n6/3q2B1/3R3Q/2k1N3/6p1/1nN3K1/4P3/8 w - - 0 1 | 7 | 5 | 24 | 16 | 8 | 3 | null | 4 | 2 | 0.5 |
| 60 | 6B1/4pPq1/5k2/8/5P1r/8/8/1QK1R3 w - - 0 1 | 6 | 4 | 19 | 15 | 4 | 3 | 1994 | 4 | 5 | 0.38461538 | 58.027 | 62.711 | 24.684 | 4Nk2/3P1P2/1N4P1/8/6K1/8/8/8 w - - 0 1 | 6 | 1 | 9 | 0 | 9 | 3 | 1975 | 1 | 2 | 0.41176471 |

**APPENDIX C: The 90 DSNS-Generated Compositions Evaluated by the Human Experts**

| # | FEN | Solution | Source |
|---|---|---|---|
| 1 | 3K4/3b4/4p2k/8/8/4R3/8/1R6 w - - 0 1 | 1. Rg1 e5 2. Kxd7 e4 3. Rh3# 0-1 | tg2500+photo |
| 2 | 1b6/P7/k4r2/8/1P6/1KP5/2R1P3/8 w - - 0 1 | 1. axb8=Q Rb6 2. Qa8+ Kb5 3. c4# 0-1 | tg2500+photo |
| 3 | 2k5/7P/8/8/K7/b7/1p3Q2/8 w - - 0 1 | 1. Qf7 b1=Q 2. h8=Q+ Bf8 3. Qhxf8# 0-1 | tg1500+photo |
| 4 | 8/8/8/4Q3/2k5/1R3q2/8/1K6 w - - 0 1 | 1. Rxf3 Kb4 2. Rf4+ Kb3 3. Qe3# 0-1 | tg2500+photo |
| 5 | 8/6b1/P5R1/2B5/3R4/7K/4n3/6bk w - - 0 1 | 1. a7 Nf4+ 2. Rxf4 Bh2 3. a8=Q# 0-1 | tg2500+photo |
| 6 | 8/8/8/8/R1K5/1r6/8/1k6 w - - 0 1 | 1. Kxb3 Kc1 2. Rd4 Kb1 3. Rd1# 0-1 | tg1500+photo |
| 7 | 6B1/8/6KR/8/4k3/8/7Q/8 w - - 0 1 | 1. Qd2 Ke5 2. Rh5+ Ke4 3. Bd5# 0-1 | tg2500+photo |
| 8 | 1B6/3kP3/3P4/1P1p4/5R1K/5b2/8/8 w - - 0 1 | 1. Rf8 Bh5 2. Kxh5 d4 3. e8=Q# 0-1 | tg2500+photo |
| 9 | Q7/8/3Q4/8/1K2n3/8/5k2/8 w - - 0 1 | 1. Qxe4 Kf1 2. Qd1+ Kf2 3. Qde1# 0-1 | tg1500+photo |
| 10 | 8/6B1/3n4/5B1K/3R4/pp6/k7/8 w - - 0 1 | 1. Rd1 b2 2. Be6+ Nc4 3. Bxc4# 0-1 | tg1500+photo |
| 11 | 8/1K6/8/8/8/k1N5/6R1/2R5 w - - 0 1 | 1. Nd1 Kb3 2. Rb2+ Ka3 3. Ra1# 0-1 | comp3.5 |
| 12 | 8/k2P4/8/3N4/8/1n4K1/8/8 w - - 0 1 | 1. d8=Q Na5 2. Qb6+ Ka8 3. Nc7# 0-1 | tg2500+photo |
| 13 | 8/1Q6/7K/3B4/8/6k1/8/8 w - - 0 1 | 1. Qf7 Kh3 2. Qf2 Kg4 3. Be6# 0-1 | comp3.5 |
| 14 | kN6/8/K7/4N3/8/8/3R4/b7 w - - 0 1 | 1. Rd8 Bxe5 2. Nc6+ Bb8 3. Rxb8# 0-1 | tg1500+photo |
| 15 | 4k3/bQ6/8/6K1/8/6R1/8/8 w - - 0 1 | 1. Qxa7 Kf8 2. Re3 Kg8 3. Re8# 0-1 | tg1500+photo |
| 16 | 8/8/8/8/4K3/1Q5n/8/k7 w - - 0 1 | 1. Kd3 Nf4+ 2. Kc2 Nd5 3. Qb1# 0-1 | comp3.5 |
| 17 | 8/4K3/5Nk1/6B1/6r1/8/8/6R1 w - - 0 1 | 1. Rxg4 Kg7 2. Bf4+ Kh8 3. Rg8# 0-1 | tg1500+photo |
| 18 | 1Rn5/8/4K3/8/8/kp6/3B4/2R5 w - - 0 1 | 1. Bc3 b2 2. Bxb2+ Ka4 3. Ra1# 0-1 | comp3.5 |
| 19 | 8/P3k1P1/1p3R1p/8/p4K2/8/4p3/8 w - - 0 1 | 1. g8=Q h5 2. Qg7+ Ke8 3. a8=Q# 0-1 | tg1500+photo |
| 20 | 1k6/1P1R4/4p3/1K6/8/4n3/8/8 w - - 0 1 | 1. Kc6 e5 2. Rd8+ Ka7 3. Ra8# 0-1 | tg2500+photo |
| 21 | 7B/2K4P/8/7k/8/8/6R1/7n w - - 0 1 | 1. Bf6 Ng3 2. Rxg3 Kh6 3. h8=Q# 0-1 | tg1500+photo |
| 22 | 5Nk1/8/3R4/8/8/8/K7/2R5 w - - 0 1 | 1. Rd7 Kh8 2. Ng6+ Kg8 3. Rc8# 0-1 | tg1500+photo |
| 23 | 8/8/3K4/7B/8/4p2k/8/1Q6 w - - 0 1 | 1. Qg1 e2 2. Bxe2 Kh4 3. Qg4# 0-1 | tg2500+photo |
| 24 | 8/8/K1B2Q2/8/1k6/8/5b2/2N5 w - - 0 1 | 1. Bb5 Bg3 2. Nd3+ Kb3 3. Qb2# 0-1 | comp3.5 |
| 25 | 5n2/4K1Pb/1R6/8/8/k7/2p5/8 w - - 0 1 | 1. gxf8=Q Be4 2. Qf1 c1=Q 3. Qa6# 0-1 | comp3.5 |
| 26 | k7/8/1P6/P2R4/2R5/3rN3/8/6K1 w - - 0 1 | 1. a6 Rd1+ 2. Rxd1 Kb8 3. Rd8# 0-1 | comp3.5 |
| 27 | 8/7R/2p5/5K2/8/7p/1p1R4/7k w - - 0 1 | 1. Rb7 h2 2. Rbxb2 c5 3. Rb1# 0-1 | comp3.5 |
| 28 | 5k2/2P5/8/8/5K2/8/8/3Rn3 w - - 0 1 | 1. Rd7 Nd3+ 2. Kf5 Nb4 3. c8=Q# 0-1 | tg1500+photo |
| 29 | 2k5/K3Pb2/2P5/8/p1r2b2/8/5Q2/4R3 w - - 0 1 | 1. Qb6 Be3 2. Rxe3 a3 3. Qd8# 0-1 | tg1500+photo |
| 30 | 8/1PK5/k5q1/4N3/bP6/8/3N4/8 w - - 0 1 | 1. Nxg6 Bb5 2. b8=Q Bc6 3. Qb6# 0-1 | tg2500+photo |
| 31 | 8/2R5/1P6/6K1/1k1p4/6p1/8/8 w - - 0 1 | 1. b7 g2 2. b8=Q+ Ka4 3. Ra7# 0-1 | comp3.5 |
| 32 | 4k1B1/8/2KB4/6N1/6n1/8/8/8 w - - 0 1 | 1. Kc7 Ne5 2. Ne6 Nc6 3. Ng7# 0-1 | comp3.5 |
| 33 | 8/3P2kr/8/K1R4p/8/8/8/4R3 w - - 0 1 | 1. d8=Q h4 2. Qd7+ Kg8 3. Re8# 0-1 | tg1500+photo |
| 34 | 6kN/4R3/4K3/8/8/8/8/8 w - - 0 1 | 1. Kf6 Kxh8 2. Kg6 Kg8 3. Re8# 0-1 | tg1500+photo |
| 35 | 1K3k2/1P6/4PNBP/8/8/4N3/r7/7n w - - 0 1 | 1. Nf5 Ra8+ 2. bxa8=Q Nf2 3. e7# 0-1 | comp3.5 |
| 36 | 8/8/1R6/4k3/2K5/8/2QP4/8 w - - 0 1 | 1. Qg6 Kf4 2. Rf6+ Ke5 3. d4# 0-1 | tg2500+photo |
| 37 | 8/3P4/2k1K3/p7/1R6/2P5/8/8 w - - 0 1 | 1. Rb8 a4 2. d8=Q a3 3. Qd6# 0-1 | comp3.5 |
| 38 | 8/2B3P1/8/7k/8/8/7K/8 w - - 0 1 | 1. g8=Q Kh6 2. Bf4+ Kh5 3. Qg5# 0-1 | tg1500+photo |
| 39 | 8/1R2K3/6bR/8/k7/8/8/8 w - - 0 1 | 1. Rxg6 Ka5 2. Rh6 Ka4 3. Ra6# 0-1 | comp3.5 |
| 40 | 4K3/5R2/8/4b2k/6R1/8/3B4/8 w - - 0 1 | 1. Rg8 Bg7 2. Rfxg7 Kh4 3. Rh8# 0-1 | tg2500+photo |
| 41 | 8/1R6/1q6/8/8/8/K1R5/3N3k w - - 0 1 | 1. Rxb6 Kg1 2. Nc3 Kh1 3. Rb1# 0-1 | tg2500+photo |
| 42 | 8/2P5/8/8/8/8/k1K5/8 w - - 0 1 | 1. c8=Q Ka3 2. Qg4 Ka2 3. Qa4# 0-1 | tg2500+photo |
| 43 | 8/8/1B5R/4K2B/8/8/8/4k3 w - - 0 1 | 1. Be3 Kf1 2. Bf3 Ke1 3. Rh1# 0-1 | comp3.5 |
| 44 | 7R/1p6/p7/8/1b2R2p/5K2/5p2/5k2 w - - 0 1 | 1. Rhxh4 Kg1 2. Reg4+ Kf1 3. Rh1# 0-1 | tg1500+photo |
| 45 | 7k/8/8/3B2K1/4p3/8/8/6B1 w - - 0 1 | 1. Kg6 e3 2. Bh2 e2 3. Be5# 0-1 | tg1500+photo |
| 46 | B3k3/8/4KP2/4B3/8/8/8/8 w - - 0 1 | 1. Bd6 Kd8 2. f7 Kc8 3. f8=Q# 0-1 | tg2500+photo |
| 47 | 1K6/8/8/8/1k2N3/8/Q7/1B6 w - - 0 1 | 1. Bc2 Kb5 2. Qa7 Kb4 3. Qa4# 0-1 | tg2500+photo |
| 48 | 8/4P2n/2K5/8/7k/5R2/6R1/8 w - - 0 1 | 1. e8=Q Nf8 2. Qd8+ Kh5 3. Qg5# 0-1 | comp3.5 |
| 49 | 8/R4P2/4k3/1B6/6p1/4K3/8/8 w - - 0 1 | 1. f8=Q g3 2. Bc4+ Ke5 3. Qf4# 0-1 | tg1500+photo |
| 50 | 8/7P/8/4pR2/8/p6K/8/6k1 w - - 0 1 | 1. h8=Q a2 2. Qg8+ Kh1 3. Qg2# 0-1 | tg1500+photo |
| 51 | 8/K2N3P/8/6k1/3N2P1/5nPP/1n6/8 w - - 0 1 | 1. h8=Q Kg6 2. Qg8+ Kh6 3. Nf5# 0-1 | comp3.5 |
| 52 | 8/4BP2/2N5/3k4/4p3/2K5/8/8 w - - 0 1 | 1. Nd4 e3 2. f8=Q e2 3. Qf5# 0-1 | tg2500+photo |
| 53 | 8/8/6R1/3k4/8/8/RR6/2K5 w - - 0 1 | 1. Rgb6 Ke5 2. R2b4 Kf5 3. Ra5# 0-1 | tg2500+photo |
| 54 | 2R5/3P4/8/1k6/8/1p1K4/8/R5b1 w - - 0 1 | 1. d8=Q Bb6 2. Qe8+ Kb4 3. Qa4# 0-1 | comp3.5 |
| 55 | 8/8/3Qr3/kB1n4/8/N7/1K6/8 w - - 0 1 | 1. Qc5 Re2+ 2. Bxe2+ Ka4 3. Bd1# 0-1 | tg2500+photo |
| 56 | 1n6/P1K5/5Q2/8/7B/8/4k3/8 w - - 0 1 | 1. axb8=Q Kd2 2. Qb3 Ke2 3. Qf2# 0-1 | tg2500+photo |
| 57 | 8/8/6p1/7k/8/8/2Q3n1/2BK3N w - - 0 1 | 1. Qxg2 Kh4 2. Qg3+ Kh5 3. Qh3# 0-1 | comp3.5 |
| 58 | 8/6Q1/8/8/8/7B/1K6/4k3 w - - 0 1 | 1. Qg2 Kd1 2. Bg4+ Ke1 3. Qe2# 0-1 | tg2500+photo |
| 59 | 3K4/8/3k4/6R1/4R2b/8/8/8 w - - 0 1 | 1. Rxh4 Ke6 2. Rf4 Kd6 3. Rf6# 0-1 | tg2500+photo |
| 60 | 8/5P1k/1Rr5/4R3/2P5/3p4/8/5K2 w - - 0 1 | 1. f8=Q Rf6+ 2. Rxf6 d2 3. Re7# 0-1 | tg1500+photo |
| 61 | 5Nk1/1p6/8/8/8/8/8/K3Q3 w - - 0 1 | 1. Qe7 b6 2. Ne6 b5 3. Qg7# 0-1 | tg2500+photo |
| 62 | 8/6Q1/8/1K1k4/8/8/8/6N1 w - - 0 1 | 1. Qf6 Ke4 2. Kc4 Ke3 3. Qd4# 0-1 | tg1500+photo |
| 63 | 5N2/6R1/5k1N/3Pp3/8/8/1K6/8 w - - 0 1 | 1. Ne6 e4 2. Ng4+ Kf5 3. Rg5# 0-1 | tg1500+photo |
| 64 | 1k6/4p3/8/8/RK6/8/8/6BB w - - 0 1 | 1. Bb6 Kc8 2. Bc6 e6 3. Ra8# 0-1 | comp3.5 |
| 65 | 4K3/1P4p1/R4b2/2bk4/5RN1/8/8/8 w - - 0 1 | 1. b8=Q Bce7 2. Qa8+ Kc5 3. Qc6# 0-1 | tg1500+photo |
| 66 | 8/8/8/2p1p3/8/k7/2B5/1K2R3 w - - 0 1 | 1. Re4 c4 2. Rxc4 e4 3. Ra4# 0-1 | comp3.5 |
| 67 | 8/4N3/1Q6/1b6/8/2K2B1k/8/8 w - - 0 1 | 1. Nf5 Bc6 2. Qb8 Bd7 3. Qg3# 0-1 | comp3.5 |
| 68 | k7/2P5/3K4/8/8/8/8/8 w - - 0 1 | 1. Kc6 Ka7 2. c8=R Ka6 3. Ra8# 0-1 | comp3.5 |
| 69 | 8/8/8/8/k4nK1/2P5/2P3Q1/8 w - - 0 1 | 1. Qb7 Ka5 2. Kxf4 Ka4 3. Qb4# 0-1 | tg2500+photo |
| 70 | 5K1k/8/1B6/8/8/1p6/4n3/2R3N1 w - - 0 1 | 1. Rc7 b2 2. Nxe2 b1=Q 3. Bd4# 0-1 | tg1500+photo |
| 71 | 8/3P1N2/3NN3/3k1p1p/3B4/8/8/2K5 w - - 0 1 | 1. d8=Q h4 2. Qc8 h3 3. Qc4# 0-1 | comp3.5 |
| 72 | 4N3/3K4/4N3/4k3/4B3/8/2p2P2/3N4 w - - 0 1 | 1. Bxc2 Kd5 2. Ne3+ Ke5 3. f4# 0-1 | tg1500+photo |
| 73 | 2k5/8/4P3/4B3/4R3/6K1/8/8 w - - 0 1 | 1. Rb4 Kd8 2. Bd6 Ke8 3. Rb8# 0-1 | tg1500+photo |
| 74 | 8/3K2R1/8/3Q4/8/8/5k2/2n5 w - - 0 1 | 1. Qe4 Na2 2. Rg2+ Kf1 3. Qe2# 0-1 | comp3.5 |
| 75 | 8/3N4/6Q1/8/2k5/B5K1/8/8 w - - 0 1 | 1. Nc5 Kd4 2. Qd6+ Kc4 3. Qd3# 0-1 | comp3.5 |
| 76 | 8/8/8/5r1Q/8/1k6/8/R1K5 w - - 0 1 | 1. Qxf5 Kc3 2. Ra4 Kb3 3. Qc2# 0-1 | tg2500+photo |
| 77 | 8/8/3Rr3/8/1R6/5k2/8/7K w - - 0 1 | 1. Rxe6 Kg3 2. Rf6 Kh3 3. Rf3# 0-1 | tg2500+photo |
| 78 | 1b5b/8/Q7/7p/8/4p3/K7/2k1b3 w - - 0 1 | 1. Qd3 h4 2. Kb3 h3 3. Qc2# 0-1 | tg1500+photo |
| 79 | 8/3P4/8/8/8/k7/B2B1pB1/1K6 w - - 0 1 | 1. d8=Q f1=Q+ 2. Bxf1 Ka4 3. Qa5# 0-1 | tg2500+photo |
| 80 | 5n2/7P/8/B7/2N5/8/8/1k3K2 w - - 0 1 | 1. h8=Q Kc1 2. Qc3+ Kd1 3. Qd2# 0-1 | comp3.5 |
| 81 | 2K5/k3N3/8/1B6/8/8/p7/8 w - - 0 1 | 1. Kc7 a1=Q 2. Nc8+ Ka8 3. Bc6# 0-1 | comp3.5 |
| 82 | 4K3/8/8/8/Q3B3/8/7k/8 w - - 0 1 | 1. Qa3 Kg1 2. Qg3+ Kf1 3. Bd3# 0-1 | tg2500+photo |
| 83 | 7k/8/n4K2/8/8/r1p5/8/2R5 w - - 0 1 | 1. Kf7 c2 2. Rh1+ Rh3 3. Rxh3# 0-1 | comp3.5 |
| 84 | 3k1K2/8/4P3/1B6/1R6/3r4/8/1b3R2 w - - 0 1 | 1. Rc4 Rf3+ 2. Rxf3 Bc2 3. e7# 0-1 | comp3.5 |
| 85 | 5R2/6N1/8/5K2/7k/1b6/6p1/8 w - - 0 1 | 1. Kf4 Bf7 2. Rh8+ Bh5 3. Rxh5# 0-1 | tg1500+photo |
| 86 | 7N/P7/1B1k4/7K/3N4/8/8/8 w - - 0 1 | 1. a8=Q Ke5 2. Ng6+ Kf6 3. Qf8# 0-1 | comp3.5 |
| 87 | 4R3/K2R1p2/8/8/8/8/8/6Nk w - - 0 1 | 1. Rg8 Kh2 2. Rxf7 Kh1 3. Rh7# 0-1 | tg2500+photo |
| 88 | 4B3/5Pk1/8/7P/7B/8/5b2/3K4 w - - 0 1 | 1. Be7 Bg3 2. f8=Q+ Kh7 3. Bg6# 0-1 | tg1500+photo |
| 89 | 8/8/5Q2/8/k3p3/8/K7/2N5 w - - 0 1 | 1. Qb6 e3 2. Nb3 e2 3. Nc5# 0-1 | tg1500+photo |
| 90 | 8/P7/8/8/2k5/2b5/2R5/1K1Q4 w - - 0 1 | 1. a8=Q Kb4 2. Qb7+ Kc4 3. Qdd5# 0-1 | tg2500+photo |

## APPENDIX D: The Human Expert Evaluations of the 90 DSNS-Generated Compositions

| # | Jana Krivec | Vlaicu Crisan | Matej Guid | Average |
|---|---|---|---|---|
| 1 | 0.0 | 0.5 | 0.5 | 0.33 |
| 2 | 1.0 | 0.1 | 1.0 | 0.70 |
| 3 | 1.0 | 0.0 | 0.0 | 0.33 |
| 4 | 0.0 | 0.1 | 0.1 | 0.07 |
| 5 | 2.0 | 0.1 | 4.0 | 2.03 |
| 6 | 2.0 | 0.3 | 0.1 | 0.80 |
| 7 | 2.0 | 0.0 | 3.0 | 1.67 |
| 8 | 1.0 | 0.2 | 0.1 | 0.43 |
| 9 | 1.0 | 0.1 | 0.2 | 0.43 |
| 10 | 2.0 | 0.2 | 0.2 | 0.80 |
| 11 | 2.0 | 0.5 | 3.0 | 1.83 |
| 12 | 1.0 | 0.2 | 0.0 | 0.40 |
| 13 | 1.0 | 0.3 | 3.0 | 1.43 |
| 14 | 1.0 | 0.1 | 1.0 | 0.70 |
| 15 | 1.0 | 0.0 | 0.0 | 0.33 |
| 16 | 0.0 | 0.2 | 0.1 | 0.10 |
| 17 | 1.0 | 0.1 | 0.2 | 0.43 |
| 18 | 1.0 | 0.1 | 0.2 | 0.43 |
| 19 | 1.0 | 0.1 | 0.2 | 0.43 |
| 20 | 2.0 | 0.3 | 0.2 | 0.83 |
| 21 | 1.0 | 0.0 | 0.0 | 0.33 |
| 22 | 2.0 | 0.0 | 0.5 | 0.83 |
| 23 | 0.0 | 0.2 | 0.1 | 0.10 |
| 24 | 1.0 | 1.0 | 3.0 | 1.67 |
| 25 | 2.0 | 0.2 | 2.0 | 1.40 |
| 26 | 2.0 | 0.1 | 0.3 | 0.80 |
| 27 | 0.0 | 0.0 | 0.0 | 0.00 |
| 28 | 0.0 | 0.1 | 0.2 | 0.10 |
| 29 | 2.0 | 0.2 | 1.5 | 1.23 |
| 30 | 0.0 | 0.2 | 0.0 | 0.07 |
| 31 | 0.0 | 0.3 | 0.0 | 0.10 |
| 32 | 2.0 | 0.1 | 2.0 | 1.37 |
| 33 | 0.0 | 0.0 | 0.0 | 0.00 |
| 34 | 2.0 | 2.0 | 0.2 | 1.40 |
| 35 | 2.0 | 0.1 | 0.2 | 0.77 |
| 36 | 2.0 | 0.1 | 3.0 | 1.70 |
| 37 | 1.0 | 0.1 | 0.2 | 0.43 |
| 38 | 0.0 | 0.1 | 0.0 | 0.03 |
| 39 | 2.0 | 0.0 | 0.0 | 0.67 |
| 40 | 1.0 | 0.2 | 0.3 | 0.50 |
| 41 | 0.0 | 0.1 | 0.0 | 0.03 |
| 42 | 2.0 | 0.0 | 0.2 | 0.73 |
| 43 | 1.0 | 0.0 | 0.2 | 0.40 |
| 44 | 1.0 | 0.2 | 0.0 | 0.40 |
| 45 | 2.0 | 0.0 | 0.2 | 0.73 |
| 46 | 1.0 | 0.2 | 0.4 | 0.53 |
| 47 | 3.0 | 0.0 | 0.8 | 1.27 |
| 48 | 0.0 | 0.1 | 0.2 | 0.10 |
| 49 | 0.0 | 0.1 | 0.3 | 0.13 |
| 50 | 1.0 | 0.0 | 0.0 | 0.33 |
| 51 | 1.0 | 0.3 | 0.3 | 0.53 |
| 52 | 2.0 | 0.3 | 1.0 | 1.10 |
| 53 | 3.0 | 0.0 | 0.1 | 1.03 |
| 54 | 0.0 | 0.1 | 0.6 | 0.23 |
| 55 | 0.0 | 0.2 | 3.0 | 1.07 |
| 56 | 0.0 | 0.2 | 0.1 | 0.10 |
| 57 | 1.0 | 0.1 | 0.1 | 0.40 |
| 58 | 1.0 | 0.1 | 0.1 | 0.40 |
| 59 | 2.0 | 2.0 | 0.1 | 1.37 |
| 60 | 1.0 | 0.3 | 0.3 | 0.53 |
| 61 | 1.0 | 0.3 | 0.5 | 0.60 |
| 62 | 2.0 | 0.4 | 3.0 | 1.80 |
| 63 | 2.0 | 0.4 | 2.0 | 1.47 |
| 64 | 2.0 | 0.2 | 0.1 | 0.77 |
| 65 | 2.0 | 0.1 | 0.3 | 0.80 |
| 66 | 1.0 | 1.0 | 0.1 | 0.70 |
| 67 | 1.0 | 0.0 | 0.6 | 0.53 |
| 68 | 1.0 | 1.0 | 1.0 | 1.00 |
| 69 | 3.0 | 0.3 | 0.1 | 1.13 |
| 70 | 3.0 | 0.3 | 2.0 | 1.77 |
| 71 | 2.0 | 0.0 | 3.0 | 1.67 |
| 72 | 2.0 | 0.3 | 1.5 | 1.27 |
| 73 | 3.0 | 0.2 | 0.2 | 1.13 |
| 74 | 3.0 | 0.3 | 0.2 | 1.17 |
| 75 | 3.0 | 0.1 | 3.0 | 2.03 |
| 76 | 2.0 | 0.2 | 0.2 | 0.80 |
| 77 | 3.0 | 0.2 | 0.0 | 1.07 |
| 78 | 3.0 | 0.2 | 0.5 | 1.23 |
| 79 | 2.0 | 0.0 | 0.1 | 0.70 |
| 80 | 2.0 | 0.1 | 0.1 | 0.73 |
| 81 | 4.0 | 0.3 | 2.0 | 2.10 |
| 82 | 3.0 | 0.0 | 0.3 | 1.10 |
| 83 | 1.0 | 0.3 | 0.0 | 0.43 |
| 84 | 2.0 | 0.3 | 0.2 | 0.83 |
| 85 | 2.0 | 0.3 | 0.1 | 0.80 |
| 86 | 2.0 | 0.2 | 0.1 | 0.77 |
| 87 | 2.0 | 0.0 | 0.1 | 0.70 |
| 88 | 1.0 | 0.3 | 0.5 | 0.60 |
| 89 | 1.0 | 0.1 | 0.3 | 0.47 |
| 90 | 2.0 | 0.0 | 0.1 | 0.70 |

## APPENDIX E: Unedited Expert Commentary on the DSNS-Generated Problems.

*Note that these comments were provided with no knowledge about the actual composer of the problems. The experts would have had to guess whether they were human or computer-generated. The parts in bold refer to the additional commentary the expert provided for the compositions he identified as most likely having been composed by a human. In the move notation 'S' and 'N' both refer to the knight.*

| # | FM (C) / IM (S) Vlaicu Crisan | FM Matej Guid |
|---|---|---|
| 1 | Very weak key takes three flights. Short threat. Two variations ending with grab theme and same mate as in the threat. No duals in the real play a plus. | obvious, plain |
| 2 | Awful key: major promotion from en prise position capturing a black officer with short threat and taking all four flights. wRc2 used only as threat. Dual in the variation 3.Qa4#. | slightly pretty, but with obvious moves **there are many pieces and pawns on board (harder to design a problem automatically), perception of beauty is relatively humanlike (although this could be very subjective)..** |
| 3 | Double solution: 1.Qa7! ~ 2.h8=Q/R[+] ~ 3.Q[R]xf8#. The key takes three flights to the black King. The threat is unstoppable. | plain |
| 4 | Weak give-and-take key: en prise rook captures the black Queen (last officer). The dualistic mate 3.Qb2# ruins the intended beauty of a mirror ideal mate. | most straightforward |
| 5 | Short dualistic threat after the key. Three useless pieces (wRd4, wBc5 and bBbg7) - they could be easily replaced by white pawns d4 and e3. Poor construction. | fairly difficult to find (there are reasonable alternatives at disposal)... also beautiful in the sense that a quite march forward by the pawn results in an effective check(mate) on the long diagonal **this one was fairly complex, with a nice motif.. I also gave it the highest aesthetic rating... most computer-like compositions are far more simple (also in a computational sense)** |
| 6 | Bad key, capturing a whole Rook (last black piece) and taking two flights. Poor construction: bR can be replaced by bP. A second variation can be added easily (Ke3, Re1 - Kh2, pe4). **The wR could actually stand anywhere from a4 to a8. I think a4 would be the preferred choice of a human, because of the figurative initial setting.** | very obvious moves, hard to miss any one of them |
| 7 | Double solution: 1.Qg3! Kd4 2.Rh5 Ke4 3.Rh4#. Dual in the main variation 1.Qd2 Ke5 2.Rh3 Ke4 3.Qe3#. No dual in the second variation 1.Qd2 Kf3 2.Rh3+ Kg4(Ke4) 3.Be6(Qe3)#. | nice geometric position of the pieces a the end... but the fact that the black king is the only black piece makes it easier to find the solution... still: pretty **it is possible that this one was designed by a computer... I find the geometry really pretty, and for this reason it seems to me that it is more likely that a human was the composer** |
| 8 | Weak key, taking two flights of which one is provided in the set play (e8) and short threat. No real fight: black must sacrifice its Bishop to stop the immediate mate. No duals. | the strongest - and very obvious - continuation wins, which makes it less beautiful |
| 9 | Obvious key, capturing the remaining black piece and taking five flights. Duals after 1...Kf1 2.Qd2/Qg6/Qc2/Qf3. Usage of two white Queens rather dubious. | not capturing the knight would be beautiful, the solution involving capturing it is not |
| 10 | Key takes two flights and threatens two short mates. One variation only, with a line closing and line opening, ending with capture. Fortunately, no duals. | possible interposition of the black knight makes the solution less beautiful |
| 11 | Ampliative though pretty obvious key, giving an extra flight. However, all three black moves have the same continuation (2.Rb2 and 3.Ra1#) - so no real actual fight. | a strange and somewhat unexpected white's knight move makes the solution beautiful, in particular since it is the only way to deliver checkmate in time **this 1.Nd1 is very appealing and unexpected... due to the small number of pieces it is quite possible that the computer designed it... but would computer know to appreciate the beauty of this move?** |
| 12 | Promotion key takes two flights. Long threat is actually unstoppable. bS defense only stops a second similar threat (2.Qc7+ Ka8/Ka6 3.Sb6/Qb6#). | a very obvious solution |
| 13 | Obvious key takes two prominent flights. Dual after 1...Kh2 2.Qf2/Qf3. Compare with Shinkman's composition (Ke4, Qb5, Ba5 - Kc8): 1.Qb2 Kd7 2.Qe5 Kc8(Kc6) 3.Qc7(Qd5)# | nice geometry... the fact that this is the only way to deliver checkmate in time makes it quite pretty |
| 14 | Key takes bK flight, creates a strong unstoppable battery and threatens a dualistic mate (2.Sc6/Sd7) which Black defense can not actually prevent. wSe5 useless. No real fight. | sacrificing of the knight in order to enable a checkmate to the "determined" rook is not that plain... |
| 15 | Three double solutions: 1.Kf5/Kh5/Kf6 all threatening 2.Rg8#. 1...Bc5 or 1...Kf8 can both be answered by 2.Rg7 followed by 3.Qf7#. Brutal key capturing the last black piece. | an obvious route to checkmate |
| 16 | Dualistic mate 3.Qa4/Qa3/Qb2# are also possible. No real challenge for a player: the mate is possible only by approaching wK. Black has no defense against the threat. | plain march by the white knight... not beautiful |
| 17 | Duals in the main variation: 2.Bc1/Bd2/Be3+. Extra variation is better: 1...Kf5 2.Be3 Ke5 3.Rg5#. Bad key capturing last black officer, taking a flight and setting up a strong battery. | the other possible solution would be quite pretty: 1... Kf5 2. Be3 Ke5 3. Rg5# (but the actual continuation is not) |
| 18 | Dual in the main variation: 2.Rxb2 also works, followed by 3.Ra1#. The key takes a flight and threatens a short mate, which black can only delay. No duals with wKe7, bSc5 and wBa1. | not taking the en-prise knight is ok, but still very plain |
| 19 | Major promotion key takes two flights. Dualistic threat 2.Qf7+ also works. A secondary variation adds interest: 1...Kxf6 2.a8=Q Ke7 3.Qad8# | nice to see the new queen promoted... but the rook move would deliver the checkmate as well... not really beautiful |
| 20 | Black pieces are rather superfluous - even without them, the same solution would appear (1.Kb6?? Stalemate). Solution is however obvious, because no real fight.<br>**A computer might have seen that bSe3 and bPe6 are useless. A human would probably add them just in order to equilibrate the balance of initial forces.** | too obvious |
| 21 | Double solution: 1.Be5! Sg3 2.Rxg3 Kh4/Kh6 3.h8=Q[R]#. Bad key taking a flight and threatening a short mate. | obvious, besides 1.Be5 is another solution |
| 22 | Double solution: 1.Rc7! Kh8 2.Rd8 Kg8 3.Se6/Sg6/Sh7#. Dual in the intention after 1...Kh8 2.Se6/Sg6+/Rc8. Bad key taking two flights. | leaving the knight en prise is somewhat beautiful |
| 23 | Obvious key takes three flights. It would have been better to place bP on e4 in order to set up a continuous zugzwang: 1.Qg1! Zz 1...e3 2.Be2! Zz 2...Kh4 3.Qg4#.<br>**Human are perhaps more likely to create zugzwang based compositions.** | too obvious |
| 24 | Key takes flight c4 and threatens the variation given in the solution 2.Sd3+ Kb3/Ka3 3.Qb2#. Also particularly interesting is the second variation: 1...Bd4 2.Qxd4+ Ka3 3.Qc3#.<br>**Neat changed mate after 2...Ka3 in two variations is particularly pleasing.** | nice geometry... also the only way to deliver checkmate in time... quite pretty<br>**the geometry is really not trivial... humans know to appreciate it... the quality seems to be much higher that in many of these 90 problems** |
| 25 | Key is bad, capturing a black piece that can give check and introduces a short threat. In the main variation, black second move can stop only one of the threatened mates. No duals. | nice geometric moves by the newly promoting queen... long moves... |
| 26 | Dual after 1...Rd1+ 2.Sxd1. Flight taking key. wSe3 could be removed, by placing for instance wK on g2 and wRc4 on c5. | very obvious... not taking the rook is ok.. but still... |
| 27 | Double solution: 1.Rxb2! ~ 2.Ra7/Rd7 ~ 3.Ra1/Rd1#. Dual in the intention after 2...c5 3.Rd1#. The threat is unstoppable - no real fight by Black. | not unexpected at all |
| 28 | Duals after 1...Sd3+ 2.Ke3/Kg3/Kg5. Key takes three flights and threatens a short mate. Poorly constructed position, with duals in the second variation: 1...Sg2+ 2.Ke5/Kg5/Kg3/Ke4. | ignoring the en-prise knight... still, too straightforward |
| 29 | Flight taking key, threatening a short mate. Black defense pins wQ, but the pin is released through a brutal capture of pinner. Duals: 2...a3 3.Qb7/Qb8#. Nothing subtle. | actually the only way by far... that's quite nice, but still not very beautiful<br>**lots of pieces and pawns makes it more difficult for the computers to compose...** |
| 30 | Another horrible key, taking the most prominent black figure (bQ). 2...Bc6 ensures that 3.Qb7# is avoided. wSs can be traded for wPe2, when bQ is replaced on d3: 1.exd3! | some checkmate combination with an underpromotion to a knight would be beautiful, capturing the black queen on the first move isn't |
| 31 | Straightforward play, with no surprises. There is a dual 2.b8=R which occurs after 1...g2. No need to have black pawns on the board.<br>**Black pawns d4 and g3 are seemingly useless.** | very plain |
| 32 | Multiple duals in intention after 1...Se5 2.Bb3/Bd5/Be6/Ba3/Bb4/Bc5/Kc8. The flight taking key is not pleasant. | the helpless black knight, a unique solution... |
| 33 | Double solution: 1.Rc6! with the following variations: 1...Rh8/Rh6 2.Re7+ Kf8 3.d8=Q# and 1...Kf8/Kg8/Kf7 2.d8=Q[+] ~ 3.Re7#. Bad key taking three flights. | the most obvious solution |
| 34 | The second variation 1...Kf8 2.Sg6+ Kg8 3.Rg7# enhances the content of this composition. The key is not bad and a better one can not be found. Best from the lot.<br>**This is clearly the best of the lot. A human composer would even consider publishing this composition!** | too simple |
| 35 | Duals: 2.bxa8=R/B/S/Kxa8. The second variation 1...Re2 2.h7 adds some interest. The key takes an unprovided flight and creates a short threat. | 1... Re2 2. h7 with an overloaded black rook would be slightly more pretty... just giving the rook away is not |
| 36 | Duals: 3.Qf5# and 2.Kd3 Kf3(Ke5) 3.Rf6(Qe4/Qg5)#. With same material, Healey created a memorable composition (Kd6, Qf2, Rc5, pc3 - Kd3): 1.Kd7 Ke4 2.Rd5 Kxd5 3.Qd4#. | the pawn unexpectedly joins in the checkmating construction.. also the only way for #3... quite nice **the role of the pawn in this checkmate in terms of making this composition appealing has to be appreciated... seems humanlike perception of beauty** |
| 37 | Duals in the main variation: 3.Qb6/Qc8# and 2.d8=R ~ 3.Rdc8#. The key is not very good, because wR is en prise in the initial position. | too obvious |
| 38 | Dual: 2.Bd8 (instead of 2.Bf4+), avoidable by shifting wBc7 to b8. Again a major promotion key taking three flights. Second variation 1...Kh4 2.Bf4/Bd8 Kh5 3.Qg5#. | too obvious |
| 39 | Double solutions: 1.Rh1! Bb1 2.Rxb1 Ka5/Ka3 3.Ra1# and 1.Rh8! Be8 2.Rxe8 Ka5/Ka3 3.Ra8#. Duals in the intention after 1...Ka5 any waiting move e.g. 2.Kf7/Rb8/Rc6 etc. will mate. | too obvious |
| 40 | Dual after 2...Kh4 3.Rh7#. Give and take key introducing a short threat, with wR playing from en prise position. | 1. Rg8 is not the first move that comes too mind.. that slightly contributes to the aesthetic value |
| 41 | Multiple duals: 2.Rh6/Rf6/Sc3/Se3. The key is awful, capturing the only black unit left on the board (bQ). Everything is very obvious. | too straightforward |
| 42 | Double solution: 1.c8=R! Ka3 2.Rc4 Ka2 3.Ra4#. Nice stalemate avoidance in the intention 2.Qc4?? **In spite of the double solution, the stalemate 2.Qc4 would appeal much to humans, as it reminds the famous Barbieri-Saavedra endgame.** | underpromotion to a rook would be more pretty... 2. Qg4 is not too unexpected |
| 43 | Double solution: 1.Rd6! Kf1 2.Rd2/Bf3 Ke1 3.Rd1#. Dual in the intention as well: 2.Rg6 Ke1 3.Rg1#. Bad key taking an unprovided flight. | too obvious |
| 44 | Capturing key, creating a short threat. The defense is unique, but the continuation is rather dull. With wR placed on g4 and bPs a6, b7 and h4 removed, the key would have been improved. **Black pawns a6 and b7 are seemingly useless. As stated in the commentary, the position could have been improved.** | an obvious choice |
| 45 | Double solution: 1.Kh6! e3 2.Kg6/Ba2-f7[Bh2] e2 3.Bd4[Be5]#. Similar duals in the intention: 2.Kh6/Ba2-f7. Bad key taking an unprovided flight. | most obvious |
| 46 | Dual: 3.f8=R#. Flight taking key. The play could have been improved by adding a twin (e.g. Shift bK to h8 with the solution: 1.Be4 Kg8 2.Ke7 Kh8 3.f7#). | there is only one possible #3 solution... and that is slightly surprising.. but fairly easy one to find |
| 47 | Double solutions: 1.Kb7/Kc7! Kb5 2.Bc2 Kb4 3.Qa4#. Duals in the intention: 2.Kb7/Kc7. Also dual after 2...Kb4 3.Qc5#. Key piece out of play in the initial position. | nice to engage the queen in such geometry... there are however several alternative solutions |
| 48 | After a major promotion key taking a provided flight, the play is dualistic: 1...Sf8 2.Qh5+/Qe1+/Qe4+/Qe7+/Rh2+/Qe3/Qe5/Qf7/Qb8/Qc8/Qxf8/. Same apply to other variations. | so many alternatives on the 2nd move... |
| 49 | Major promotion key takes three flights. Duals in the main variation after 2...Ke5 3.Ra5# or 1...g3 2.Ra6+/Kd4/Ke4. Spoilt by duals in all variations. | nice geometry... but certainly not the unique solution |
| 50 | Double solutions: 1.h8=R! ~ 2.Rg8+ Kh1 3.Rf1# and 1.Kg3! ~ 2.h8=Q ~ 3.Qh2#. Duals in the intention: 2.Qg7+/Qxe5/Kg3. Bad key (major promotion) and several threats. | most obvious |
| 51 | Promotion key takes a provided flight and introduces a short threat. Just one line, with no duals. | unique, but too obvious solution |
| 52 | Give (e5) and take (c6, e6) key by an en prise piece. The rest is forced, in spite of what black moves. No duals. | the mating net by the knight and bishop is nice... |
| 53 | Interversion of moves possible: 1.Rb4! Kc5 2.Rgb6 Kd5 3.Ra5#. Double solution: 1.Ra4! Kc5 2.K any (or R waiting) Kd5 3.Rb5#. Dual in intention: 2.Ra4 Kd5 3.R2b5#. | too elementary |
| 54 | Duals in the main variation: 1...Bb6 2.Qd5+/Qg5+/Qd7+ and 2...Kb4 3.Ra4/Rc4#. Again major promotion key takes flight and threatens a short mate. | diagonal checks on white squares vs. the opposite color bishop is a nice choice, but still fairly plain |
| 55 | Dual in the main variation: 3.Qb5#. The key played by en prise wQ threatens a short mate (2.Sc4#). The alternate black defense 1...Se3 is dualistic: 2.Qc7+/Bd3+/Bd7+/Bf1+/Be2+/Bc4+. | quite pretty **relatively complex... although I wouldn't be too surprised if the computer composed this one (the absence of pawns reduce the complexity in the computational sense)** |
| 56 | Key takes black piece with major promotion. Dual 2.Qb1 after 1...Kd2. Duals also after 1...Ke3 2.Qb2/Qd8 - not only 2.Qb3+. | too obvious |
| | Not exciting play. | |
| 57 | Flight taking key, capturing the only black officer and threatening two short mates. Dual after 1...Kh4 2.Qg5+/Qxg6/Sf2. | too obvious... e.g. a unique solution without capturing the knight (at least not on the first move) would be much prettier |
| 58 | Dual 2.Kc3 in the main variation. Bad key, taking three flights. Again this must be compared with Shinkman's cited at no 13. | obvious |
| 59 | Key captures the remaining black officer, but provides a flight. Two ideal mirror echo mates delivered thanks to zugzwang, the second after 1...Kc6 2.Rb4 Kd6 3.Rb6#.<br>**Two echo variations - something that humans will appreciate and try to create.** | Plain |
| 60 | Duals: 3.Rh5/Rh6#. Other not dualistic variations: 1...Kg6 2.Rxc6+ Kh7 3.Re7/Rh5/Rh6# and 1...Rxb6 2.Qf7+ Kh6(Kh8) 3.Rh5(Re8)#. Key takes three flights, though. | obvious... still, at least the rook wasn't captured immediately |
| 61 | Key takes three flights, defending the wS en prise. No real fight: white threat can not be actually stopped. | obvious, but looks efficient |
| 62 | Only two flights taken by key - defect would have been avoided by putting wQ initially on e7 (or Kb4 Qf6 Sf3 - Kd5; B: wQf6 -> e8). Continued zugzwang and midboard mate.<br>**An appealing diagonal mate. An equally appealing setting would be Kc4 Qd4 Sh4 - Ke4.** | nice geometry, unique solution... quite pretty<br>**very appealing... the quality of this problem seems to be much higher than of the average problem here... hence the decision** |
| 63 | Single line, but key is obvious, defending wR en prise. Black active distant selfblock is exploited in the mate.<br>**For the nice symmetry - asymmetry.** | nice geometry, unique solution, minor pieces... pretty.. still relatively obvious (hard to decide whether or not it was composed by a human)<br>**it was quite difficult to decide on this one... still, the minor pieces combine an appealing net... and it is probably relatively difficult computationally to operate with this subset of pieces** |
| 64 | Flight taking first move followed by flight taking second move with imparable mate. Pretty ordinary combination, often seen in practice. | obvious |
| 65 | Major promotion threatens three short mates. Main variation spoilt by many duals: 1...Bce7 2.Qb3+/Qb5+/Qb7+/Qb6/Qc7/Qc8. Second bB not justified. | diagonal check on white squares vs. the opposite color bishops.. but still not really beautiful |
| 66 | Clear single line, with forced continuation. Key takes flight and stops pawn's advance. bPe5 used in order to avoid stalemate.<br>**Just for the fact that Re4-c4 and Rc4-e4 are equipollent vectors and Kb1, Bc2, Re4 are aligned on the same diagonal. Many similar settings are possible.** | very straightforward |
| 67 | Double solutions: 1.Qg1/Qg6! with two unstoppable threats: 2.Qg2+ Kh4 3.Qg4# and 2.Qg4+ Kh2 3.Qg2# (also 1.Qf2 cooks). Duals in the intention: 2.Qc7/Qf2/Qg1. Flight taking key. | mating net with bishop and knight.. the second move (Qb8) is certainly more beautiful than 2.Qf2... |
| 68 | Nice minor promotion, but bad key taking flight. A better presentation of this idea was shown by Mackenzie (Kc6, Bb8, pc7 - Ka8): 1.Ba7 Kxa7 2.c8=R Ka6 3.Ra8# with stalemate avoidance.<br>**For the minor promotions. Human even composed the following: Ke4 pc7 d6 e7 f6 g7 - Ke6: 1.e8=B Kxd6(Kxf6) 2.c8=R(g8=R) Ke6 3.Rc6(Rg6)#. Are computers able to do something similar?** | underpromotion to a rook is nice... also this is a unique solution... with so little pieces.. quite pretty.. although easy to find |
| 69 | Dual 3.Qa6#. Again the key takes an unprovided flight and the second move captures the remaining officer, hence putting black in zugzwang. | not capturing the knight on the first move... still, an obvious solution |
| 70 | No surprise: key takes flight, second move takes main black officer and unavoidable mate follows. No real fight. | a unique solution... each piece takes part.. black succeeds promoting to a queen, still helpless against the timely mate... pretty (and quite humanlike if it was composed by the computer)<br>**such geometry is something that humans know to appreciate (and compose)** |
| 71 | Double solution: 1.d8=S! ~ 2.Ba7/Bb7/Bf2/Bg1 ~ 3.Sc7/Sf4#. In the intention Black can not parry the threat - actually no real fight. | many alternatives for black, but Qd8-c8-c4 is always decisive in time... quite pretty!<br>**the fact that the same maneuver wins in all variations makes me believe that it is very likely that a human composed this one** |
| 72 | En prise officer takes the remaining black piece, which was almost promoting and threatening to capture another white officer. The rest is forced, with interesting pawn mate. | pretty... it's a pity that the pawn must be taken! |
| 73 | Typical ending for normal game: flight taking key, followed by a flight taking second move with unavoidable mate. Lone bK | obvious |
| | hunting. | |
| 74 | Key takes three flights. Black is in zugzwang and white plays 2.Rg2 and 3.Qe2# regardless black's moves. Nothing really exciting. | not so trivial... but also not very beautiful |
| 75 | Duals 2.Qd3+/Qe4+/Qf5. Flight taking key guarding also key squares a6 and d3. | this one is nice... not easy, unique solution, geometry **with these pieces there are many possible checkmates, but this one is quite appealing... and that makes it more likely in my opinion that a human composed it** |
| 76 | Key takes bR threatening wQ. Neat mate in the main variation, similar with 1...Kc4. Dual after 1...Kb4 2.Kb2/Kc2 Kc4 3.Ra4#. Black Rook could be replaced by black S/B. **For the appealing diagonal mate (like 62). Another appealing setting would be Kc1 Qc2 Bc5 - Kc3.** | Obvious |
| 77 | Key takes bR threatening wR and also three flights. Similar second variation: 1...Kf2 2.Rb3 Kf1 3.Rf3#. Black Rook could be removed when wRe6 is shifted initially on b6. **Mirror ideal echo mates - long time humans considered them a symbol of perfection. Nowadays, however, strategic school prevails.** | most straightforward |
| 78 | Horror: two promoted black Bishops can't do anything against the threat. Key takes three flights. No real fight. | white diagonals vs. black colored bishops... quite pretty |
| 79 | Double solutions: 1.d8=R! f1=Q+ 2.Bxf1 Ka4 3.Ra8# and 1.Bf1! Ka4 2.d8=Q[R] Ka3 3.Q[R]a8#. Dual in the intention: 3.Qa8#. The key threatens two short mates. No real fight from Black. | promoting to queen is obvious |
| 80 | Duals in the main variation 3.Sb2/Se3# and 2.Qb2+/Sa3/Ke2. Major promotion threatens short mate and takes flight. | straightforward |
| 81 | Typical endgame mate. Key takes flight and threatened mate can't be avoided. With similar material better composition were shown (Ke7 Ba1 Sh6 - Kh8 pg7 h7 B: Rotate table 180). **A typical position to be shown by humans, especially when explaining the finale KBS vs K.** | just in time mate... unique solution (actually, also the only one that wins)... quite nice **so efficiently pretty... the knight must go to a square previously occupied by the king, which makes the solution a bit harder to find... that may have been one idea of the (human) composer** |
| 82 | Double solution: 1.Qb3! Dual in intention 2.Qf3 Kh2 3.Qg2# | plenty of solutions, and this geometric one is certainly not the ugliest |
| 83 | Key takes flight g8 and threatens short mate. No black defense can change anything, except 1...Ra1 2.Rxa1 and 3.Rh1#. Mate on the file is often met. | obvious |
| 84 | Key takes two flights and threatens short mate. Other variations: 1...Re3 2.Rf7 and 1...Rd7 2.Bxd7/exd7. | not most obvious... not too pretty either |
| 85 | Key takes flight g3 and threatens short mate on the file. Other variation: 1...Bg8 2.Rxg8. | straightforward |
| 86 | Key takes flight thanks to major promotion. Short mates if black plays other moves. Dual after 2...Kd6 3.Qc6/Qd8# | promoting to queen on the first move is generally not the prettiest way... also, white has quite a lot of pieces... |
| 87 | Double solutions: 1.Rxf7! ~ 2.Rg8 ~ 3.Rh7#; 1.Rd2! ~ 2.Re1 ~ 3.Sf3/Sh3# and 1.Re2! ~ 2.Rd1 ~ 3.Sf3/Sh3#. Duals in intention: 2.Rd3/Rd4 ~ 3.Rh3/Rh4#. Bad key taking a flight. | not capturing the pawn would be somewhat prettier... the solution is very obvious |
| 88 | Key takes flight, with wB leaving from en prise position. The long threat is actually unavoidable. However the line opening for wBe8 is neat. | the moves are obvious, but the setup is relatively pretty **perhaps this one is not so convincing... could have been a computer, the solution is somewhat too obvious... not a composition of a high quality.. still, interesting for beginners of chess...** |
| 89 | Duals in the main variation: 3.Qa5# and 2.Sd3 also work. Key immobilizes bK, by taking three flights. | efficient, but obvious |
| 90 | Double solution: 1.a8=R! [2.Rb8 Kc5 3.Rxc3#] 1...Kb5/Kc5 2.Rxc3[+] ~ 3.Qb3# 1...Kb3 2.Rb2+ Kc4 3.Rc8# and 1...Kb4 2.Qd5 B~ 3.Qb7/Qc4#. Duals in intention: 2.Qa6/Qd5/Qd6+. | lots of alternatives, very obvious |

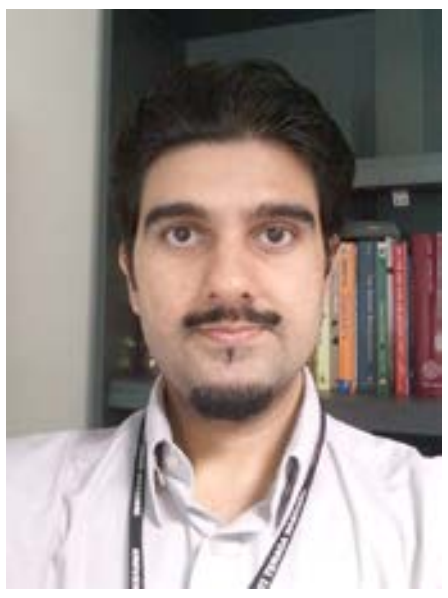

**Azlan Iqbal** received the B.Sc. and M.Sc. degrees in computer science from Universiti Putra Malaysia (2000 and 2001, respectively) and the Ph.D. degree in computer science (artificial intelligence) from the University of Malaya in 2009. He has been with the College of Information Technology, Universiti Tenaga Nasional since 2002, where he is senior lecturer. He is a member of the IEEE and AAAI, and chief editor of the electronic Journal of Computer Science and Information Technology (eJCSIT). He has won numerous awards and published dozens of reputable scholarly articles. His research interests include computational aesthetics and computational creativity in games.

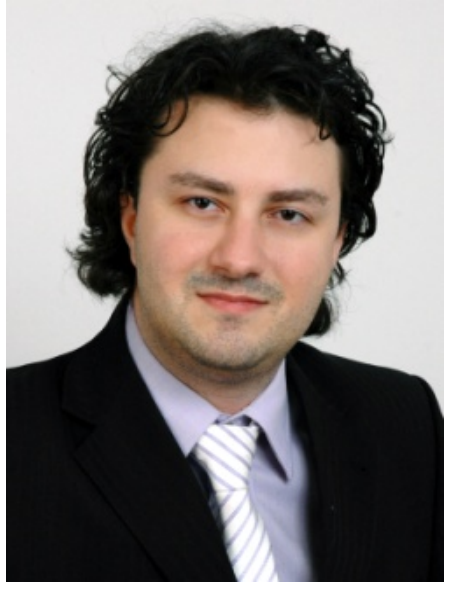

**Matej Guid** received his B.Sc. (2005) and Ph.D. (2010) degrees in computer science from the Faculty of Computer and Information Science at the University of Ljubljana, Slovenia. He is a assistant professor at the Artificial Intelligence Laboratory, University of Ljubljana. His research interests include heuristic search, computer game-playing, automated explanation and tutoring systems, and argument-based machine learning. Chess has been one of his favorite hobbies since childhood. He was also a junior champion of Slovenia a couple of times, and holds the title of FIDE master.

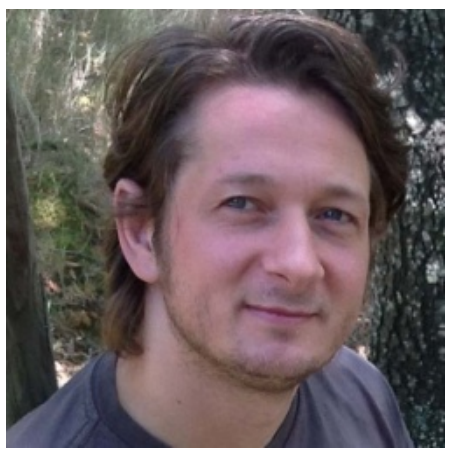

**Simon Colton** is Professor of Computational Creativity and EPSRC Leadership Fellow in the Department of Computing of Goldsmiths College, University of London; and previously Reader in Computational Creativity at Imperial College, London. He heads the Computational Creativity Group, which studies notions related to creativity in software. He has published over 120 papers on AI topics such as machine learning, constraint solving, computational creativity, evolutionary techniques, the philosophy of science, mathematical discovery, visual arts and game design. He is the author of the programs HR and The Painting Fool.

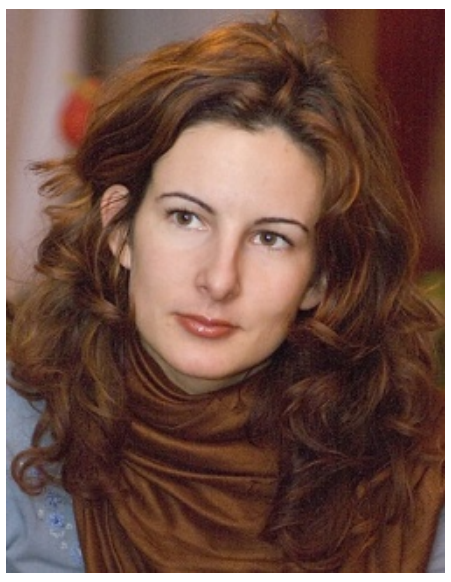

**Jana Krivec** received her B.Sc. in psychology at the University of Ljubljana in 2004 and her PhD in psychology (cognitive science) in 2013 at the same university. She works as a researcher at Jožef Stefan Institute since 2004. Her research interests include cognitive science combined with artificial intelligence, in particular, the game of chess. Notably, she is also a chess woman grandmaster and several times Slovenian champion.

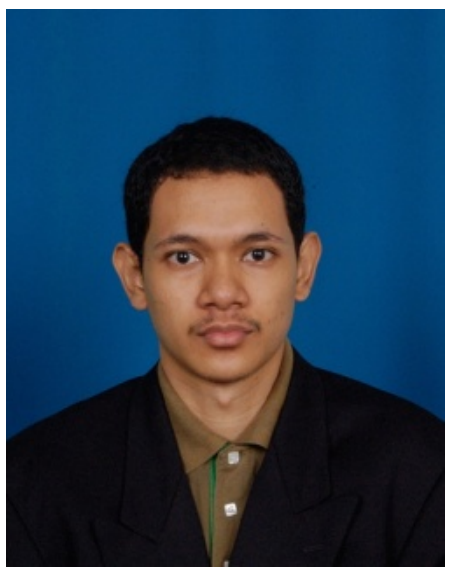

**Shazril Azman** earned Diploma in Information Technology (IT) in 2009 and Bachelor's degree in Graphic and Multimedia in 2012. He has been involved in projects and workshops related to web development, graphic design, computer animation and augmented reality throughout his years of study. He currently is pursuing his Master's degree in Information Technology. His areas of interest are Artificial Intelligence, arts and Computational Creativity.

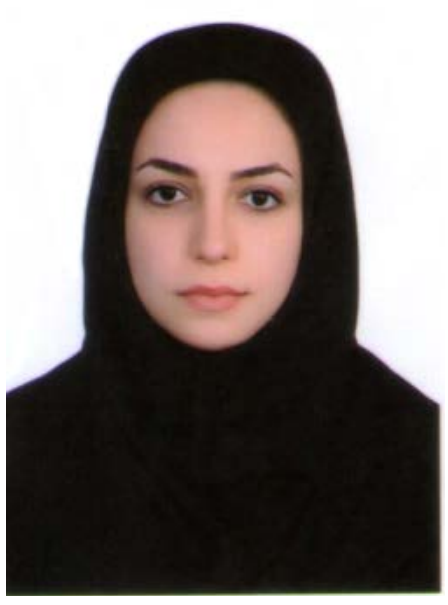

**Boshra Talebi Haghighi** (originally from Iran) received her B.Sc. in Systems and Networking (Hons. 2009) and M. Sc. in Information Technology (2012) from Universiti Tenaga Nasional, Malaysia. She served as research assistant for 18 months to Azlan Iqbal under the eScienceFund research grant (01-02-03-SF0240). Her main interest is working as academic and researcher. Her research interests include Human Computer Interaction, E-Learning, Information Systems and Artificial Intelligence.